\documentclass[]{fairmeta}

\title{ARE: scaling up agent environments and evaluations}

\author{Meta Superintelligence Labs\textsuperscript{1}
\newline
\textnormal{\textsuperscript{1}A detailed contributor list can be found in the appendix of this paper.}
}

\abstract{
We introduce \textbf{Meta Agents Research Environments (\are)}, a research platform for scalable creation of environments, integration of synthetic or real applications, and execution of agentic orchestrations. \are\ provides simple abstractions to build complex and diverse environments, each with their own rules, tools, content, and verifiers, helping to bridge the gap between model development and real-world deployment.
We also propose \textbf{Gaia2}, a benchmark built in \are\ and designed to measure general agent capabilities. Beyond search and execution, \gaiatwo\ requires agents to handle ambiguities and noise, adapt to dynamic environments, collaborate with other agents, and operate under temporal constraints. Unlike prior benchmarks, \gaiatwo\ runs asynchronously, surfacing new failure modes that are invisible in static settings. Our experiments show that no system dominates across the intelligence spectrum: stronger reasoning often comes at the cost of efficiency, and budget scaling curves plateau, highlighting the need for new architectures and adaptive compute strategies.
Perhaps more importantly, \are\ abstractions enable continuous extension of \gaiatwo\ to other environments, empowering the community to rapidly create new benchmarks tailored to their domains. In AI’s \href{https://ysymyth.github.io/The-Second-Half/}{``second
 half''}, progress increasingly depends on defining meaningful tasks and robust evaluations to drive frontier capabilities forward.
}
\date{\today}
\correspondence{
\email{rfroger@meta.com},
\email{amineben@meta.com},
\email{gmialon@meta.com},
\email{tscialom@meta.com}
}

\metadata[Code]{\url{https://github.com/facebookresearch/meta-agents-research-environments}}
\metadata[HF Space]{\url{https://huggingface.co/meta-agents-research-environments}}

\usepackage{xspace}

\newcommand{\are}{ARE\xspace}
\newcommand{\gaiatwo}{Gaia2\xspace}
\newcommand{\gaia}{Gaia\xspace}

\newcommand{\readact}{\texttt{read}\xspace}
\newcommand{\writeact}{\texttt{write}\xspace}
\newcommand{\env}{\texttt{Env}\xspace} 
\newcommand{\user}{\texttt{User}\xspace} 
\newcommand{\agent}{\texttt{Agent}\xspace} 
\newcommand{\myenv}{\texttt{Mobile}\xspace} 
\newcommand{\chats}{{Chats}\xspace}

\newcommand{\sendmessagetouser}{\texttt{send\_message\_to\_user}\,}
\newcommand{\sendmessagetoagent}{\texttt{send\_message\_to\_agent}\,}

\newcommand{\llamaIII}{Llama 3.3 70B Instruct\xspace}

\newcommand{\claudeSonnet}{Claude Sonnet 3.7\xspace}
\newcommand{\Gemini}{Gemini 2.5 Pro\xspace}

\newcommand{\verifier}{\are Verifier\xspace}
\newcommand{\inContextVerifier}{In-context Verifier\xspace}

\usepackage{listings}
\usepackage{xcolor}
\usepackage{float}
\usepackage{bbm}
\usepackage{wrapfig}
\usepackage[dvipsnames]{xcolor}

\definecolor{backcolour}{rgb}{0.95,0.95,0.92}
\definecolor{codegreen}{rgb}{0,0.6,0}
\definecolor{codegray}{rgb}{0.5,0.5,0.5}
\definecolor{codepurple}{rgb}{0.58,0,0.82}
\definecolor{keywordcolor}{rgb}{0.8,0,0.8}
\definecolor{stringcolor}{rgb}{0.6,0.1,0.1}

\lstdefinestyle{pythonstyle}{
    backgroundcolor=\color{backcolour},   
    commentstyle=\color{codegreen},
    keywordstyle=\color{keywordcolor},
    numberstyle=\tiny\color{codegray},
    stringstyle=\color{stringcolor},
    basicstyle=\ttfamily\footnotesize,
    breakatwhitespace=false,         
    breaklines=true,                 
    captionpos=b,                    
    keepspaces=true,                 
    numbers=left,                    
    numbersep=5pt,                  
    showspaces=false,                
    showstringspaces=false,
    showtabs=false,                  
    tabsize=2,
    language=Python,
    morekeywords={def, class, self, pass, str, import, from, as},
    morestring=[b]",
    morestring=[b]',
    morestring=[b]""",
    morecomment=[l]{\#},
}

\begin{document}

\maketitle

\section{Introduction}

\label{section:intro}
\begin{figure}[!b]
    \vspace{-0.6cm}
    \centering
    \includegraphics[width=0.9\linewidth]{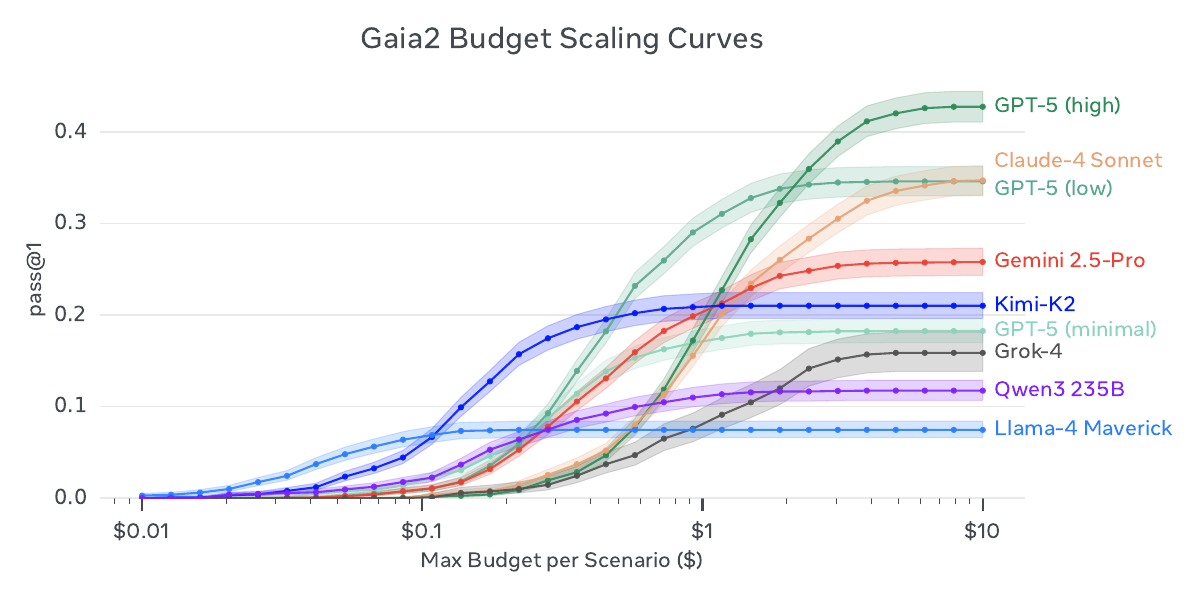}
    \vspace{-0.4cm}
    \caption{
    \gaiatwo budget scaling curve: for each $\text{max\_budget}$, we plot $\sum \mathbbm{1}\{\text{scenario\_result} = \text{True} \land \text{scenario\_cost} < \text{max\_budget}\}$. Equipped with a simple ReAct-like scaffold (see Section~\ref{sec:agent_orchestration}), no model evaluated here dominates across the intelligence spectrum—each trades off capability, efficiency, and budget. At equal cost, some models fare better, yet all curves plateau, suggesting that standard scaffolds and/or models miss key ingredients for sustained progress. Cost estimates from \href{https://artificialanalysis.ai/models}{Artificial Analysis} model pricing data (accessed September 10, 2025).
    }
    \label{fig:gaia2_scenario_budget_scaling_curve}
    \vspace{-0.3cm}
\end{figure}

Scaling large language model (LLM) training with reinforcement learning (RL) is a promising path towards continuous model improvements and, eventually, superintelligence. In particular, reinforcement learning from verifiable rewards (RLVR) has recently emerged as a more scalable alternative to reliance on reward models in settings like reasoning, coding~\citep{openai2024openaio1card,openai2025openaio3o4minicard,deepseekai2025deepseekr1incentivizingreasoningcapability,mistralai2025magistral}, agent tool use~\citep{moonshotai2025kimik2}, or even chat~\citep{mu2024rulebasedrewardslanguage}.
Concurrently, models are now addressing tasks involving deeper interactions with the outside world over longer time periods, as reflected by the emergence of new benchmarks~\citep{mialon2023gaia,jimenez2024swebench,yao2024taubenchbenchmarktoolagentuserinteraction,backlund2025vendingbenchbenchmarklongtermcoherence} and products~\citep{google2024deepresearch,openai2025deepresearch}.
 
In this context, we posit that model improvement through experience and deployment in production are bounded by the controllability, diversity, and realism of available environments. First, while the web is a great environment for supporting agent tasks like search,
it is constantly evolving, making reproducibility for evaluation~\citep{mialon2023gaia, wei2025browsecompsimplechallengingbenchmark} and study of complex behaviors challenging, in particular those involving \writeact operations.
Creating simulated environments is an appealing alternative providing more control to developers, but requires spending sufficient effort on diversity and realism. 
Existing environments are therefore tightly coupled to narrow sets of tasks and capabilities, and even agent modeling. Since environments saturate quickly with model progress, moving to new environments and tasks requires frequent rewriting of a lot of boilerplate code. As of the writing of this paper, there are few open-source and flexible libraries for developing and studying practical LLM agents~\citep{brown_verifiers_2025}.

Second, most environments reflect idealized models of agent interaction. These idealized models reduce task diversity and do not map to real-world deployment conditions. For example, $\tau$-bench~\citep{yao2024taubenchbenchmarktoolagentuserinteraction} and SWE-bench~\citep{jimenez2024swebench} agents operate sequentially and the environment is paused while the agent is working, preventing the state of the world from changing in the interim and effectively giving away many valuable real-world capabilities, such as asynchronous communication with users and adaptation to new events.

We therefore propose Meta Agents Research Environments (\are), a research platform that supports the running of orchestrations, creation of environments, and connection of synthetic or real world apps for agent development and evaluations. \are\ does so by: (i) proposing abstractions for simulation and verification that facilitate both the creation of diverse environments and tasks, as well as the integration of existing ones, like $\tau$-bench; and (ii) supporting a shift from sequential to asynchronous interaction between an agent and its environment, unlocking new tasks and capabilities in the process, like handling time. Though simulated, the platform is not unrealistic. \are\ supports connection of real apps e.g., through Model Context Protocol (MCP) integration~\citep{anthropic2024mcp}, so that model development, evaluation, and production deployment can be consistent. In addition to RL, \are\ enables generation of high-quality SFT traces.

Building on \are, we introduce \gaiatwo, a new evaluation for agents. \gaiatwo\ is composed of 1,120 verifiable, annotated scenarios taking place in a \myenv\ environment that mimics a smartphone, with apps such as email, messaging, calendar, etc. as well as their associated content, similar to AppWorld~\citep{appworld-acl24} or ToolSandbox~\citep{lu2024toolsandboxstatefulconversationalinteractive}. \gaiatwo\ is designed to address the need for challenging yet holistic evaluations for agents beyond pure search-and-execution~\citep{mialon2023gaia,wei2025browsecompsimplechallengingbenchmark,yao2024taubenchbenchmarktoolagentuserinteraction,appworld-acl24,lu2024toolsandboxstatefulconversationalinteractive}. 
\gaiatwo retains the core principles of \gaia, consisting of verifiable tasks that are simple for humans but challenging for today's models, and that align with actual model use. The benchmark also integrates our learnings from using \gaia. We propose more diverse tasks in a simulated but dense and realistic environment with built-in tools so that signal-to-noise ratio and reproducibility are better. Furthermore, we target new capabilities with scenarios that require agents to exhibit adaptability and to effectively handle ambiguity, noise, time, and collaboration with other agents. If developed, these new capabilities will unlock a breadth of practical use cases.

\gaiatwo\ departs from most agent benchmarks in two ways. (i) Deeper and more realistic interactions between the agent and the environment, since both run asynchronously. In particular, we shift from tasks to scenarios spanning an arbitrary period of time. Environment time passes independent of whether the agent acts or not, and the environment state is continuously updated with random or scheduled events, such as a friend replying to message sent by a user or an agent. This unlocks a breadth of new scenarios requiring the agent to adapt at test-time to either ignore incoming events or to be always-on and proactive, performing actions in due time. (ii) We propose a robust verification system that lends itself to RL, comparing agent \writeact\ actions only to annotated oracle \writeact\ actions, and for each \writeact\ action, evaluating arguments, that can be seen as rubrics, via a soft (LLM judge) or hard (exact-match) comparison depending on the argument type.

While today's frontier models are far from solving \gaiatwo, we do not consider it to be an ``AGI-level'' benchmark; in the LLM and RL era, we expect rapid hill-climbing. However, we believe \gaiatwo, with its richer interactions with an environment and resulting spectrum of scenarios, is aligned with actual progress towards practically useful agents. We also expect that doing well on multi-agent and time-based scenarios will require some modeling effort beyond scaling training. Finally, \gaiatwo\ lends itself to extensions. There is no progress without reliable evaluations, and we hope removing the need for writing boilerplate environment and runtime code will help the community continue creating benchmarks that challenge current modeling standards.

\begin{figure}[t!]
    \centering
    \includegraphics[width=0.95\textwidth]{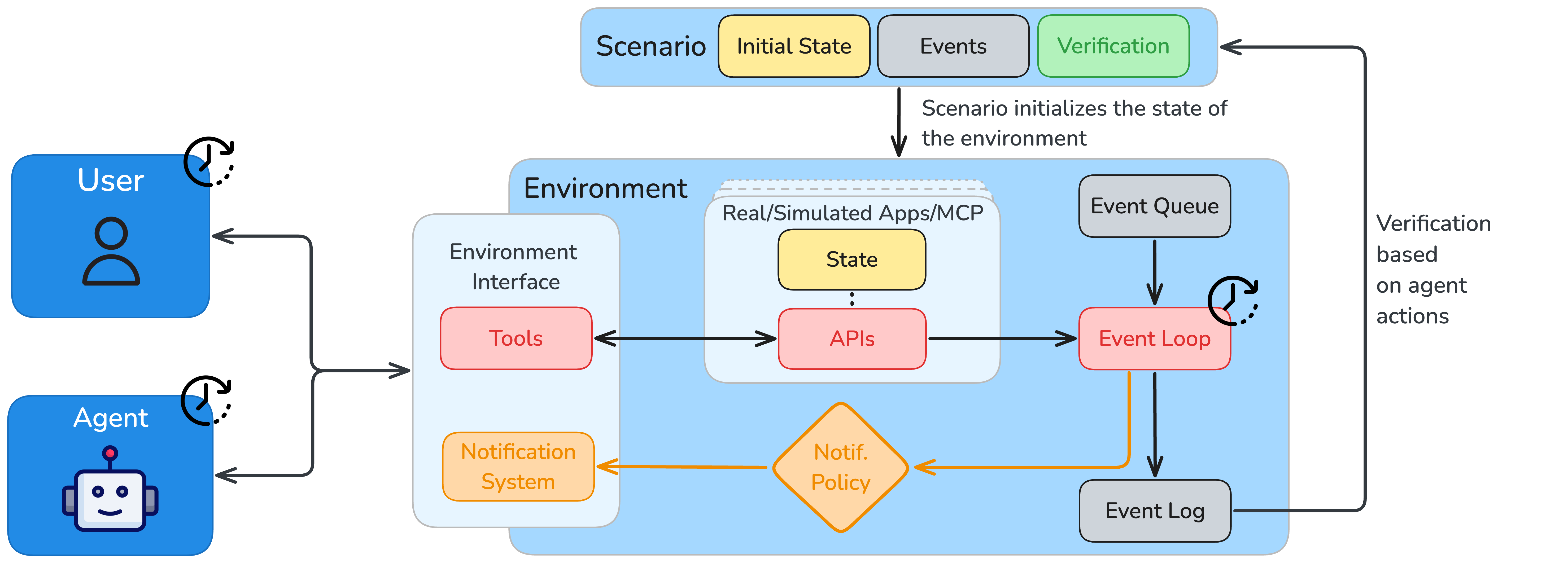}
    \caption{\are\ environments are event-based, time-driven simulations, that run asynchronously from the agent and the user. \are\ environments allow to play scenarios, which typically contain tasks for the agent and verification logic. Whether initiated by agent or user, interactions happen through the same interfaces and can be either tool calls, or tool output/notification observations. Extensive simulation control and logging allow precise study of agents behavior.}
    \label{fig:are_high_level}
\end{figure}

\section{\are: A Research Platform to Create Environments and Run Agents}

\are\ is a research platform for creating simulated environments, running agents on scenarios within them, and analyzing their behavior.
\are\ environments evolve continuously and are strictly decoupled from the agent. Time advances in the simulation, and the environment continuously introduces events. The agent runs asynchronously and interacts with the user and the environment through a dedicated interface.
\autoref{fig:are_high_level} provides an overview of \are’s architecture and high-level abstractions.

\subsection{\are\ Foundations}

\are\ is time-driven and built on the principle that ``\textbf{everything is an event}’’. Specifically, five core concepts work together:
\begin{enumerate}
\item \textbf{Apps} are stateful API interfaces that typically interact with a data source.
\item \textbf{Environments} are collections of Apps, their data, and governing rules that define system behavior.
\item \textbf{Events} are anything that happens in the Environment. All Events are logged.
\item \textbf{Notifications} are messages from the Environment that inform the agent about Events. They are configurable and enable selective observability of the Environment.
\item \textbf{Scenarios} are sets of initial state and scheduled Events that take place in an Environment, and can include a verification mechanism.
\end{enumerate}

\subsubsection{Apps}

Apps are collections of tools that interact with a data source. For instance, an Emails app contains tools like \texttt{send\_email} and \texttt{delete\_email} that all operate on the same email database. Similar approaches have been explored in AppWorld~\citep{appworld-acl24} and ToolSandbox~\citep{lu2024toolsandboxstatefulconversationalinteractive}.

\paragraph{Apps maintain their own state} \label{sec:app_stateful}
Each app starts in the simulation with an initial state and keeps track of changes as agents use its tools or as events occur in the environment. Apps store their data internally rather than relying on external databases. This design makes it convenient to study agent tasks that require to modify the state of the environment, and ensures that experiments can be reproduced consistently.

\paragraph{Tool creation and taxonomy} \label{sec:app_api_abstraction} 
Apps are implemented by adding Python methods within an \texttt{App} class. When the simulation runs, these methods are automatically converted into properly formatted tool descriptions that agents can understand and use.
\are classifies tools into two types via decorators: \readact, which only read app states (e.g., \texttt{search\_emails}), and \writeact, which modify app states (e.g., \texttt{send\_email}).
\are classifies tools into two types via decorators: \readact, which only read app states (e.g., \texttt{search\_emails}), and \writeact, which modify app states (e.g., \texttt{send\_email}).
This distinction is helpful \textit{e.g.} for verification, see Section~\ref{sec:are-verifier}. Tools are role-scoped—\texttt{agent}, \texttt{user}, or \texttt{env}. For example, certain user tools may be unavailable to the agent due to sensitivity. See Appendix~\ref{app:app_tool_creation} for example code snippets.

\paragraph{Extensibility} \label{sec:extensibility}
Beyond \textit{ad hoc} app creation, \are\ can also connect with external APIs through MCP compatibility~\citep{anthropic2024mcp}. The framework also offers flexible options for data storage. While our current implementation stores data in memory, users can easily connect SQL databases or other storage systems without changing the core framework.

\paragraph{Core apps} 
Developers can choose which apps to include in their environment or create new ones. However, every \are\ environment includes two core apps that handle the basic interaction between agents and their environment:
\begin{itemize}
    \item \texttt{AgentUserInterface} is the communication channel between users and agents: messages are tool calls, and user messages generate notifications (Section~\ref{sec:notification_system}) that agents can process asynchronously. This enables asynchronous interactions during task execution. The interface supports two modes: \emph{blocking} (the agent waits for a user reply) and \emph{non-blocking} (the agent continues loop regardless of reply).
    
    \item \texttt{System} provides core simulation controls like \texttt{get\_current\_time} (query time), \texttt{wait} (pause for a duration), and \texttt{wait\_for\_next\_notification} (pause until an event). When any wait tool is invoked, the simulation accelerates: it switches from real time to a queue-based, event-to-event loop. Scenarios that would take hours in the real world can thus run in minutes, enabling practical long-horizon testing.
\end{itemize}

\subsubsection{Environment}
An environment is a Markov Decision Process with states, observations, actions, and transition rules. The environment state includes the states of all apps, the time manager, and the notification system. Apps define the action space by exposing their tools.
The environment runs deterministically given a fixed starting state and seed, ensuring reproducible evaluations. It can host one or multiple agents simultaneously, supporting both single-agent and multi-agent setups. The environment's rules define time progression, action permissions, reward computation, and how agent actions affect the environment state.

\subsubsection{Events}
\label{sec:are_events}

In \are, an event is any agent action or app-state change. Each event is timestamped, logged. Events can be scheduled, e.g., a friend’s message 1 minute after simulation start. This design yields (i) \textit{deterministic execution}—events run in scheduled order; (ii) \textit{complete auditability}—all actions can be replayed and analyzed; and (iii) \textit{flexible scheduling}—events can be set at absolute times or relative to others.

\paragraph{Event lifecycle} Events flow through four stages described in \autoref{fig:are_high_level}: (i) \textit{creation} - events are created from tool calls or scheduled by the simulation; (ii) \textit{scheduling} - events enter a time-ordered \texttt{EventQueue} with dependency management using directed acyclic graphs, supporting both absolute timing (at specific timestamps) and relative timing (relative to other events or conditions); (iii) \textit{execution} - the \texttt{EventLoop} processes events and captures results, state changes, and exceptions; and (iv) \textit{logging} - executed events are stored in an \texttt{EventLog} with detailed metadata for analysis, debugging, and validation of agent behavior.

\paragraph{Event types} There are different types of events. While most events track interactions within the environment, other special events are needed to enable dynamic scenarios and verification strategies:

\begin{itemize}
    \item \textbf{Agent/User/Env events} are generated by tool calls. \textit{Agent Events} are initiated by the agent (e.g., sending a message), \textit{User Events} by the user (e.g., replying to the agent), and \textit{Environment Events} by the simulation itself to introduce external changes (e.g., a scheduled message from a friend).

    \item \textbf{Conditional events} periodically check predefined conditions and complete when criteria are met (e.g., cancel a ride only if one was booked).

    \item \textbf{Validation events} check milestone achievement or constraint violations for verification, and fail the simulation if not completed on timeout (e.g., stop if no ride is booked within 30 seconds of the user request).

    \item \textbf{Oracle events} are pre-scheduled ``ground truth” actions used by a verifier for comparison (see Section~\ref{sec:are_foundations_scenarios}).

\end{itemize}

\begin{figure*}[t]
    \centering
    \includegraphics[width=0.8\textwidth]{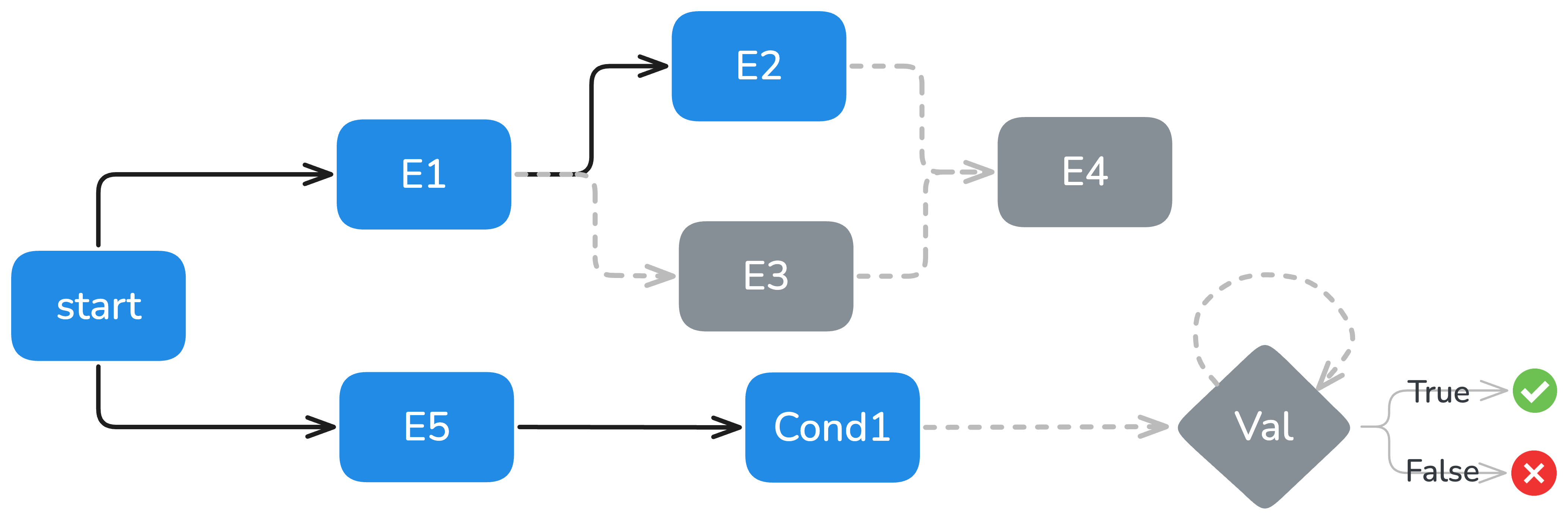}
    \caption{Event dependency graph illustrating \are\ scheduling patterns. Events \texttt{E1} and \texttt{E5} execute in parallel after simulation start, and \texttt{E2}/\texttt{E3} executing in parallel after their prerequisites, both need to be executed for \texttt{E4} to execute. Conditional execution is shown through \texttt{Cond1} leading to validation (\texttt{Val}) with true/false outcomes.}
    \label{fig:are_foundations_complex_dag}
\end{figure*}

\paragraph{Dependencies and scheduling} Events are modeled as Directed Acyclic Graphs (DAGs) as illustrated in \autoref{fig:are_foundations_complex_dag}. An event can only be triggered upon successful completion of all its predecessors (\textit{e.g.}, \texttt{e1} processes immediately at simulation start, \texttt{e4} needs both \texttt{e2} and \texttt{e3} to be completed). This data structure also supports multiple branches running simultaneously to model independent events. Conditional and Validation events can be used in the DAG to trigger other events and make the environment more dynamic.

\subsubsection{Notification System}
\label{sec:notification_system}
At each environment step, processed events can trigger notifications according to a notification policy (see \autoref{fig:are_high_level}), similar to mobile device notifications. 
Apart from tool outputs, notifications are the only signals agents receive from the environment.
Notifications are queued by timestamp and exposed to agents through a notification queue, enabling asynchronous interactions (\autoref{fig:async_interaction}). In our orchestration (Section~\ref{sec:agent_orchestration}), notifications are injected into the agent's context at the beginning of each agent step.

\paragraph{Notification policy}
The notification system follows a configurable policy—i.e., a whitelist of events authorized to emit notifications.
\are\ pre-defines three verbosity levels: \texttt{low} (only user messages are notified), \texttt{medium} (emails, messages and calendar events are notified), and \texttt{high} (everything is notified), creating a graduated spectrum of environmental observability. More details on notification policies are given in Appendix \ref{appendix:notifications}.

\paragraph{Notifications and agent proactivity} 
Notifications are not the only way for agents to observe environment changes. For example, even if the notification policy doesn't alert the agent when messages arrive from contacts, the agent can still proactively check for new messages by browsing the user's inbox. 
Notifications add realism and complexity to environments, potentially creating different agent behaviors based on whether the environment is notification-rich or notification-poor. This system enables researchers to tackle new capabilities such as proactivity.

\subsubsection{Scenarios}
\label{sec:are_foundations_scenarios}
\are\ shifts from static, single-turn tasks to dynamic \textit{scenarios}. Scenarios attempt to capture real-world complexity through temporal dynamics, events, and multi-turn interactions. This enables evaluation of agent capabilities that cannot be assessed through traditional request-response paradigms. In practice, scenarios are implemented in a \texttt{scenario.py} containing the apps, scheduled events, and arbitrary verification logic. Appendix~\ref{app:scenario_py} provides more details.

\begin{figure}[b!]
    \centering
    \includegraphics[width=0.7\textwidth]{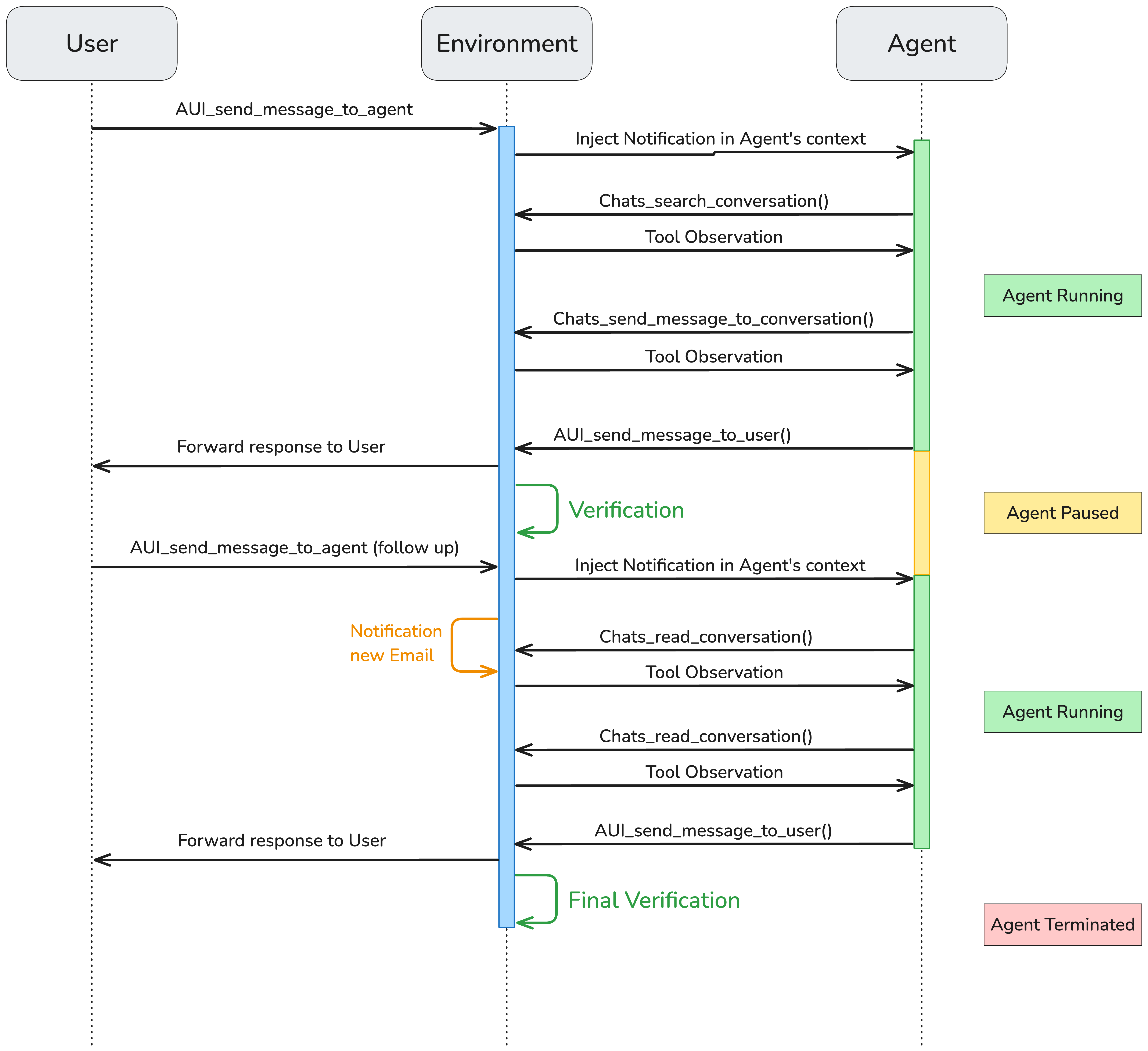}
    \caption{Sequence diagram of a multi-turn scenario in \are. The agent is paused between turns, i.e., between calling \sendmessagetouser and receiving \sendmessagetoagent, and adapts its strategy in response to an asynchronous notification from the environment, a new email.}
    \label{fig:sequence_diagram}
\end{figure}

\paragraph{Scenario runtime} Scenarios typically start with an environment instance and a \texttt{send\_message\_to\_agent} tool call, waking the agent up. The environment operates on discrete time steps, executing scheduled events and managing state transitions until the agent reaches an exit condition, see~\autoref{fig:are_foundations_complex_dag}.
All interactions with the user are through the \texttt{AgentUserInterface}, with verification triggered upon task completion.

\paragraph{Scenario example} Consider this two-turn scenario (see \autoref{fig:are_high_level} and \autoref{fig:sequence_diagram}): a user asks the agent via \texttt{AgentUserInterface} \textit{``Can you ask my mom to send me our family streaming password?"}. The agent is initialized from this first notification, starts checking messages, and requests the password in the \emph{Chats} app; the tool calls modify the \emph{Chats} app state and are recorded in the \texttt{EventLog}. 
The agent confirms to user that the request was sent, after which the environment pauses execution and applies first-turn validation.

At turn two, the user asks a follow up question:
\textit{``As soon as I receive the password from my mother, transfer it to my father”}. The agent resumes upon the \texttt{send\_message\_to\_agent} notification, and looks for the mother's reply in the \emph{Chats} app (where it previously requested it).
In the meantime, a scheduled environment event is triggered and an \emph{Email} from the mother containing the code is received. The agent reacts to this email notification by stopping searching the \emph{Chats} app, processes the \emph{Email}, extracts the code, forward it to the father, and report success to the user.
Final verification reviews the complete interaction in the \texttt{EventLog}, and the environment issues a termination signal to end execution.

\paragraph{Scenario hints} In addition to event DAGs, scenario creators optionally provide step by step solution to scenarios in natural language, which we call hints.
Hints serve multiple purposes: they help validate scenario correctness during QA by clarifying the approach intended by the original creator for solving the scenario. Hints are useful during RL training when scenarios prove too challenging for the agent - they can be rephrased and injected to provide high-level guidance.

\subsection{Mobile, an Initial Environment in \are} 
\label{sec:are-env}

We release \are\ with an environment analogous to a mobile device, \myenv, in which \gaiatwo lives. Mobile device environments offer a broad range of tasks with complex interactions between agents and their environment that are aligned with actual model use. \myenv is composed of a set of rules defining the interaction model between the agent and \myenv, and a set of apps, including their content, that are typically found on mobile devices such as \texttt{Messages}, \texttt{Contacts}, or \texttt{Calendar}, providing 101 tools in total, displayed in \autoref{fig:app_usage_distribution}.

\subsubsection{Environment Rules}

\myenv uses turn-based interaction. A turn begins when the agent receives a user instruction or environment event, and ends when the agent reports back to the user (via \texttt{send\_message\_to\_user}), signaling task completion or requesting further input.

During a turn, the environment operates asynchronously—the simulation clock advances while the agent processes information and selects actions. The agent's computational time directly consumes simulated time, making slow responses quantifiably impact the simulation. Between turns, the simulation pauses while awaiting user input. Scenarios terminate under three conditions:
\begin{itemize}
    \item \textbf{Successful completion}: Agent signals task completion via \texttt{send\_message\_to\_user} with no further user messages scheduled.
    
    \item \textbf{Constraint failure}: Agent exceeds predefined limits on simulation time, total steps, or number of turns.
    
    \item \textbf{Verification failure}: At the end of a turn, Agent actions do not pass verification, see Section~\ref{sec:are-verifier}.
\end{itemize}

\subsubsection{Environment Population}
\label{sec:are-universe}

\begin{wrapfigure}{r}{0.35\textwidth}
    \centering
    \includegraphics[width=0.35\textwidth, trim=0 15 0 0, clip]{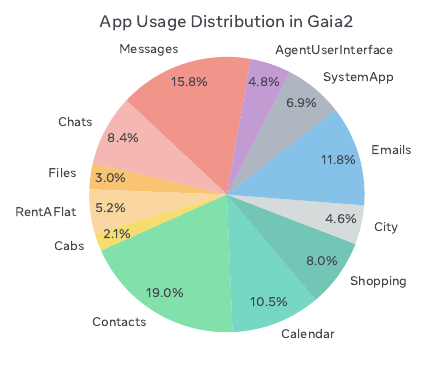}
    \caption{App usage distribution across the 12 \myenv apps in \gaiatwo for Llama 4 Maverick.}
    \label{fig:app_usage_distribution}
    \vspace{-0.5cm}
\end{wrapfigure}

We create content for \myenv apps with synthetic data generated using \llamaIII. The primary challenge lies in generating coherent data across all applications -- contacts in the \texttt{Contacts} app must match those in messaging apps, calendar events should align with user descriptions, etc. To address this, we define an app dependency graph (\autoref{fig:app_dependency}, Appendix~\ref{appendix:universe}) to guide the generation process, though more complex inter-app dependencies remain unhandled at this stage. The root node is a persona from PersonaHub \citep{ge2024scaling}, from which we infer plausible universe locations and countries as foundational information. 

To maximize diversity, we initiate populations by generating unstructured content, subsequently processed through structured decoding guided by individual app schemas. We repeat this process to create diverse \myenv instances—termed \emph{``universes''}—sharing identical rules and applications but containing distinct content. For example, one universe centers on a retired French physics professor, while another focuses on a Chinese professional athlete. Each universe contains approximately 400K tokens of raw unstructured content on average. When accounting for the complete structured representation, universes reach approximately 800K tokens. Both estimates constitute lower bounds, as they do not include the contents of the filesystem. Additional implementation details are provided in Appendix~\ref{appendix:universe}.

\subsubsection{Scenario Creation}
\label{sec:scenario_creation}

\are\ enables \myenv scenario creation through an annotation interface (not released) described in Section \ref{sec:are-annotation-ui}. The novelty of \myenv scenario creation is that it focuses on collecting the DAG of \writeact\ events as ground truth, including user, oracle, and environment events. The \are\ Verifier (Section \ref{sec:are-verifier}) validates that agent \writeact\ actions match annotated oracle actions. \myenv scenarios are designed such that there is a unique set of \writeact\ actions that solves the scenario, while \readact\ actions are not explicitly verified as they do not change the environment state. This encompasses the evaluation paradigm of the original Gaia benchmark, where the correct \writeact\ action consists of sending a message to the user with the final answer.

\subsubsection{Implementing other Environments with \are}

\myenv is an example of the environments that can be built in \are: beyond \myenv, \are\ abstractions encompass many existing agent benchmarks. For example, we internally replicated $\tau$-bench~\citep{yao2024taubenchbenchmarktoolagentuserinteraction} and BFCLv3~\citep{patil2025bfcl} in \are\ without major modifications. In particular for $\tau$-bench, a domain such as Airline is implemented as a single app environment, leverages \are\ LLM user abstraction, and Oracle Events are parsed from $\tau$-bench ground truth actions. The simulation is stopped with a Conditional Event monitoring that the agent defers to a human assistant or the user stops the interaction, and the trajectory is verified by implementing $\tau$-bench verification logic as validation events. Our MCP integration also allows for reproducibility within \are\ of MCP-based benchmarks \citep{wang2025mcpbenchbenchmarkingtoolusingllm, mcpmark_2025, gao2025mcpradarmultidimensionalbenchmarkevaluating}.

\subsection{An initial Verifier for \are}
\label{sec:are-verifier}

Verifiable rewards have proven crucial for improving  reasoning \citep{deepseekai2025deepseekr1incentivizingreasoningcapability,lambert2024tulu}, code generation \citep{gehring2024rlef}, agentic web browsing \citep{openai2025deepresearch,wei2025webagent} and software engineering \citep{yang2025swe,moonshotai2025kimik2}. Similarly, recent reasoning and agent benchmarks adopted short-formed answers that can be easily matched \citep{hendrycks2021measuring,mialon2023gaia}, or binary feedback from an execution environment \citep{yao2024taubenchbenchmarktoolagentuserinteraction,jimenez2024swebench}. We propose a rubric-like verifier for \are\ and \myenv\ checking each agent \writeact operation.

\subsubsection{Verification Mechanism}

\begin{figure}[t!]
     \centering
     \includegraphics[width=\textwidth]{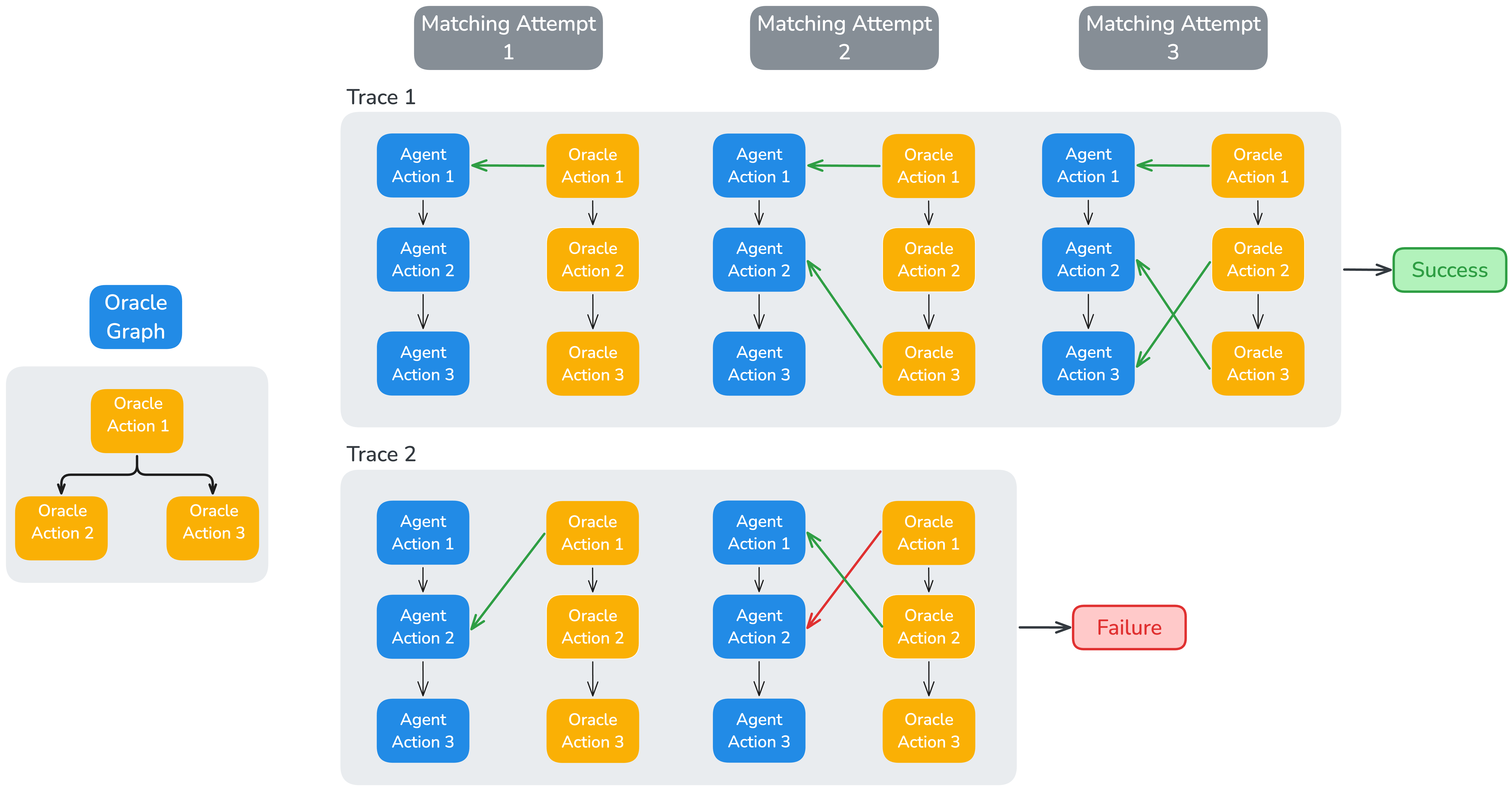}
    \caption{Illustration of a success (top) and a failure (down) of the matching trajectory process.}
    \label{fig:verifier}
\end{figure}

We verify scenario successful completion by comparing agent actions with a ground truth, defined as the minimal sequence of \writeact actions needed to solve a task. We exclude \readact actions from verification since multiple reading strategies can lead to the correct set of \writeact actions.
\autoref{fig:verifier} provides an overview of the verification procedure. In a preliminary phase, the verifier checks that used tool names counters are identical in both the oracle actions and the agent's \writeact actions. If this test is successful, the verifier sorts the oracle actions in a topological order based on the oracle graph, which reflects their dependencies. Then, the verifier proceeds to mapping each oracle action to an agent action by checking:
\begin{itemize}
    \item \textbf{Consistency:} the verifier tests whether the oracle action and the candidate agent’s action are equivalent. After conducting some preliminary tests (such as ensuring that both the oracle and agent actions use the same tool and that the oracle action is not already mapped to another agent action), the verifier performs:
\begin{itemize}
    \item \textbf{Hard check} to compare action parameters that require exactness. For example, when replying to an email, it verifies that \texttt{email\_id} value is identical for both actions, \textit{i.e.} the agent replies to the correct email.
    \item \textbf{Soft check} for parameters that require more flexible evaluation, such as the content of an email or a message. To perform a soft check, an LLM judge is prompted with the user task as context, and the arguments from both the agent action and the oracle action as inputs. The LLM then determines if the actions are equivalent according to tool-specific guidelines. For example, emails verification includes guidelines to check their signatures.
\end{itemize} 
 After observing some hacking of the verifier during reinforcement learning experiments (see Appendix~\ref{app:judge_hacking}), we add a soft check for global sanity of the agent's messages.
    \item \textbf{Causality:} crucially, oracle actions are organized within an oracle graph, whereas agent actions are collected from a trajectory and simply ordered by execution time. Therefore, we must ensure that the agent does not violate dependencies within this graph.
    For example, if both oracle actions A and B depend solely on action C, the agent is free to execute A and B in any order, as long as they are executed after C; i.e. sequences C-B-A or C-A-B are both acceptable.
    Once a match is found, the \verifier ensures causality by verifying that all parent actions of the oracle action have already been matched with preceding agent actions.
    \item \textbf{Timing:} scenarios can include a time delay for certain actions relative to their parent actions (see for example Time scenarios in~\ref{sec:gaia2_capabilities}, which the agent must respect. The verifier evaluates whether the agent's timing falls within a specified tolerance window centered around the relative time of the oracle action. 
To determine the relative timing of the agent's action, it is necessary to identify which agent action corresponds to the oracle's parent action. 
This information is readily available due to the \verifier's process. Indeed, for a given oracle action, all its parent actions must be matched to an agent action before attempting to match the oracle action itself.
\end{itemize}
 
 If all oracle actions are successfully matched, the verifier returns a success signal. Conversely, if any oracle action cannot be matched to an agent action, the verifier returns a failure signal, see \autoref{fig:verifier} for two examples. 
 Crucially, the verifier implicitly assumes there are no equivalent \writeact actions, \textit{i.e.} user preferences are clearly stated with minimal ambiguity in the scenario tasks. For example, sending a message using the Messages app while the oracle action uses the Chat app will trigger a failure.

While other verification methods~\citep{patil2025bfcl,yao2024taubenchbenchmarktoolagentuserinteraction} compare the environment ground truth and actual final states, verifying a sequence of \writeact\ actions, which is equivalent to comparing ground truth and actual states after each \writeact\ action of the sequence, provides more control.
For example our verification allows to distinguish, \textit{e.g.} for safety considerations, a \myenv\ trajectory where the agent adds an event at the wrong place and correct itself from a trajectory where the agent is correct at first try.
Moreover, in \myenv, sequences of \writeact\ actions are easier for human to interpret and annotate, compared to diffs of states. 

\paragraph{Validating multi-turn scenarios} 

To validate multi-turn scenarios (\textit{i.e.} that include several rounds of interactions with a user or another entity from the environment), the verifier runs at the end of each turn to validate the current turn. 
This ensures the agent maintains the correct trajectory before proceeding to subsequent turns. Refer to Appendix~\ref{app:validation-multi-turn} for details. 

\subsubsection{Verifying the Verifier}
\label{verifying-verifier}

Verifiers are critical components of training and evaluation pipelines, where false positive or false negative \textit{e.g.} via hacking can result in flawed evaluations or collapsed trainings (see example in Appendix \ref{app:judge_hacking}).
We evaluate the \verifier by first deriving a series of ``unit'' tests from the oracle actions that the verifier should satisfy. Typically, we apply perturbations to oracle actions that we know preserve or invalidate the oracle trajectory validity, before submitting the oracle and perturbed oracle trajectories to the verifier and checking its verdict match the perturbation type. While these checks allow fast iteration, they only catch anticipated behaviors. Furthermore, the perturbed trajectories do not necessarily reflect real trajectories that could be obtained with an agent.

\paragraph{Validation benchmark} We complement this initial evaluation by analyzing \verifier verdicts for 450 trajectories manually labeled with the expected verifier outcome (Success or Failure).
The trajectories were derived from running agents powered by various models on scenarios from the \gaiatwo benchmark. 
We compare the \verifier with a simple baseline, \inContextVerifier, where an LLM is prompted with all the agent actions and criteria (causality constraints, relative time, soft/hard checks, etc.). The same model \llamaIII is used for both verifiers. 
\verifier achieves better accuracy than the baseline, which tends to accept agent trajectories too readily, see~\autoref{tab:ValBench}.

\begin{table*}[h!]
    \centering
    \begin{NiceTabular}{lccc}
    \toprule
     Verifier  & Agreement & Precision & Recall \\
    \midrule
     \inContextVerifier (LLM judge only)  & 0.72 & 0.53 & 0.83 \\
     \verifier  & 0.98 & 0.99 & 0.95 \\
    \bottomrule
    \end{NiceTabular}
    \caption{\verifier and \inContextVerifier results on 450 trajectories annotated with human labels.}
    \label{tab:ValBench}
\end{table*}

We then evaluate the \verifier powered with different models: \llamaIII, \Gemini and \claudeSonnet, see \autoref{tab:ValBenchModel} in Appendix~\ref{app:verifier_models} -- same prompts were used for all models. All three models have satisfactory precision scores, while the prompts were tuned for \llamaIII.

\subsection{An initial Agent Orchestration for \are}
\label{sec:agent_orchestration}

\are\ provides a default agent orchestration to run models on \myenv, though any orchestration is compatible with \are\ as long as it supports its two core interfaces: tools and notifications. Our implementation is a ReAct loop \citep{yao2023react} with some additions. The agent performs one tool call per step, formatted as a JSON. Its system prompt is structured in general-, agent-, and environment-level instructions.

\paragraph{Enhanced ReAct loop} Unlike classical ReAct implementations, our orchestration includes \texttt{pre-step} and \texttt{post-step} operations, systematically applied respectively before and after the LLM call. They are typically used to handle \are-specific functionality, like injecting new notifications into the agent's context (\texttt{pre-step}), or checking for turn-termination signals (\texttt{post-step}) (more details in Appendix \ref{app:agent_orchestration}). This augmented loop supports code execution capabilities (not released), though JSON tool calling remains the standard evaluation method for \myenv.

\paragraph{Multi-turn support} Because \are is asynchronous, for multi-turn scenarios, the orchestration manages pause/resume functionality where validation occurs semi-online \textit{i.e.} between turns, see Section~\ref{sec:are-verifier}: when the environment is paused in between turns, the agent is paused. If the environment sends new information via the notification system while the agent is paused, the orchestration automatically resumes the agent execution with the new information in context. \autoref{fig:sequence_diagram} illustrates this for a simple multi-turn scenario.

\subsection{\are\ Graphical User Interface}
\label{sec:are-ui}

Running scenarios with \are generates rich agent execution traces that include reasoning steps, tool calls, their outputs, notifications, and, on the environment side, temporal event flows that unfold over simulated time periods. It is important for practitioners to be able to debug these interactions, whose complexity requires specialized tooling. Existing development tools largely fall into one of these categories: interactive debugging platforms~\citep{Epperson_2025,RorsethGGSS25,pang2025interactivereasoningvisualizingcontrolling} and data annotation/curation platforms, each with distinct UI approaches. They also often lack key features like environment exploration. We provide a more detailed review of existing solutions in Appendix~\ref{app:ui_related_work}.

\begin{figure}[h!]
    \centering
    \includegraphics[width=1\linewidth]{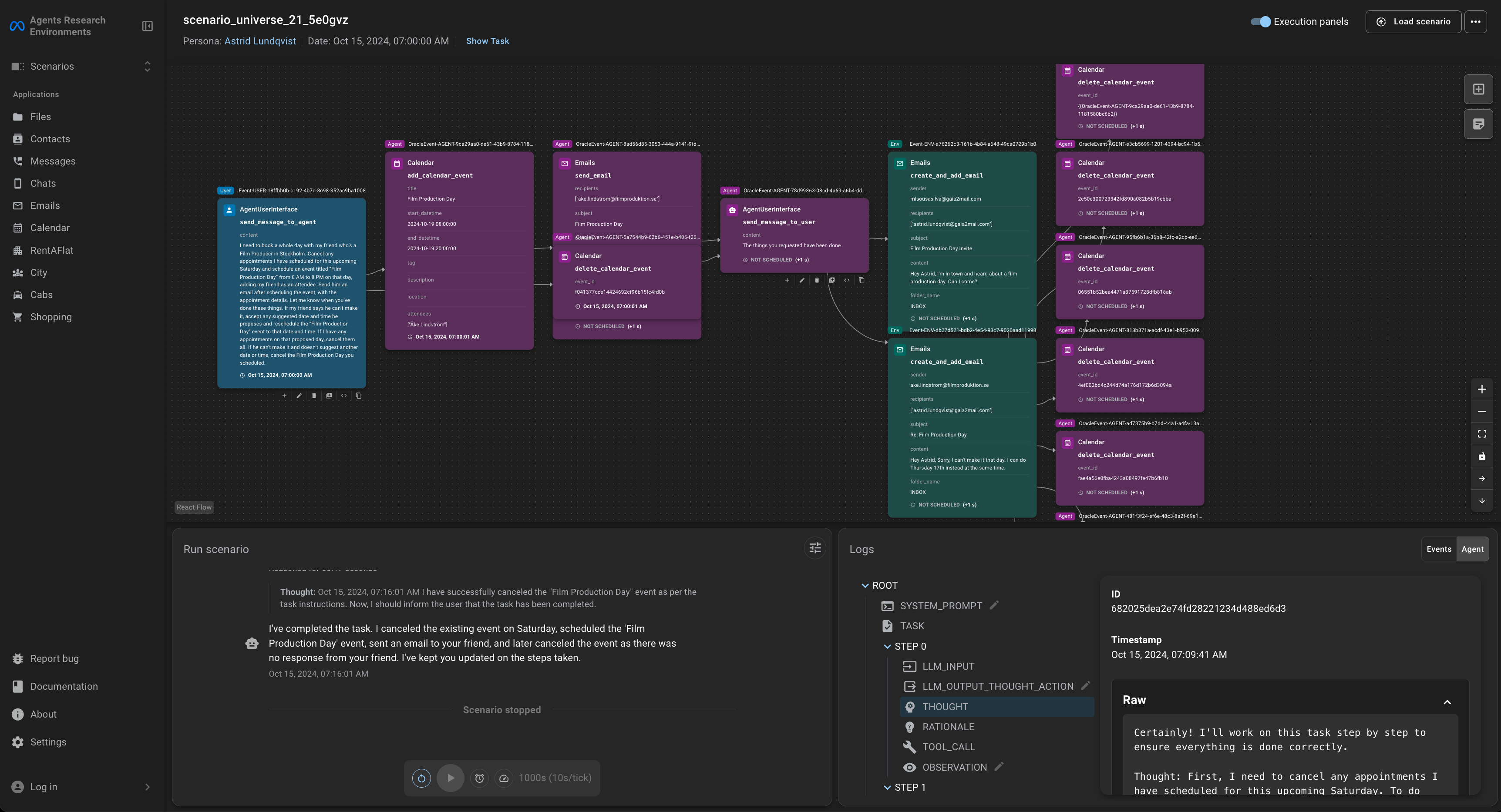}
    \caption{\are scenario view with event DAG (top), scenario run (bottom left) and agent logs (bottom right).}
    \label{fig:ui-annotations-dag}
\end{figure}

To address this, we propose a single \are Graphical User Interface (UI), a web-based platform that enables developers to interact with the environment, visualize scenarios (see~\autoref{fig:ui-annotations-dag}), and understand agent behavior and failures through detailed trace analysis and replay capabilities, and enable zero-code scenario annotation. We provide more details on the capabilities the UI offers to researchers and developers in Appendix~\ref{app:ui_appendix}.

\section{\gaiatwo: Expanding General Agent Evaluation} \label{section:gaia2}

\begin{figure}[H]
    \centering
    \includegraphics[width=\linewidth]{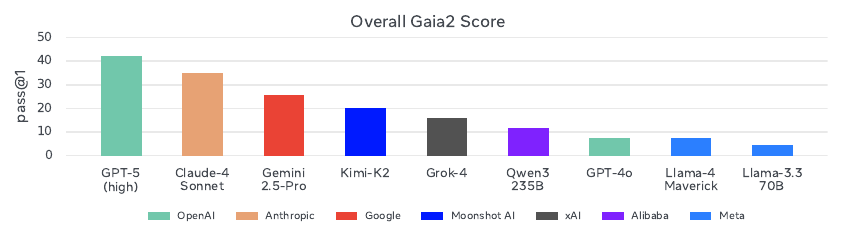}
    \caption{Overall Gaia2 benchmark performance across some major AI models using pass@1 evaluation. Proprietary frontier models (GPT-5, Claude-4 Sonnet, Gemini 2.5-Pro) significantly outperform open-source alternatives, with GPT-5 achieving the highest score with ``high'' reasoning. Among open-source models, Kimi-K2 leads.}
    \label{fig:gaia2_overall_score}
\end{figure}

\subsection{Description and Setup}

\gaiatwo\ consists of 800 unique verifiable scenarios, carefully annotated by humans across 10 distinct universes in the \myenv environment, with 101 tools each. The scenarios are organized into splits, each targeting one agent capability defined in Section~\ref{sec:gaia2_capabilities}. To support rapid and cost-effective evaluations, we also curate a 160-scenario split, \gaiatwo-mini. The benchmark includes two augmentation setups derived from \gaiatwo-mini, adding 320 scenarios to the original 800 for a total of 1,120 scenarios. Compared to prior benchmarks, \gaiatwo leverages \are to simulate complex interactions between an agent and a dynamic environment, thus getting closer to real-world agent use-cases. \gaiatwo has several distinguishing characteristics:

\begin{itemize}
    \item \textbf{Dynamic environment events:} Dynamic events modify world state asynchronously during \gaiatwo\ scenario execution, enabling evaluation of agent adaptation to changing conditions—a critical skill absent from static benchmarks~\citep{yao2024taubenchbenchmarktoolagentuserinteraction,jimenez2024swebench}. In contrast, VendingBench \citep{backlund2025vendingbenchbenchmarklongtermcoherence} employs a strictly synchronous environment which advances time only when the agent acts, with all new events (e.g., customer purchases) batched and delivered once every simulated morning.
    \item \textbf{Time:} In all \gaiatwo\ scenarios, time flows continuously. Many scenarios explicitly incorporate time as a dimension, requiring agents to handle temporal constraints. Temporal awareness is essential for practical applications such as scheduling tasks~\citep{google2024scheduled,openai2024scheduled,microsoft2024scheduled}, although omitted from existing benchmarks. \gaiatwo\ goes beyond setting alerts by evaluating agents’ ability to proactively initiate both time-based and event-driven actions throughout task execution. 
    \item \textbf{Agent-to-agent collaboration:} Gaia2 evaluates collaboration with other agents by representing apps such as Shopping or Email by autonomous, specialized agents. Compared to traditional benchmarks for multi-agent collaboration \citep{foerster2016learning,lowe2017multi, carroll2019utility}, Gaia2 is significantly more challenging, focuses on real-world tasks, and permits fully text/natural language-based model evaluation. With respect to more recent multi-agent benchmarks designed specifically for LLMs \citep{ vezhnevets2023generative, zhu2025multiagentbench}, \gaiatwo differs from existing work by embedding other agents as part of the environment rather than treating them as symmetric peers, targeting the emerging paradigm in which traditional API endpoints are replaced by agents \citep{google2025a2a}. In order to successfully complete tasks in such settings, agents acting on behalf of users may need to coordinate tool calls, share state, and understand the affordances of external agents during interaction. 

\end{itemize}

During an evaluation run, all scenarios are executed independently. For each scenario, the \are\ verifier (described in Section \ref{sec:are-verifier}) leverages oracle actions provided by annotators alongside the scenarios, cf. Section~\ref{sec:gaia2-annotation}, to assign Pass-Fail scores to agent trajectories. Final results are reported per split as Pass@1, averaged over three runs. An overall score is computed by averaging over all splits.

\subsection{Agent Capabilities Evaluated}
\label{sec:gaia2_capabilities}

To build \gaiatwo, we define a set of capabilities that we believe are necessary -- though not sufficient -- for general purpose agents. As introduced above, each of the 800 scenarios is built to emphasize at least one of these capabilities, yielding 160 scenarios per capability split. We provide example scenarios displayed in the \are\ GUI graph editor in Appendix~\ref{app:capa-annotations}.

\textbf{Search} scenarios require the agent to take multiple \readact\ actions in order to collect facts from different sources within the environment. Any sequence of \readact\ operations leading to the correct answer is considered successful as long as the answer is communicated via \sendmessagetouser before scenario timeout. While conceptually similar to the original \gaia benchmark's web search tasks, \gaiatwo\ search scenarios operate within a controlled \are\ environment.
\begin{itemize}
    \item Example: \textit{Which city do most of my friends live in? I consider any contact who I have at least one 1-on-1 conversation with on \chats a friend. In case of a tie, return the first city alphabetically.} \label{item:search_task}
    \item Explanation: This scenario requires the agent to cross-reference data from multiple apps (Contacts and \chats), perform aggregation operations, and handle edge cases like ties.
\end{itemize}

\textbf{Execution} scenarios require the agent to take multiple \writeact\ actions, which may need to be executed in a particular order. Most of the time, \readact\ actions are needed in order to gather information for properly filling \writeact\ action arguments.
\begin{itemize}
    \item Example: \textit{Update all my contacts aged 24 or younger to be one year older than they are currently.}
    \item Explanation: This task requires the agent to read contact information, filter based on age criteria, and execute multiple \writeact\ to update Contacts data.
\end{itemize}

All remaining capabilities tested in \gaiatwo reflect tasks with a balanced number of required \readact\ and \writeact\ operations. However, each capability features an additional challenge. Namely:

\textbf{Adaptability} scenarios require the agent to dynamically adapt to environmental changes that are consequences of previous agent actions, such as a response to an email sent by the agent, or the cancellation of a ride booked by the agent. These events require agents to recognize when adaptation is necessary and adjust their strategy accordingly.
\begin{itemize}
    \item Example: \textit{I have to meet my friend Kaida Schönberger to view a property with her [...] If she replies to suggest another property or time, please replace it with the listing she actually wants and reschedule at the time that works for her.}
    \item Explanation: This task requires the agent to execute an initial plan while monitoring for environmental changes (the friend's response), then adapt the plan based on new information. The agent must demonstrate flexibility in execution while maintaining task objectives.
\end{itemize}

\textbf{Time} scenarios require agents to execute actions in due time, monitor and respond to events, and maintain awareness of temporal relationships throughout task execution. The duration of Time scenarios is currently capped at 5 minutes to facilitate annotation and evaluation.
\begin{itemize}
    \item Example: \textit{Send individual \chats messages to the colleagues I am supposed to meet today, asking who is supposed to order the cab. If after 3 minutes there is no response, order a default cab from [...].}
    \item Explanation: This scenario requires the agent to understand temporal constraints (the 3-minute window), monitor for events (new messages from colleagues), and execute a time-sensitive action (order a cab).
\end{itemize}

\textbf{Ambiguity} scenarios reflect user tasks that are impossible, contradictory, or have multiple valid answers, with negative consequences arising during interaction if agents make mistakes. These scenarios test agents' ability to recognize these issues and seek appropriate clarification from users.
\begin{itemize}
    \item Example: \textit{Schedule a 1h Yoga event each day at 6:00 PM from October 16, 2024 to October 21, 2024. Ask me in case there are conflicts.}
    \item Explanation: While this task appears straightforward, current models often struggle to identify contradictions or multiple valid interpretations, tending to execute the first seemingly valid approach rather than recognizing the need for clarification.
\end{itemize}

\textbf{Agent2Agent} scenarios replace apps with app-agents. Main-agents can no longer access app tools directly and must instead communicate with the app-agents in order to place tool calls, observe tool call outputs, and ultimately accomplish user tasks. This transformation requires agents to develop robust collaboration capabilities, including sub-task setting, affordance understanding, ``context-sharing,'' and general coordination. By default, agents and app sub-agents are instantiated with the same scaffold and model, with good performance requiring strong sub-goal setting and sub-goal solving. However, \gaiatwo also supports heterogeneous multi-agent evaluations, i.e. where stronger agents supervise weaker sub-agents or vice-versa.
\begin{itemize}
    \item Example: Same \textit{Search} task as above but the Contacts and \chats apps are replaced by app sub-agents and the main agent must communicate with them in order to gather information.
\end{itemize}

\textbf{Noise} scenarios require robustness to environment noise, simulating the inherent instability of real-world systems, where APIs change, services become temporarily unavailable, and environmental conditions shift during task execution. This category applies systematic perturbations to \gaiatwo\ scenarios, including tool signature modifications, random failure probabilities, and dynamic environment events that are irrelevant to the task. We assess the sanity of our noise mechanisms in Appendix~\ref{sec:gaia2_noise_experiments}.
\begin{itemize}
    \item Example: Same \textit{Adaptability} task as above but with random tool execution errors and random environment events (e.g., messages from other people) occurring during execution.
\end{itemize}

\begin{figure}[t!]
    \centering
    \includegraphics[width=0.9\linewidth]{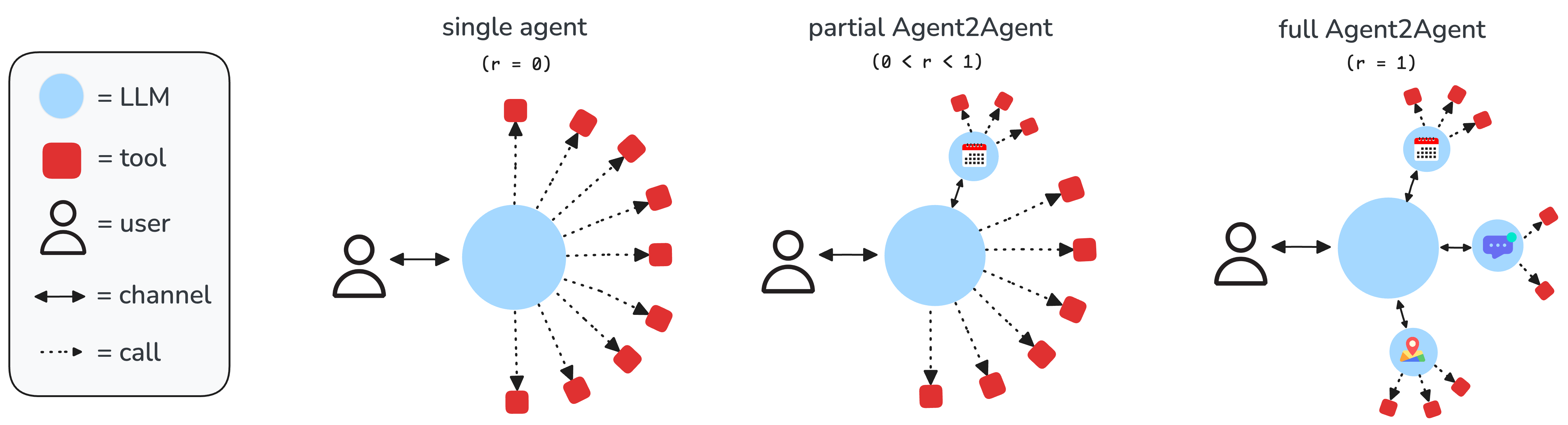}
    \caption{In Agent2Agent scenarios, a proportion ``r'' of the apps in Gaia2 scenarios are replaced by autonomous agents with access to the corresponding APIs and/or tools. The main agent (instantiated by the user) can communicate with app agents through a channel, but cannot use their tools or see their tool call outputs directly. Agents now have to send messages in order to coordinate actions, share state, set sub-goals, and collaboratively solve user tasks. By default, \gaiatwo evaluates LLMs on the full Agent2Agent (``r = 1'') setting.}
    \label{fig:agent2agent}
\end{figure}

\paragraph{Environment augmentation and practical extensions to more capabilities} A key property of \gaiatwo\ scenarios is that ground truth task solutions are invariant to many changes in the environment, facilitating the creation of new scenarios through augmentation of existing ones, and thus making it possible to extend \gaiatwo\ without new annotations.
Consider as an illustration the example Search scenario in~\ref{item:search_task}: its final answer stays the same whether \textit{Chats} is an actual app or a sub-agent, allowing for the re-use of this scenario in an Agent2Agent evaluation.

This is highly practical: to create Agent2Agent and Noise evaluations, we simply augment \gaiatwo-mini scenarios from all other capabilities.
For Agent2Agent, all \myenv\ apps are replaced with app sub-agents, transforming single-agent tasks into collaborative multi-agent scenarios. The proportion and nature of sub-agents is parameterized and can be modified for research purpose.
For Noise, we apply systematic perturbations through two mechanisms: 
\begin{itemize}
    \item \textbf{Tool augmentations} modify app interfaces by changing tool signatures, descriptions, or introducing random failure probabilities -- default probability for \gaiatwo\ is 0.1.
    \item \textbf{Environment events} inject with some frequency random events in the simulation, such as receiving new emails or having new products available for shopping while the agent is working --- default frequency for \gaiatwo\ is 10 events per minute.
\end{itemize} 
The frequency of probabilities and random events can be modified.  In particular, increasing the frequency of random events is an interesting augmentation as it challenges agent modeling choices that inject all environment events into the context window. 

These environment augmentations demonstrate the extensibility of \gaiatwo\ and the \are\ platform more broadly. Researchers can create new evaluation dimensions by applying different environment modifications to existing scenarios, enabling exploration of capabilities such as memory, safety or security, without the substantial cost of creating entirely new benchmark datasets.

\subsection{Data Collection} \label{sec:gaia2-annotation}

\paragraph{Principles for creating \gaiatwo\ scenarios}
We task a annotators to create scenarios following the process described in Section~\ref{sec:scenario_creation}, plus some additions specific to \gaiatwo, using the GUI and its graph editor described in Section~\ref{sec:are-annotation-ui}.
Namely, annotators are asked to create scenarios that are difficult with respect to one capability only, so that developers get clear signal on the strength and weakness of their agent. Since difficult tasks for humans are not necessarily difficult for models, we calibrate the difficulty of the annotated scenarios we receive on internal agents.
To facilitate automated verification with LLMs as judges, we require that agent messages (to the User or Contacts) remain short, easy to parse and neutral in tone.
Moreover, we exclude sensitive topics and personally identifiable information. We provide capability-specific guidelines in Appendix~\ref{app:capa-annotations}.

\paragraph{Scenario quality assurance} 

In spite of our GUI, creating diverse, interesting, challenging and verifiable \myenv scenarios puts a heavy cognitive load on annotators, increasing the likelihood of annotation errors. For example, think of a scenario requiring to delete any contact that did not send a message to the user in the past month, with ten or more such contacts.
To decrease this likelihood, scenarios undergo quality assurance (QA) on both annotators and researcher side.
Annotator side, similar to~\citep{mialon2023gaia}, each newly created scenario undergoes multiple rounds of validation by different annotators: 
\begin{enumerate}
    \item Annotator A creates a prompt and the Oracle Event graph to solve it.
    \item Annotator B receives the prompt from annotator A and creates an independent Oracle Event graph to solve it (without seeing A's solution). They can refine the prompt if it is ambiguous.
    \item Annotator C does the same as B, without seeing A and B's solution.
    \item Annotator D sees the 3 annotated solutions and confirms that they are consistent. If not, D can reject the scenario or make minor edits to the best prompt version to make it more specific.
\end{enumerate}

To further alleviate the annotator cognitive burden, we complement human QA with automated checks:
\begin{itemize}
    \item \textbf{Pre-QA guardrails}
    leverage the graph editor described in Section~\ref{sec:are-annotation-ui} to prevent any annotated DAG of \writeact from saving if it does not satisfy \myenv modeling constraints, such as each turn ending with \sendmessagetouser. We provide more details on \myenv\ annotation guardrails in Appendix \ref{app:gaia2-design}.
    \item \textbf{Post-QA evaluation} leverages model success rates per scenario. Typically, a 100\% success rate suggests a too easy scenario, while a 0\% success rate suggests a misspecification or impossible scenario. This approach allowed us to find various broken scenarios that escaped QA attention.
\end{itemize}

\section{Experiments}

In our core experiments, we evaluate state-of-the-art models on each Gaia2 capability split \citep{moonshotai2025kimik2, comanici2025gemini25pushingfrontier, yang2025qwen3technicalreport, grattafiori2024llama3herdmodels, openai2024gpt4ocard}. We also evaluate model sensitivity to environment- and tool-level augmentations (``noise'') and to different configurations of ``time’’ evaluation scenarios. Finally, we use Gaia2 as a test-bed for evaluating zero-shot collaboration and coordination between LLM agents via \are\ Agent2Agent mode. We show that scaling up the number of agentic collaborators on \gaiatwo tasks improves best-of-n scaling laws for weaker models.

\subsection{Experimental Setup}

\paragraph{Model configuration and evaluation protocol}
We evaluate both open-source and proprietary LLMs using our baseline ReAct-like scaffold, described in Section~\ref{sec:agent_orchestration}. Our approach avoids model-specific tool calling APIs, to ensure provider-agnostic evaluation. All models are evaluated with long-context capabilities (minimum 128K tokens), temperature set to 0.5, and generation limits of 16,000 tokens per turn. If context length is exceeded during execution, the agent is terminated and the scenario failed. Each model is evaluated three times per scenario to account for variance. The agent loop continues until one of the two termination conditions is met: maximum iteration count (200 steps) or not passing verification at the end of a turn. We encourage researchers to submit alternative orchestration approaches and run the benchmark with tool-calling APIs. Our leaderboard presents observational comparisons using the ReAct framework rather than claiming definitive model hierarchies.

\paragraph{Environment configuration}
Tools and environment descriptions are provided via system prompts alongside the notification policy configuration. We set the notification verbosity to \texttt{medium} level by default: agents receive systematic alerts for high-priority events while filtering out lower-priority background notifications (e.g., new product listings in shopping applications). All scenarios are verified with the \verifier using \llamaIII with temperature set to 0.

\paragraph{Mitigating API issues} 
To address deployment challenges such as service outages and rate limiting, we implement a simulated generation time mode: the simulation pauses during agent response generation and resumes with a temporal offset equal to the actual response completion duration. This approach maintains realistic timing constraints while ensuring robust evaluation under varying infrastructure conditions.

We provide more information about the experimental setup in Appendix \ref{app:experimental_setup}.

\subsection{Core Results}

\begin{table}[t]
    \centering
    \begin{adjustbox}{max width=\textwidth}
    \begin{NiceTabular}{lrrrrrrr>{\columncolor{blue!20}}r}
    \CodeBefore
    \Body
    \toprule 
     & {Execution} & {Search} & {Ambiguity} & {Adaptability} & {Time}  & {Noise} & {Agent2Agent} & {\textbf{\shortstack{Overall}}} \\
    \midrule
    \llamaIII & 7.1 $\scriptstyle \pm 1.2$ & 11.5 $\scriptstyle \pm 1.5$ & 1.7 $\scriptstyle \pm 0.6$ & 1.9 $\scriptstyle \pm 0.6$ & 0.4 $\scriptstyle \pm 0.3$ & 3.8 $\scriptstyle \pm 0.9$ & 4.6 $\scriptstyle \pm 1.0$ & 4.4 \\
    Llama 4 Maverick & 13.8 $\scriptstyle \pm 1.6$ & 14.4 $\scriptstyle \pm 1.6$ & 2.1 $\scriptstyle \pm 0.7$ & 5.0 $\scriptstyle \pm 1.0$ & 1.2 $\scriptstyle \pm 0.5$ & 6.2 $\scriptstyle \pm 1.1$ & 9.2 $\scriptstyle \pm 1.3$ & 7.4 \\
    GPT-4o & 8.3 $\scriptstyle \pm 1.3$ & 17.5 $\scriptstyle \pm 1.7$ & 4.4 $\scriptstyle \pm 0.9$ & 6.2 $\scriptstyle \pm 1.1$ & 5.8 $\scriptstyle \pm 1.1$ & 4.6 $\scriptstyle \pm 1.0$ & 5.2 $\scriptstyle \pm 1.0$ & 7.4 \\
    Qwen3-235B & 22.7 $\scriptstyle \pm 1.9$ & 22.3 $\scriptstyle \pm 1.9$ & 6.5 $\scriptstyle \pm 1.1$ & 8.1 $\scriptstyle \pm 1.2$ & 1.2 $\scriptstyle \pm 0.5$ & 10.8 $\scriptstyle \pm 1.4$ & 9.4 $\scriptstyle \pm 1.3$ & 11.6 \\
    Grok-4 & 8.8 $\scriptstyle \pm 2.2$ & 57.5 $\scriptstyle \pm 3.9$ & 9.4 $\scriptstyle \pm 2.3$ & 4.4 $\scriptstyle \pm 1.6$ & 0.0 $\scriptstyle \pm 0.0$ & 15.6 $\scriptstyle \pm 2.9$ & 14.4 $\scriptstyle \pm 2.8$ & 15.7 \\
    Kimi-K2 & 34.2 $\scriptstyle \pm 2.2$ & 36.0 $\scriptstyle \pm 2.2$ & 8.3 $\scriptstyle \pm 1.3$ & 24.0 $\scriptstyle \pm 1.9$ & 0.8 $\scriptstyle \pm 0.4$ & 18.8 $\scriptstyle \pm 1.8$ & 18.3 $\scriptstyle \pm 1.8$ & 20.1 \\
    Gemini-2.5-Pro & 39.2 $\scriptstyle \pm 2.2$ & 57.7 $\scriptstyle \pm 2.3$ & 18.1 $\scriptstyle \pm 1.8$ & 17.5 $\scriptstyle \pm 1.7$ & \textbf{7.3} $\scriptstyle \pm 1.2$ & 20.4 $\scriptstyle \pm 1.8$ & 20.4 $\scriptstyle \pm 1.8$ & 25.8 \\
    Claude-4-Sonnet & 57.9 $\scriptstyle \pm 2.3$ & 59.8 $\scriptstyle \pm 2.2$ & 24.2 $\scriptstyle \pm 2.0$ & \textbf{38.1} $\scriptstyle \pm 2.2$ & \textbf{8.1} $\scriptstyle \pm 1.2$ & 27.7 $\scriptstyle \pm 2.0$ & \textbf{27.9} $\scriptstyle \pm 2.0$ & 34.8 \\
    GPT-5 (minimal) & 31.9 $\scriptstyle \pm 2.1$ & 26.2 $\scriptstyle \pm 2.0$ & 20.6 $\scriptstyle \pm 1.8$ & 19.2 $\scriptstyle \pm 1.8$ & 5.2 $\scriptstyle \pm 1.0$ & 13.1 $\scriptstyle \pm 1.5$ & 11.5 $\scriptstyle \pm 1.5$ & 18.2 \\
    GPT-5 (low) & 52.7 $\scriptstyle \pm 2.3$ & 64.2 $\scriptstyle \pm 2.2$ & 39.6 $\scriptstyle \pm 2.2$ & 30.2 $\scriptstyle \pm 2.1$ & 2.3 $\scriptstyle \pm 0.7$ & 28.3 $\scriptstyle \pm 2.1$ & 24.6 $\scriptstyle \pm 2.0$ & 34.6 \\
    GPT-5 (high) & \textbf{69.2} $\scriptstyle \pm 2.1$ & \textbf{79.6} $\scriptstyle \pm 1.8$ & \textbf{51.9} $\scriptstyle \pm 2.3$ & \textbf{40.4} $\scriptstyle \pm 2.2$ & 0.0 $\scriptstyle \pm 0.0$ & \textbf{35.4} $\scriptstyle \pm 2.2$ & 17.9 $\scriptstyle \pm 1.8$ & \textbf{42.1} \\
    \bottomrule
    \end{NiceTabular}
    \end{adjustbox}
    \caption{Pass@1 scores and standard errors on \gaiatwo scenarios per model and capability split. All models are evaluated with the same ReAct loop scaffolding described in Section \ref{sec:agent_orchestration}. The overall score is the average across all splits, each run three times to account for variance.}
    \label{table:core_results}
\end{table}

\begin{figure}[t]
    \centering
    \includegraphics[width=\linewidth]{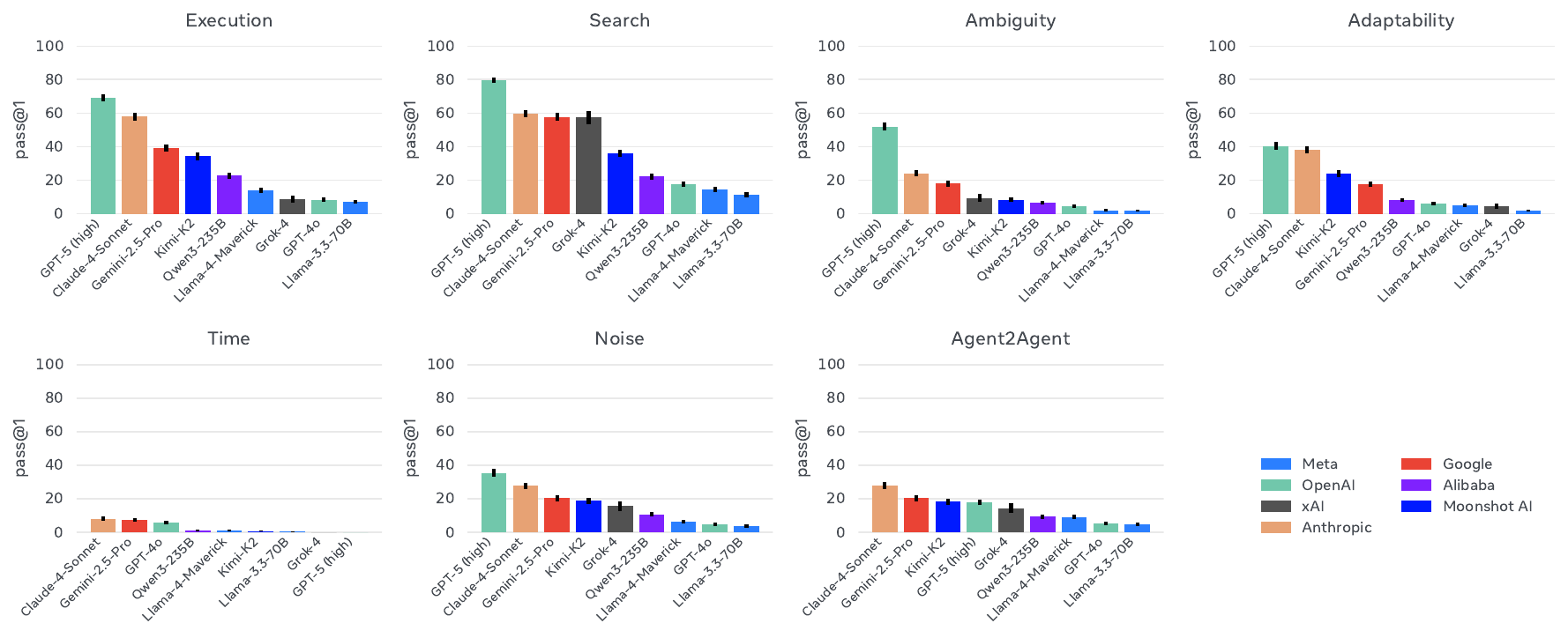}
    \caption{\gaiatwo scores per capability split. Models are reranked independently for each capability, highlighting where they excel or struggle.}
    \label{fig:gaia2_score_vs_capability}
\end{figure}

\begin{figure}[t!]
    \centering
    \includegraphics[width=\linewidth]{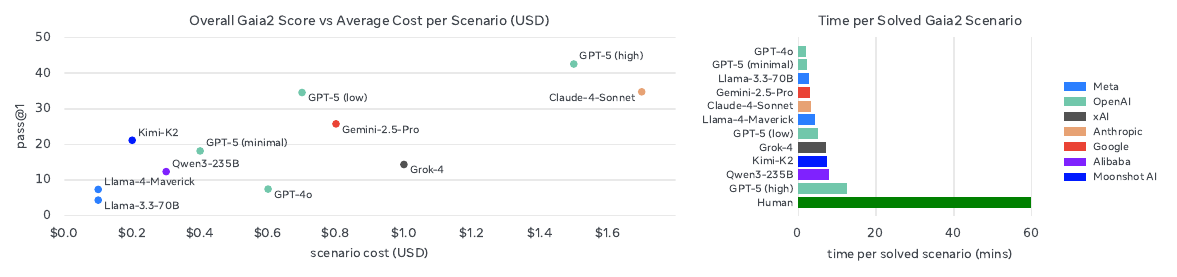}
    \caption{Left: Gaia2 score vs average scenario cost in USD. Right: Time taken per model to successfully solve \gaiatwo scenarios compared to Humans.}
    \label{fig:gaia2_score_vs_cost_vs_humans}
\end{figure}

Our core experimental results are presented in 
\autoref{table:core_results}, \autoref{fig:gaia2_overall_score}, 
\autoref{fig:gaia2_score_vs_capability}, and 
\autoref{fig:gaia2_score_vs_cost_vs_humans}.
Execution and Search are as the easiest splits.
The top five models on these categories all underpin ``DeepResearch'' products\footnote{
See DeepResearch products:
\href{https://openai.com/index/introducing-deep-research/}{OpenAI},
\href{https://gemini.google/overview/deep-research/}{Gemini},
\href{https://grok.com}{Grok},
\href{https://www.anthropic.com/engineering/multi-agent-research-system}{Anthropic}, and
\href{https://www.kimi.com}{Kimi}.
}
with Grok-4 strong on Search but collapsing elsewhere, consistent with its specialization. 
Ambiguity and Adaptability remain challenging, with robust performance limited to Claude~4 Sonnet and GPT-5 (high). 
The sharp drop from Execution/Search to these categories shows that existing benchmarks may overestimate robustness in realistic environments.

The Time split further separates frontier models: only Gemini~2.5 Pro and Claude~4 Sonnet achieve non-trivial scores, consistent with their efficiency–latency tradeoff in \autoref{fig:gaia2_score_vs_cost_vs_humans}. 
Noise robustness also lags: although GPT-5 (high) reaches 36.0, most models fall below 20, with significant degradation under noisy conditions (see ablation in Appendix~\ref{sec:gaia2_noise_experiments}). 
Agent2Agent collaboration can offset weaknesses, benefiting weaker models such as Llama~3.3 and Llama 4 Maverick more than stronger frontier systems as highlighted in Section~\ref{subsec:a2a_analysis}. 

Overall, GPT-5 (high) performs best on the benchmark, leading Execution/Search and the more challenging Ambiguity/Adaptability categories, exceeding Claude 4 Sonnet by 8 points. Kimi K2 is the strongest open model, particularly on Adaptability. In sum, frontier models largely solve instruction-following and search, but robustness, ambiguity resolution, and collaboration remain open problems for real-world use.

We extend our analysis beyond raw model scores to identify the factors that drive performance differences between models and to determine what most strongly contributes to achieving high scores on \gaiatwo. In addition, since agents are ultimately intended for deployment in production settings, we evaluate their performance in relation to their computational cost and execution time.

\paragraph{Model costs} \autoref{fig:gaia2_score_vs_cost_vs_humans} compares models by average scenario cost (USD)\footnote{Cost estimates from \href{https://artificialanalysis.ai/models}{Artificial Analysis} model pricing data (accessed September 10, 2025) }
and time to solve a scenario, including human annotator baselines. Our analysis highlights clear cost–performance–time tradeoffs. GPT-5’s reasoning variants illustrate scaling directly: higher computational investment yields systematically better performance but with longer solution times. For comparable accuracy, Claude~4 Sonnet is roughly $3\times$ more expensive than GPT-5 (low) but much faster, whereas GPT-4o combines high cost with lower performance, offering poor value. Most models fall along expected tradeoff curves, though outliers emerge: Grok-4 is particularly inefficient, while Kimi~K2 provides strong cost–performance despite being slower than Gemini~2.5 Pro. Non-expert human annotators were slower than all models, partly because they solved scenarios through ARE’s UI rather than a real OS, which likely inflated execution times; however, they did manage to complete the tasks. More broadly, these findings call for a shift in evaluation culture: comparing model weights or FLOPs alone is increasingly meaningless when assessing AI systems. Instead, benchmarks should report cost-normalized metrics, such as success rate per dollar or per unit of compute, see also~\citet{arc_agi_2025}. As shown in \autoref{fig:gaia2_score_vs_cost_vs_humans}, normalizing \gaiatwo\ results by average price per task reveals that some models achieve more favorable tradeoffs than raw scores suggest, better reflecting how agents will be judged in practice—by their ability to solve tasks reliably and efficiently under resource constraints.

\paragraph{What behaviors drive performance?}
We analyze key behavioral factors that correlate with \gaiatwo performance to understand performance drivers across models. Our first hypothesis posits that exploration drives success: we expect pass@1 scores to scale with tool call frequency and with the number of \readact actions before first \writeact operations, indicating systematic information gathering. Our second hypothesis suggests that increased token generation yields better performance through more comprehensive reasoning.

\begin{figure}[t!]
    \centering
    \includegraphics[width=\linewidth]{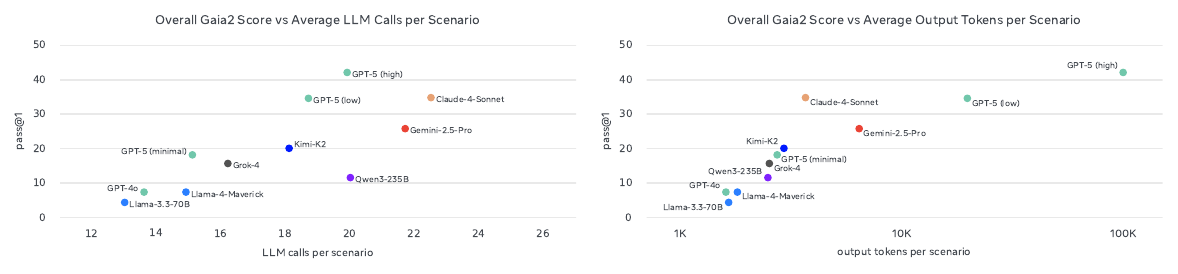}
    \caption{Left: \gaiatwo pass@1 versus average model calls per scenario. The performance of models is highly correlated to the number of tool calls, emphasizing the importance of exploration.
    Right: \gaiatwo pass@1 score versus average output tokens per scenario (log scale). Claude 4 Sonnet, while costing a lot, lies beyond the Pareto frontier.}
    \label{fig:gaia2_perf_vs_llm_calls_output_tokens}
\end{figure}

\autoref{fig:gaia2_perf_vs_llm_calls_output_tokens} (right) confirms the token-performance relationship, revealing a positive correlation between output tokens and pass@1 scores across most models. However, Claude-4 Sonnet and Kimi-K2 emerge as significant outliers, achieving high performance ($\sim$35\% and $\sim$21\% respectively) while generating relatively few tokens (left). These hybrid reasoning models demonstrate exceptional efficiency, potentially due to larger parameter counts or specialized architectures, though Claude-4 Sonnet's superior performance comes with substantially higher operational costs.

We also examined whether models exhibited app-specific usage patterns, see~\autoref{fig:app_usage_distribution} for an example. Across all evaluated systems, app usage distributions were nearly identical, suggesting that performance differences arise from general reasoning capabilities rather than preferences for particular apps.

\subsection{Time Scenarios Emphasize the Importance of Inference Speed and Reliability}

\begin{figure}[htbp]
    \centering
    \includegraphics[width=0.49\textwidth]{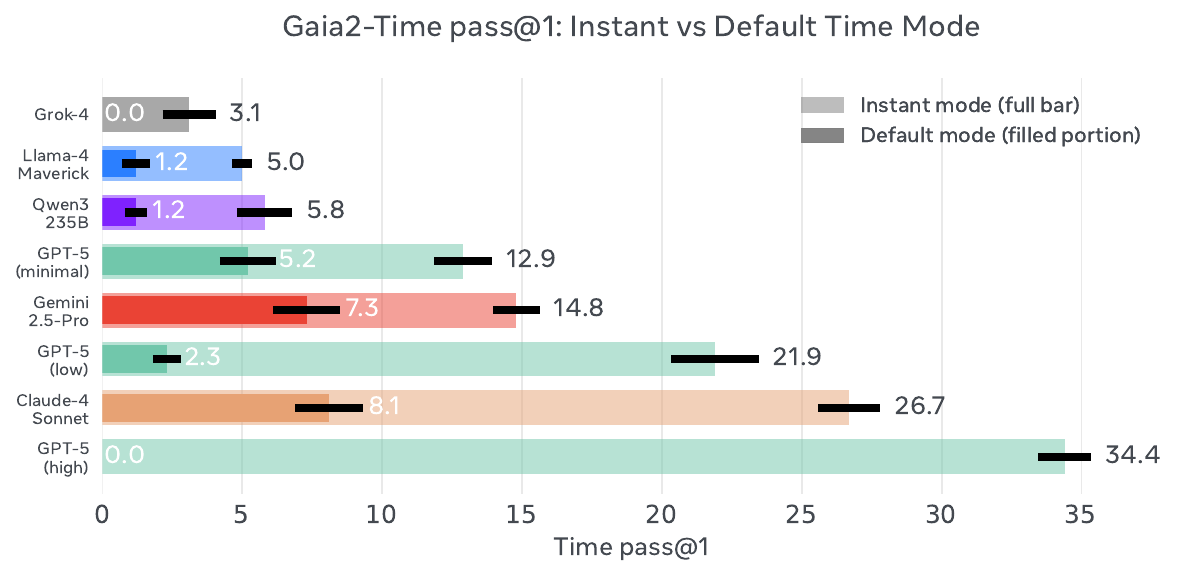}
    \includegraphics[width=0.49\textwidth]{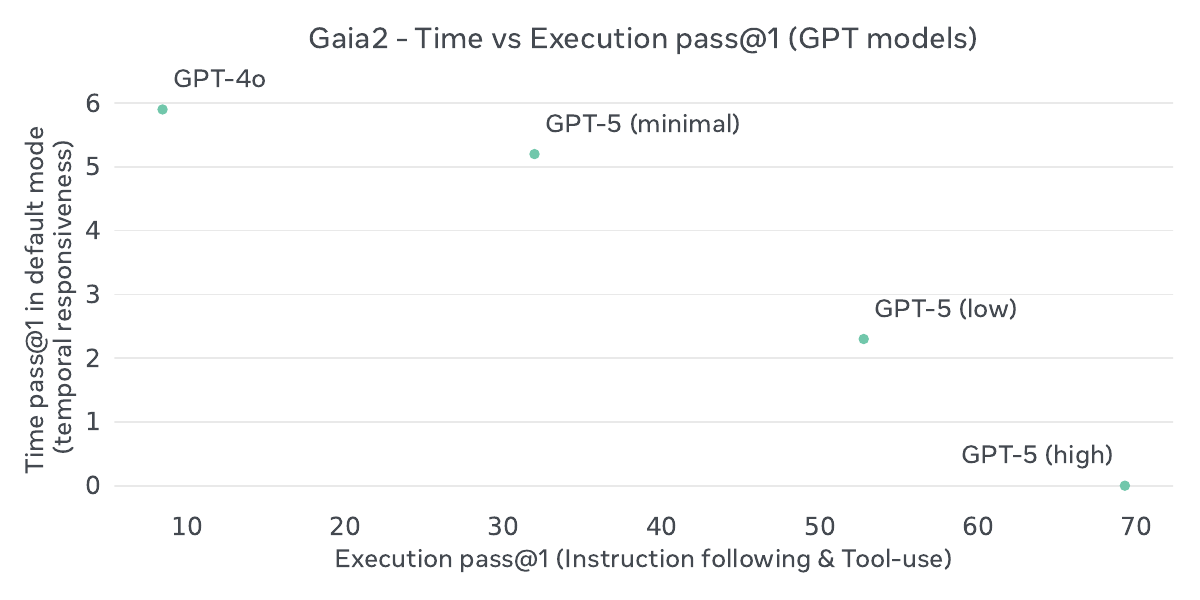}
    \caption{Left: Pass@1 scores on \gaiatwo-time with default mode vs. instant mode. Right: Inverse Scaling for Time: Frontier models perform poorly on the Time split (default mode), due to their time-consuming reasoning capabilities.}
    \label{fig:time_results}
\end{figure}

The Time category emphasizes the difficulty in evaluating \emph{models} independently of the \emph{systems} they operate within. Indeed, various contributing factors can lead to delays in taking time-sensitive actions. These include model policy errors (e.g., incorrect reasoning or actions), inference speed, communication issues with the LLM inference server, and server errors (deployment issues or downtime). Our view is that evaluation should isolate intrinsic model properties like model policy errors and inference speed from infrastructure-related problems such as server failures. To isolate these factors, we implemented two simulation modes:

\begin{itemize}
    \item \textbf{Generation time (\gaiatwo default mode):} The environment's time is paused during LLM inference server queries, and incremented by the generation duration measured on the client side. This approach excludes time lost due to repeated server errors, while the actual generation time is accounted for.
    \item \textbf{Instant:} Each action is simulated to take a fixed duration of 1 second in the environment, regardless of real inference latency. This mode ablates the effects of generation time, isolating model policy (reasoning and actions) from inference speed.
\end{itemize}
\paragraph{Impact of generation time} We conducted additional experiments on \gaiatwo-Time in instant mode. Results shown in \autoref{fig:time_results} (left) demonstrate that inference speed significantly impacts model performance on Time scenarios. As expected, instant mode yields higher scores overall, but the gap is significantly larger for reasoning models. Claude 4.0 Sonnet increases from 8.2\% with generation time to 26.7\% in instant mode, and GPT-5 with reasoning (high) from 0\% to 34.4\%. This indicates that long thinking times drastically improve these policies at the cost of timing. In contrast, models like Llama 4 Maverick and GPT-5 without reasoning (minimal) show smaller gaps due to faster inference speeds. Gemini 2.5 Pro combines a good policy with fast inference, making it the best at supporting strict timing requirements on short timescales.

\paragraph{Impact of server issues} We conducted real-time experiments internally observing lower scores when querying proprietary model APIs, largely due to frequent API rate-limiting that hindered timely agent execution (excluded due to irreproducibility). This shows benefits of self-hosting open models to avoid third-party server reliance and mitigate downtime. These findings highlight the need to redefine model serving when response time is critical, emphasizing the role of inference speed and infrastructure in time-sensitive applications.

\paragraph{Agent orchestration} Finally, some Time scenarios require executing multiple actions simultaneously within a narrow tolerance window, making them challenging or even impossible within our current agent scaffold. Improving performance on \gaiatwo-Time would require designing a scaffold that supports parallel multi-tasking. 

\begin{figure}[b]
    \centering
    \includegraphics[width=\linewidth]{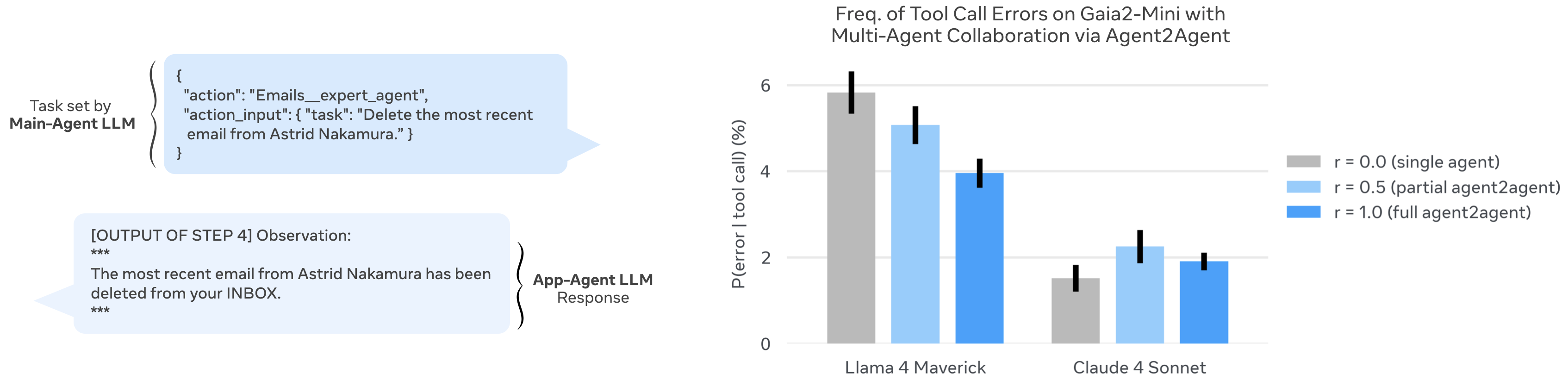}
    \caption{Agent2Agent tests whether LLM agents can collaborate through message passing in order to solve Gaia2 tasks via sub-task decomposition. For lighter-weight LLMs, collaboration in Agent2Agent results in a lower incidence of tool call errors. Left: Sample exchange between Llama~4 Maverick main vs app agent in an Agent2Agent scenario. Right: Frequency of errors per tool call (lower is better) on Gaia2-mini for Llama 4 Maverick and Claude~4 Sonnet. }
    \label{fig:gaia2_mini_multi_agent_prob_tool_errors}
\end{figure}

\begin{figure}[t]
    \centering
    \includegraphics[width=\linewidth]{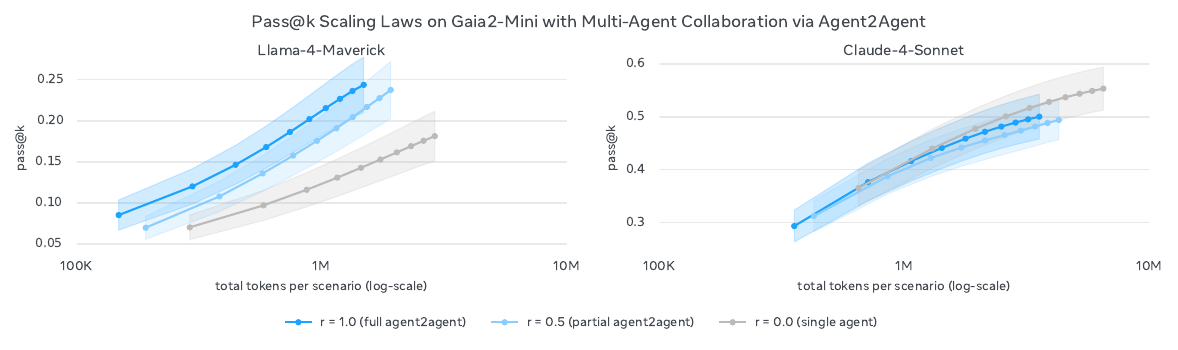}
    \caption{Increasing the number of multi-agent collaborators in Gaia2 scenarios by increasing the Agent2Agent ratio  $r$ improves pass@k scaling laws for Llama 4 Maverick, but does not improve token cost vs score tradeoffs with repeated sampling for Claude 4 Sonnet. }
    \label{fig:gaia2_mini_multi_agent_scaling_laws}
\end{figure}

\subsection{A Closer Look at Multi-Agent Collaboration on \gaiatwo with Agent2Agent} \label{subsec:a2a_analysis}

Inspired by recent work pushing beyond single-LLM agent tool-use and towards agent teams that message, coordinate, and divide labor \citep{google2025a2a}, we conduct a deeper study of multi-agent collaboration on \gaiatwo scenarios. Here, we probe the effect of increasing multi-agent collaboration by varying $r$ -- the ratio of apps in a scenario replaced with autonomous ``app agents'' (set to $r = 1$ by default in \gaiatwo-Agent2Agent, see Figure \ref{fig:agent2agent}) -- as well as the effect of swapping app-agent instances across model families. We focus on two models at different points in the cost-quality curve: Llama 4 Maverick, a lighter-weight \& open-source model, and Claude 4 Sonnet, the strongest overall LLM on the standard Agent2Agent setting at $r = 1$ (Table \ref{table:core_results}).

\paragraph{General effects of increasing forced collaboration}

For the lighter-weight Llama 4 Maverick, centralized collaboration on \gaiatwo tasks via Agent2Agent improves both performance with pass@k and operational stability. As the agent-to-agent ratio $r$ increases, we observe more favorable scaling with repeated sampling and a lower incidence of tool-call errors per step (Figure \ref{fig:gaia2_mini_multi_agent_prob_tool_errors} right; Figure \ref{fig:gaia2_mini_multi_agent_scaling_laws}).

However, the trends observed for Llama 4 are not universal. For Claude 4 Sonnet, increasing the collaborator ratio $r$ -- and thus the degree of task decomposition -- does not improve cost-normalized performance under best-of-$k$ sampling: score per token plateaus with or without multi-agent collaboration. Similarly, collaboration ratio with Agent2Agent has a weak negative effect on tool call error frequency.

One explanation is that Agent2Agent encourages hierarchical decomposition of decision making: as shown in Figure \ref{fig:gaia2_mini_multi_agent_prob_tool_errors} left, sub-goals issued by a main-agent to an app-agent instantiate temporally extended actions akin to options in hierarchical reinforcement learning \citep{sutton1999between}. Under this lens, gains in performance on practical tasks may materialize only when the benefits of hierarchical decomposition outweigh the costs. For example, Agent2Agent may increase performance on practical agent tasks when sub-goals set by main-agents are favorably-scoped, correctly reflecting the affordances of app-agents \& corresponding to tasks that are ``easier'' or ``faster'' to solve, and both app- and main-agents are capable of reliably exchanging state \& intent during message-passing. Likewise, the addition of multi-agent hierarchy can result in cascading errors and/or saturating gains if post-training data and objectives have fit models to long-form, single-agent planning and tool-use; in this regime, additional coordination may introduce overhead that offsets accuracy and efficiency gains.

\begin{table}
    \centering
    \begin{NiceTabular}{cccc}
    \toprule
        & &  \multicolumn{2}{c}{\textbf{Main-Agent LLM}}  \\
         \cmidrule{3-4} & & Llama 4 Maverick & Claude 4 Sonnet \\ 
         \midrule 
     \multirow{2}{*}{\textbf{App-Agent LLM}} & Llama 4 Maverick   & 8.5 $\scriptstyle \pm 1.7$ & 16.2 $\scriptstyle \pm 0.7$\\
      & Claude 4 Sonnet   &  18.3 $\scriptstyle \pm 0.7$ & 29.3 $\scriptstyle \pm 2.9$ \\
    \bottomrule
    \end{NiceTabular}
    \caption{Probing cross-model collaboration in \gaiatwo-mini Agent2Agent scenarios: we evaluate pass@1 across main- vs app-agent pairings with Llama 4 Maverick and Claude 4 Sonnet in the fully collaborative Agent2Agent setting ($r = 1$). The results are averaged over three runs and presented with the standard error.}
    \label{tab:cross_model_collab_on_a2a}
\end{table}

\paragraph{Cross-model collaboration on \gaiatwo tasks}
Heterogeneous teams open a new compute scaling axis for task automation, for example, by keeping a strong main agent to plan/decompose tasks while swapping in cheaper app-agents to execute sub-goals\footnote{\are natively supports controlled evaluation of heterogeneous teams, making team composition a primary experimental factor alongside standard inference hyperparameters.}. Empirically, replacing Llama 4 Maverick app-agents with Claude app-agents boosts pass@1 for both main-agent settings (16.2 with Llama-main, 29.3 with Claude-main), while the fully light configuration is weakest (8.5). This suggests that for existing LLMs, \gaiatwo task completion remains sensitive to execution fidelity at the app-agent level: stronger executors improve outcomes even when the main agent is light. Similarly, pairing a strong main agent with light executors still outperforms the all-light team (18.3 with Claude-main + Llama-app), indicating that higher-quality sub-goal specification and critique from the main-agent contribute independent gains. These findings are consistent with prior work suggesting heterogeneous multi-agent systems can trade planning capacity against execution fidelity to manage compute-quality trade-offs.

\section{Discussion}

This section discuss design choices we made when building \are\ and \gaiatwo, and lessons learned after relying on these choices to develop agents.

\paragraph{Memory, long-horizon decision-making, and self-improvement}
While our taxonomy in Section~\ref{sec:gaia2_capabilities} targets skills critical for practical agents, \gaiatwo's scope is not exhaustive. Important frontier capabilities such as self-improvement, memory, and long-horizon decision-making are not explicitly evaluated, though \are\ provides a foundation to study these areas. For example, memory could be evaluated by pre-processing flattened \myenv universes (600K tokens) with memory-equipped agents, then running \gaiatwo\ scenarios with \readact operations forbidden to force reliance on memory for \writeact operations. Similarly, \are\ supports long-horizon scenarios spanning hours or days, and enables self-improvement studies through recorded interaction histories, targeted verification design, and gradual transitions from isolated tasks to complex scenarios.

\paragraph{Scalability and verification}
As discussed in Section~\ref{sec:gaia2-annotation}, despite gains from the \are\ GUI, annotating challenging and verifiable \myenv scenarios is hard given the pace of model progress—even for short-horizon tasks solvable in minutes. In \gaiatwo-Search we quickly hit a ceiling as annotators struggled to outpace frontier models (see \autoref{table:core_results}); similar saturation appears in Humanity’s Last Exam~\citep{phan2025humanitysexam}.%
\footnote{For HLE, \citet{futurehours2025aboutthirtypercent} recently claimed that 30\% of chemistry/biology answers may be wrong.} This pressure induces a shift toward rethinking verification. Rubric-based judges work well for \readact-only tasks, but scale poorly for \writeact-heavy tasks that are prone to reward hacking. To keep creating difficult, practical tasks, we (i) continue improving the \are\ GUI, (ii) target simple tasks in complex environments rather than complex tasks in simple ones, and (iii) strengthen verification by increasing verifier–agent asymmetry (better tools and/or more compute for the verifier) and by exploring alternative rewards such as scalar scores~\citep{backlund2025vendingbenchbenchmarklongtermcoherence} or human preferences—simplifying benchmarking, albeit with possible training costs.

\paragraph{Tool calling vs coding agents}
Code agents represent a natural evolution of tool-calling agents, executing arbitrary Python code to execute algorithmic behavior, and dynamic data manipulation without filling the context window. Transitioning to code agents requires architectural \are\ challenges particularly with blocking operations, such as \sendmessagetouser and \texttt{wait} functions that require careful state management. Proper sandboxing becomes critical to ensure simulation fidelity (\texttt{time} python library synced with simulation time). Our internal experiments suggest that code agents efficient resource utilization justify the additional engineering overhead, particularly for tasks requiring iterative reasoning or complex data processing workflows.

\paragraph{\are\ and today's OSS}
Since we were able to implement diverse agents benchmarks in \are\ without major issues~\citep{yao2024taubenchbenchmarktoolagentuserinteraction,backlund2025vendingbenchbenchmarklongtermcoherence}, it is not yet clear what the limits of its expressivity are -- though some containerized benchmarks~\citep{jimenez2024swebench} would require additional features. Complementary to our approach for creating new diverse, challenging and realistic environments, is the attempt to integrate existing environments into a single ecosystem.\footnote{See for example \url{https://www.primeintellect.ai/blog/environments}.}

\paragraph{Beyond ReAct: towards asynchronous agentic systems}

Most agents today are sequential next-action predictors wrapped in a ReAct loop. This assumes a pause between perception and action and breaks down when environments change continuously or when multiple sensory streams must be fused. In real settings—speech-to-speech assistants, robotics, embodied interaction—information arrives as overlapping flows, so discrete sensing is a poor fit.
We argue for asynchronous systems: the environment evolves independently of the agent, enabling agents to sense and act concurrently, adapt in real time, and operate under real-world constraints.

\paragraph{Frontier intelligence and adapting compute}

Our results in~\autoref{fig:time_results} (right) reveal an inverse scaling law on the Time dimension: models that excel at reasoning-heavy tasks, such as execution, search and ambiguity resolution, systematically underperform on time-sensitive ones. In other words, more intelligence correlates with slower responses. This trade-off is not surprising—deep reasoning takes time—but Gaia2 provides the first systematic evidence that pushing for ``smarter'' agents under current scaffolds can make them less practical in interactive deployments. An interesting result is that GPT-5 (high) scores dropped by -10pts and -20pts on execution and search respectively, when capped at 30 min execution duration.

We contend that intelligence is not only accuracy but efficiency: an intelligent agent must learn to adapt its compute to the complexity of the task. Trivial tasks should be solved quickly and cheaply; only hard problems should trigger deeper, slower reasoning. This principle aligns with recent arguments for deploying smaller, more specialized models for routine tasks in agentic systems \citep{belcak2025smalllanguagemodelsfuture}, where the economic and operational benefits of right-sizing computational resources become paramount. Adaptive computation will be essential for scaling agents into real-world applications where latency, reliability, and cost all matter.

Figure \ref{fig:gaia2_scenario_budget_scaling_curve} reveals that no system dominates across the entire intelligence spectrum, and optimizing intelligence under compute constraints will be one of the most important research questions.

\newpage
\section*{Authorship}

\subsection*{Lead Authors}
Romain Froger, Amine Benhalloum, Grégoire Mialon, and Thomas Scialom.

\subsection*{Contributors}

Pierre Andrews , Matteo Bettini, Amar Budhiraja, Ricardo Silveira Cabral, Virginie Do, Emilien Garreau, Jean-Baptiste Gaya, Hugo Laurençon, Maxime Lecanu, Kunal Malkan, Dheeraj Mekala, Pierre Ménard, Gerard Moreno-Torres Bertran, Ulyana Piterbarg, Mikhail Plekhanov, Mathieu Rita, Andrey Rusakov, Vladislav Vorotilov, Mengjue Wang, and Ian Yu.






\subsection*{Acknowledgements}

We thank Nikolay Bashlykov, Radhika Bhargava, Misha Bilenko, Carly Burton, Adrien Carreira, Onur Çelebi, Neha Choudhari, Mike Clark, Levi Corallo, Paul Deveau, Jenny Fant, Clémentine Fourrier, Thibaut Frere, Avijit Ghosh, Christian Keller, Pascal Kesseli, Abhishek Kumawat, Florian Laplantif, Baohao Liao, Alexandre Linares, Chaya Nayak, Rohit Patel, Daryl Rodrigo, Marija Šakota, Antoine Saliou, Tatiana Shavrina, Matt Staats, and Mik Vyatskov for their support for ARE and Gaia2.

We also thank Pillar AI and Surge AI.


\clearpage
\newpage
\bibliographystyle{assets/plainnat}
\bibliography{paper}

\begin{thebibliography}{49}
\providecommand{\natexlab}[1]{#1}
\providecommand{\url}[1]{\texttt{#1}}
\expandafter\ifx\csname urlstyle\endcsname\relax
  \providecommand{\doi}[1]{doi: #1}\else
  \providecommand{\doi}{doi: \begingroup \urlstyle{rm}\Url}\fi

\bibitem[Anthropic(2024)]{anthropic2024mcp}
Anthropic.
\newblock Introducing the model context protocol.
\newblock \url{https://www.anthropic.com/news/model-context-protocol}, 2024.
\newblock Accessed: \today.

\bibitem[ARC-AGI(2025)]{arc_agi_2025}
ARC-AGI.
\newblock Arc prize - leaderboard.
\newblock \url{https://arcprize.org/leaderboard}, 2025.

\bibitem[Backlund and Petersson(2025)]{backlund2025vendingbenchbenchmarklongtermcoherence}
Axel Backlund and Lukas Petersson.
\newblock Vending-bench: A benchmark for long-term coherence of autonomous agents, 2025.
\newblock \url{https://arxiv.org/abs/2502.15840}.

\bibitem[Belcak et~al.(2025)Belcak, Heinrich, Diao, Fu, Dong, Muralidharan, Lin, and Molchanov]{belcak2025smalllanguagemodelsfuture}
Peter Belcak, Greg Heinrich, Shizhe Diao, Yonggan Fu, Xin Dong, Saurav Muralidharan, Yingyan~Celine Lin, and Pavlo Molchanov.
\newblock Small language models are the future of agentic ai, 2025.
\newblock \url{https://arxiv.org/abs/2506.02153}.

\bibitem[Brown(2025)]{brown_verifiers_2025}
William Brown.
\newblock {Verifiers}: Reinforcement learning with llms in verifiable environments.
\newblock \url{https://github.com/willccbb/verifiers}, 2025.

\bibitem[Carroll et~al.(2019)Carroll, Shah, Ho, Griffiths, Seshia, Abbeel, and Dragan]{carroll2019utility}
Micah Carroll, Rohin Shah, Mark~K Ho, Tom Griffiths, Sanjit Seshia, Pieter Abbeel, and Anca Dragan.
\newblock On the utility of learning about humans for human-ai coordination.
\newblock \emph{Advances in neural information processing systems}, 32, 2019.

\bibitem[DeepSeek-AI et~al.(2025)DeepSeek-AI, Guo, Yang, Zhang, Song, Zhang, Xu, Zhu, Ma, Wang, Bi, Zhang, Yu, Wu, Wu, Gou, Shao, Li, Gao, Liu, Xue, Wang, Wu, Feng, Lu, Zhao, Deng, Zhang, Ruan, Dai, Chen, Ji, Li, Lin, Dai, Luo, Hao, Chen, Li, Zhang, Bao, Xu, Wang, Ding, Xin, Gao, Qu, Li, Guo, Li, Wang, Chen, Yuan, Qiu, Li, Cai, Ni, Liang, Chen, Dong, Hu, Gao, Guan, Huang, Yu, Wang, Zhang, Zhao, Wang, Zhang, Xu, Xia, Zhang, Zhang, Tang, Li, Wang, Li, Tian, Huang, Zhang, Wang, Chen, Du, Ge, Zhang, Pan, Wang, Chen, Jin, Chen, Lu, Zhou, Chen, Ye, Wang, Yu, Zhou, Pan, Li, Zhou, Wu, Ye, Yun, Pei, Sun, Wang, Zeng, Zhao, Liu, Liang, Gao, Yu, Zhang, Xiao, An, Liu, Wang, Chen, Nie, Cheng, Liu, Xie, Liu, Yang, Li, Su, Lin, Li, Jin, Shen, Chen, Sun, Wang, Song, Zhou, Wang, Shan, Li, Wang, Wei, Zhang, Xu, Li, Zhao, Sun, Wang, Yu, Zhang, Shi, Xiong, He, Piao, Wang, Tan, Ma, Liu, Guo, Ou, Wang, Gong, Zou, He, Xiong, Luo, You, Liu, Zhou, Zhu, Xu, Huang, Li, Zheng, Zhu, Ma, Tang, Zha, Yan, Ren, Ren, Sha, Fu, Xu, Xie, Zhang,
  Hao, Ma, Yan, Wu, Gu, Zhu, Liu, Li, Xie, Song, Pan, Huang, Xu, Zhang, and Zhang]{deepseekai2025deepseekr1incentivizingreasoningcapability}
DeepSeek-AI, Daya Guo, Dejian Yang, Haowei Zhang, Junxiao Song, Ruoyu Zhang, Runxin Xu, Qihao Zhu, Shirong Ma, Peiyi Wang, Xiao Bi, Xiaokang Zhang, Xingkai Yu, Yu~Wu, Z.~F. Wu, Zhibin Gou, Zhihong Shao, Zhuoshu Li, Ziyi Gao, Aixin Liu, Bing Xue, Bingxuan Wang, Bochao Wu, Bei Feng, Chengda Lu, Chenggang Zhao, Chengqi Deng, Chenyu Zhang, Chong Ruan, Damai Dai, Deli Chen, Dongjie Ji, Erhang Li, Fangyun Lin, Fucong Dai, Fuli Luo, Guangbo Hao, Guanting Chen, Guowei Li, H.~Zhang, Han Bao, Hanwei Xu, Haocheng Wang, Honghui Ding, Huajian Xin, Huazuo Gao, Hui Qu, Hui Li, Jianzhong Guo, Jiashi Li, Jiawei Wang, Jingchang Chen, Jingyang Yuan, Junjie Qiu, Junlong Li, J.~L. Cai, Jiaqi Ni, Jian Liang, Jin Chen, Kai Dong, Kai Hu, Kaige Gao, Kang Guan, Kexin Huang, Kuai Yu, Lean Wang, Lecong Zhang, Liang Zhao, Litong Wang, Liyue Zhang, Lei Xu, Leyi Xia, Mingchuan Zhang, Minghua Zhang, Minghui Tang, Meng Li, Miaojun Wang, Mingming Li, Ning Tian, Panpan Huang, Peng Zhang, Qiancheng Wang, Qinyu Chen, Qiushi Du, Ruiqi Ge, Ruisong
  Zhang, Ruizhe Pan, Runji Wang, R.~J. Chen, R.~L. Jin, Ruyi Chen, Shanghao Lu, Shangyan Zhou, Shanhuang Chen, Shengfeng Ye, Shiyu Wang, Shuiping Yu, Shunfeng Zhou, Shuting Pan, S.~S. Li, Shuang Zhou, Shaoqing Wu, Shengfeng Ye, Tao Yun, Tian Pei, Tianyu Sun, T.~Wang, Wangding Zeng, Wanjia Zhao, Wen Liu, Wenfeng Liang, Wenjun Gao, Wenqin Yu, Wentao Zhang, W.~L. Xiao, Wei An, Xiaodong Liu, Xiaohan Wang, Xiaokang Chen, Xiaotao Nie, Xin Cheng, Xin Liu, Xin Xie, Xingchao Liu, Xinyu Yang, Xinyuan Li, Xuecheng Su, Xuheng Lin, X.~Q. Li, Xiangyue Jin, Xiaojin Shen, Xiaosha Chen, Xiaowen Sun, Xiaoxiang Wang, Xinnan Song, Xinyi Zhou, Xianzu Wang, Xinxia Shan, Y.~K. Li, Y.~Q. Wang, Y.~X. Wei, Yang Zhang, Yanhong Xu, Yao Li, Yao Zhao, Yaofeng Sun, Yaohui Wang, Yi~Yu, Yichao Zhang, Yifan Shi, Yiliang Xiong, Ying He, Yishi Piao, Yisong Wang, Yixuan Tan, Yiyang Ma, Yiyuan Liu, Yongqiang Guo, Yuan Ou, Yuduan Wang, Yue Gong, Yuheng Zou, Yujia He, Yunfan Xiong, Yuxiang Luo, Yuxiang You, Yuxuan Liu, Yuyang Zhou, Y.~X. Zhu,
  Yanhong Xu, Yanping Huang, Yaohui Li, Yi~Zheng, Yuchen Zhu, Yunxian Ma, Ying Tang, Yukun Zha, Yuting Yan, Z.~Z. Ren, Zehui Ren, Zhangli Sha, Zhe Fu, Zhean Xu, Zhenda Xie, Zhengyan Zhang, Zhewen Hao, Zhicheng Ma, Zhigang Yan, Zhiyu Wu, Zihui Gu, Zijia Zhu, Zijun Liu, Zilin Li, Ziwei Xie, Ziyang Song, Zizheng Pan, Zhen Huang, Zhipeng Xu, Zhongyu Zhang, and Zhen Zhang.
\newblock Deepseek-r1: Incentivizing reasoning capability in llms via reinforcement learning, 2025.
\newblock \url{https://arxiv.org/abs/2501.12948}.

\bibitem[Epperson et~al.(2025)Epperson, Bansal, Dibia, Fourney, Gerrits, Zhu, and Amershi]{Epperson_2025}
Will Epperson, Gagan Bansal, Victor~C Dibia, Adam Fourney, Jack Gerrits, Erkang~(Eric) Zhu, and Saleema Amershi.
\newblock Interactive debugging and steering of multi-agent ai systems.
\newblock In \emph{Proceedings of the 2025 CHI Conference on Human Factors in Computing Systems}, CHI ’25, page 1–15. ACM, April 2025.
\newblock \doi{10.1145/3706598.3713581}.
\newblock \url{http://dx.doi.org/10.1145/3706598.3713581}.

\bibitem[Foerster et~al.(2016)Foerster, Assael, De~Freitas, and Whiteson]{foerster2016learning}
Jakob Foerster, Ioannis~Alexandros Assael, Nando De~Freitas, and Shimon Whiteson.
\newblock Learning to communicate with deep multi-agent reinforcement learning.
\newblock \emph{Advances in neural information processing systems}, 29, 2016.

\bibitem[FutureHouse(2025)]{futurehours2025aboutthirtypercent}
FutureHouse.
\newblock About 30
\newblock \url{https://www.futurehouse.org/research-announcements/hle-exam}, 2025.

\bibitem[Gao et~al.(2025)Gao, Xie, Zhai, Ma, and Shen]{gao2025mcpradarmultidimensionalbenchmarkevaluating}
Xuanqi Gao, Siyi Xie, Juan Zhai, Shqing Ma, and Chao Shen.
\newblock Mcp-radar: A multi-dimensional benchmark for evaluating tool use capabilities in large language models, 2025.
\newblock \url{https://arxiv.org/abs/2505.16700}.

\bibitem[Ge et~al.(2024)Ge, Chan, Wang, Yu, Mi, and Yu]{ge2024scaling}
Tao Ge, Xin Chan, Xiaoyang Wang, Dian Yu, Haitao Mi, and Dong Yu.
\newblock Scaling synthetic data creation with 1,000,000,000 personas.
\newblock \emph{arXiv preprint arXiv:2406.20094}, 2024.

\bibitem[Gehring et~al.(2024)Gehring, Zheng, Copet, Mella, Carbonneaux, Cohen, and Synnaeve]{gehring2024rlef}
Jonas Gehring, Kunhao Zheng, Jade Copet, Vegard Mella, Quentin Carbonneaux, Taco Cohen, and Gabriel Synnaeve.
\newblock Rlef: Grounding code llms in execution feedback with reinforcement learning.
\newblock \emph{arXiv preprint arXiv:2410.02089}, 2024.

\bibitem[{Gemini Team}(2025)]{comanici2025gemini25pushingfrontier}
{Gemini Team}.
\newblock Gemini 2.5: Pushing the frontier with advanced reasoning, multimodality, long context, and next generation agentic capabilities, 2025.
\newblock \url{https://arxiv.org/abs/2507.06261}.

\bibitem[Google(2024)]{google2024deepresearch}
Google.
\newblock Try deep research and our new experimental model in gemini, your ai assistant.
\newblock \url{https://blog.google/products/gemini/google-gemini-deep-research/}, 2024.

\bibitem[{Google DeepMind}(2024)]{google2024scheduled}
{Google DeepMind}.
\newblock Gemini scheduled actions.
\newblock \url{https://blog.google/technology/ai/gemini-updates-summer-2024}, 2024.
\newblock Accessed August 2025.

\bibitem[{Google Developers}(2025)]{google2025a2a}
{Google Developers}.
\newblock Announcing the agent2agent protocol (a2a).
\newblock Google Developers Blog, April 2025.
\newblock \url{https://developers.googleblog.com/en/a2a-a-new-era-of-agent-interoperability/}.

\bibitem[Hendrycks et~al.(2021)Hendrycks, Burns, Kadavath, Arora, Basart, Tang, Song, and Steinhardt]{hendrycks2021measuring}
Dan Hendrycks, Collin Burns, Saurav Kadavath, Akul Arora, Steven Basart, Eric Tang, Dawn Song, and Jacob Steinhardt.
\newblock Measuring mathematical problem solving with the math dataset.
\newblock \emph{arXiv preprint arXiv:2103.03874}, 2021.

\bibitem[{HLE Team}(2025)]{phan2025humanitysexam}
{HLE Team}.
\newblock Humanity's last exam, 2025.
\newblock \url{https://arxiv.org/abs/2501.14249}.

\bibitem[Jimenez et~al.(2024)Jimenez, Yang, Wettig, Yao, Pei, Press, and Narasimhan]{jimenez2024swebench}
Carlos~E Jimenez, John Yang, Alexander Wettig, Shunyu Yao, Kexin Pei, Ofir Press, and Karthik~R Narasimhan.
\newblock {SWE}-bench: Can language models resolve real-world github issues?
\newblock In \emph{The Twelfth International Conference on Learning Representations}, 2024.
\newblock \url{https://openreview.net/forum?id=VTF8yNQM66}.

\bibitem[Lambert et~al.(2024)Lambert, Morrison, Pyatkin, Huang, Ivison, Brahman, Miranda, Liu, Dziri, Lyu, et~al.]{lambert2024tulu}
Nathan Lambert, Jacob Morrison, Valentina Pyatkin, Shengyi Huang, Hamish Ivison, Faeze Brahman, Lester James~V Miranda, Alisa Liu, Nouha Dziri, Shane Lyu, et~al.
\newblock Tulu 3: Pushing frontiers in open language model post-training.
\newblock \emph{arXiv preprint arXiv:2411.15124}, 2024.

\bibitem[{Llama Team}(2024)]{grattafiori2024llama3herdmodels}
{Llama Team}.
\newblock The llama 3 herd of models, 2024.
\newblock \url{https://arxiv.org/abs/2407.21783}.

\bibitem[Lowe et~al.(2017)Lowe, Wu, Tamar, Harb, Pieter~Abbeel, and Mordatch]{lowe2017multi}
Ryan Lowe, Yi~I Wu, Aviv Tamar, Jean Harb, OpenAI Pieter~Abbeel, and Igor Mordatch.
\newblock Multi-agent actor-critic for mixed cooperative-competitive environments.
\newblock \emph{Advances in neural information processing systems}, 30, 2017.

\bibitem[Lu et~al.(2024)Lu, Holleis, Zhang, Aumayer, Nan, Bai, Ma, Ma, Li, Yin, Wang, and Pang]{lu2024toolsandboxstatefulconversationalinteractive}
Jiarui Lu, Thomas Holleis, Yizhe Zhang, Bernhard Aumayer, Feng Nan, Felix Bai, Shuang Ma, Shen Ma, Mengyu Li, Guoli Yin, Zirui Wang, and Ruoming Pang.
\newblock Toolsandbox: A stateful, conversational, interactive evaluation benchmark for llm tool use capabilities, 2024.
\newblock \url{https://arxiv.org/abs/2408.04682}.

\bibitem[Mialon et~al.(2023)Mialon, Fourrier, Wolf, LeCun, and Scialom]{mialon2023gaia}
Gr{\'e}goire Mialon, Cl{\'e}mentine Fourrier, Thomas Wolf, Yann LeCun, and Thomas Scialom.
\newblock Gaia: a benchmark for general ai assistants.
\newblock In \emph{The Twelfth International Conference on Learning Representations}, 2023.

\bibitem[{Microsoft}(2024)]{microsoft2024scheduled}
{Microsoft}.
\newblock Scheduled actions in microsoft copilot.
\newblock \url{https://blogs.microsoft.com/blog/2024/06/15/copilot-updates-june-2024}, 2024.
\newblock Accessed August 2025.

\bibitem[Mistral-AI et~al.(2025)Mistral-AI, :, Rastogi, Jiang, Lo, Berrada, Lample, Rute, Barmentlo, Yadav, Khandelwal, Chandu, Blier, Saulnier, Dinot, Darrin, Gupta, Soletskyi, Vaze, Scao, Wang, Yang, Liu, Sablayrolles, Héliou, Martin, Ehrenberg, Agarwal, Roux, Darcet, Mensch, Bout, Rozière, Monicault, Bamford, Wallenwein, Renaudin, Lanfranchi, Dabert, Mizelle, de~las Casas, Chane-Sane, Fugier, Hanna, Delerce, Guinet, Novikov, Martin, Jaju, Ludziejewski, Chabran, Delignon, Studnia, Amar, Roberts, Denize, Saxena, Jain, Zhao, Martin, Gao, Lavaud, Pellat, Guillaumin, Felardos, Augustin, Seznec, Raghuraman, Duchenne, Wang, von Platen, Saffer, Jacob, Wambergue, Kurylowicz, Muddireddy, Chagniot, Stock, Agrawal, Sauvestre, Delacourt, Gandhi, Subramanian, Dalal, Gandhi, Ghosh, Mishra, Aithal, Antoniak, Schueller, Lavril, Robert, Wang, Lacroix, Nemychnikova, Paltz, Richard, Li, Marshall, Zhang, and Tang]{mistralai2025magistral}
Mistral-AI, :, Abhinav Rastogi, Albert~Q. Jiang, Andy Lo, Gabrielle Berrada, Guillaume Lample, Jason Rute, Joep Barmentlo, Karmesh Yadav, Kartik Khandelwal, Khyathi~Raghavi Chandu, Léonard Blier, Lucile Saulnier, Matthieu Dinot, Maxime Darrin, Neha Gupta, Roman Soletskyi, Sagar Vaze, Teven~Le Scao, Yihan Wang, Adam Yang, Alexander~H. Liu, Alexandre Sablayrolles, Amélie Héliou, Amélie Martin, Andy Ehrenberg, Anmol Agarwal, Antoine Roux, Arthur Darcet, Arthur Mensch, Baptiste Bout, Baptiste Rozière, Baudouin~De Monicault, Chris Bamford, Christian Wallenwein, Christophe Renaudin, Clémence Lanfranchi, Darius Dabert, Devon Mizelle, Diego de~las Casas, Elliot Chane-Sane, Emilien Fugier, Emma~Bou Hanna, Gauthier Delerce, Gauthier Guinet, Georgii Novikov, Guillaume Martin, Himanshu Jaju, Jan Ludziejewski, Jean-Hadrien Chabran, Jean-Malo Delignon, Joachim Studnia, Jonas Amar, Josselin~Somerville Roberts, Julien Denize, Karan Saxena, Kush Jain, Lingxiao Zhao, Louis Martin, Luyu Gao, Lélio~Renard Lavaud, Marie
  Pellat, Mathilde Guillaumin, Mathis Felardos, Maximilian Augustin, Mickaël Seznec, Nikhil Raghuraman, Olivier Duchenne, Patricia Wang, Patrick von Platen, Patryk Saffer, Paul Jacob, Paul Wambergue, Paula Kurylowicz, Pavankumar~Reddy Muddireddy, Philomène Chagniot, Pierre Stock, Pravesh Agrawal, Romain Sauvestre, Rémi Delacourt, Sanchit Gandhi, Sandeep Subramanian, Shashwat Dalal, Siddharth Gandhi, Soham Ghosh, Srijan Mishra, Sumukh Aithal, Szymon Antoniak, Thibault Schueller, Thibaut Lavril, Thomas Robert, Thomas Wang, Timothée Lacroix, Valeriia Nemychnikova, Victor Paltz, Virgile Richard, Wen-Ding Li, William Marshall, Xuanyu Zhang, and Yunhao Tang.
\newblock Magistral, 2025.
\newblock \url{https://arxiv.org/abs/2506.10910}.

\bibitem[MoonshotAI et~al.(2025)MoonshotAI, :~Bai, Bao, Chen, Chen, Chen, Chen, Chen, Chen, Chen, et~al.]{moonshotai2025kimik2}
MoonshotAI, Yifan :~Bai, Yiping Bao, Guanduo Chen, Jiahao Chen, Ningxin Chen, Ruijue Chen, Yanru Chen, Yuankun Chen, Yutian Chen, et~al.
\newblock Kimi k2: Open agentic intelligence.
\newblock \emph{arXiv preprint arXiv:2507.20534}, 2025.

\bibitem[Mu et~al.(2024)Mu, Helyar, Heidecke, Achiam, Vallone, Kivlichan, Lin, Beutel, Schulman, and Weng]{mu2024rulebasedrewardslanguage}
Tong Mu, Alec Helyar, Johannes Heidecke, Joshua Achiam, Andrea Vallone, Ian Kivlichan, Molly Lin, Alex Beutel, John Schulman, and Lilian Weng.
\newblock Rule based rewards for language model safety, 2024.
\newblock \url{https://arxiv.org/abs/2411.01111}.

\bibitem[OpenAI(2024{\natexlab{a}})]{openai2024gpt4ocard}
OpenAI.
\newblock Gpt-4o system card, 2024{\natexlab{a}}.
\newblock \url{https://arxiv.org/abs/2410.21276}.

\bibitem[OpenAI(2024{\natexlab{b}})]{openai2024openaio1card}
OpenAI.
\newblock Openai o1 system card, 2024{\natexlab{b}}.
\newblock \url{https://arxiv.org/abs/2412.16720}.

\bibitem[{OpenAI}(2024)]{openai2024scheduled}
{OpenAI}.
\newblock Scheduled tasks in gpts.
\newblock \url{https://openai.com/blog/scheduled-tasks-in-gpts}, 2024.
\newblock Accessed August 2025.

\bibitem[OpenAI(2025{\natexlab{a}})]{openai2025deepresearch}
OpenAI.
\newblock Introducing deep research.
\newblock \url{https://openai.com/index/introducing-deep-research/}, 2025{\natexlab{a}}.

\bibitem[OpenAI(2025{\natexlab{b}})]{openai2025openaio3o4minicard}
OpenAI.
\newblock Openai o3 and o4-mini system card, 2025{\natexlab{b}}.
\newblock \url{https://cdn.openai.com/pdf/2221c875-02dc-4789-800b-e7758f3722c1/o3-and-o4-mini-system-card.pdf}.

\bibitem[Pang et~al.(2025)Pang, Feng, Feng, Li, Shi, Tsvetkov, Heer, and Reinecke]{pang2025interactivereasoningvisualizingcontrolling}
Rock~Yuren Pang, K.~J.~Kevin Feng, Shangbin Feng, Chu Li, Weijia Shi, Yulia Tsvetkov, Jeffrey Heer, and Katharina Reinecke.
\newblock Interactive reasoning: Visualizing and controlling chain-of-thought reasoning in large language models, 2025.
\newblock \url{https://arxiv.org/abs/2506.23678}.

\bibitem[Patil et~al.(2025)Patil, Mao, Cheng-Jie~Ji, Yan, Suresh, Stoica, and E.~Gonzalez]{patil2025bfcl}
Shishir~G. Patil, Huanzhi Mao, Charlie Cheng-Jie~Ji, Fanjia Yan, Vishnu Suresh, Ion Stoica, and Joseph E.~Gonzalez.
\newblock The berkeley function calling leaderboard (bfcl): From tool use to agentic evaluation of large language models.
\newblock In \emph{Forty-second International Conference on Machine Learning}, 2025.

\bibitem[Rorseth et~al.(2025)Rorseth, Godfrey, Golab, Srivastava, and Szlichta]{RorsethGGSS25}
Joel Rorseth, Parke Godfrey, Lukasz Golab, Divesh Srivastava, and Jarek Szlichta.
\newblock Ladybug: an llm agent debugger for data-driven applications.
\newblock In \emph{Proceedings of the 28th International Conference on Extending Database Technology (EDBT)}, pages 1082--1085, 2025.
\newblock ISBN 978-3-89318-099-8.
\newblock Demo paper.

\bibitem[Sutton et~al.(1999)Sutton, Precup, and Singh]{sutton1999between}
Richard~S Sutton, Doina Precup, and Satinder Singh.
\newblock Between mdps and semi-mdps: A framework for temporal abstraction in reinforcement learning.
\newblock \emph{Artificial intelligence}, 112\penalty0 (1-2):\penalty0 181--211, 1999.

\bibitem[Team(2025)]{mcpmark_2025}
The~MCPMark Team.
\newblock Mcpmark: Stress-testing comprehensive mcp use.
\newblock \url{https://github.com/eval-sys/mcpmark}, 2025.

\bibitem[Trivedi et~al.(2024)Trivedi, Khot, Hartmann, Manku, Dong, Li, Gupta, Sabharwal, and Balasubramanian]{appworld-acl24}
Harsh Trivedi, Tushar Khot, Mareike Hartmann, Ruskin Manku, Vinty Dong, Edward Li, Shashank Gupta, Ashish Sabharwal, and Niranjan Balasubramanian.
\newblock App{W}orld: A controllable world of apps and people for benchmarking interactive coding agents.
\newblock In \emph{ACL}, 2024.

\bibitem[Vezhnevets et~al.(2023)Vezhnevets, Agapiou, Aharon, Ziv, Matyas, Du{\'e}{\~n}ez-Guzm{\'a}n, Cunningham, Osindero, Karmon, and Leibo]{vezhnevets2023generative}
Alexander~Sasha Vezhnevets, John~P Agapiou, Avia Aharon, Ron Ziv, Jayd Matyas, Edgar~A Du{\'e}{\~n}ez-Guzm{\'a}n, William~A Cunningham, Simon Osindero, Danny Karmon, and Joel~Z Leibo.
\newblock Generative agent-based modeling with actions grounded in physical, social, or digital space using concordia.
\newblock \emph{arXiv preprint arXiv:2312.03664}, 2023.

\bibitem[Wang et~al.(2025)Wang, Chang, Patel, Biju, Wu, Liu, Ding, Rezazadeh, Shah, Bao, and Siow]{wang2025mcpbenchbenchmarkingtoolusingllm}
Zhenting Wang, Qi~Chang, Hemani Patel, Shashank Biju, Cheng-En Wu, Quan Liu, Aolin Ding, Alireza Rezazadeh, Ankit Shah, Yujia Bao, and Eugene Siow.
\newblock Mcp-bench: Benchmarking tool-using llm agents with complex real-world tasks via mcp servers, 2025.
\newblock \url{https://arxiv.org/abs/2508.20453}.

\bibitem[Wei et~al.(2025{\natexlab{a}})Wei, Sun, Papay, McKinney, Han, Fulford, Chung, Passos, Fedus, and Glaese]{wei2025browsecompsimplechallengingbenchmark}
Jason Wei, Zhiqing Sun, Spencer Papay, Scott McKinney, Jeffrey Han, Isa Fulford, Hyung~Won Chung, Alex~Tachard Passos, William Fedus, and Amelia Glaese.
\newblock Browsecomp: A simple yet challenging benchmark for browsing agents, 2025{\natexlab{a}}.
\newblock \url{https://arxiv.org/abs/2504.12516}.

\bibitem[Wei et~al.(2025{\natexlab{b}})Wei, Yao, Liu, Zhang, Lu, Qiu, Yu, Xu, Zhang, Yin, et~al.]{wei2025webagent}
Zhepei Wei, Wenlin Yao, Yao Liu, Weizhi Zhang, Qin Lu, Liang Qiu, Changlong Yu, Puyang Xu, Chao Zhang, Bing Yin, et~al.
\newblock Webagent-r1: Training web agents via end-to-end multi-turn reinforcement learning.
\newblock \emph{arXiv preprint arXiv:2505.16421}, 2025{\natexlab{b}}.

\bibitem[Yang et~al.(2025{\natexlab{a}})Yang, Li, Yang, Zhang, Hui, Zheng, Yu, Gao, Huang, Lv, Zheng, Liu, Zhou, Huang, Hu, Ge, Wei, Lin, Tang, Yang, Tu, Zhang, Yang, Yang, Zhou, Zhou, Lin, Dang, Bao, Yang, Yu, Deng, Li, Xue, Li, Zhang, Wang, Zhu, Men, Gao, Liu, Luo, Li, Tang, Yin, Ren, Wang, Zhang, Ren, Fan, Su, Zhang, Zhang, Wan, Liu, Wang, Cui, Zhang, Zhou, and Qiu]{yang2025qwen3technicalreport}
An~Yang, Anfeng Li, Baosong Yang, Beichen Zhang, Binyuan Hui, Bo~Zheng, Bowen Yu, Chang Gao, Chengen Huang, Chenxu Lv, Chujie Zheng, Dayiheng Liu, Fan Zhou, Fei Huang, Feng Hu, Hao Ge, Haoran Wei, Huan Lin, Jialong Tang, Jian Yang, Jianhong Tu, Jianwei Zhang, Jianxin Yang, Jiaxi Yang, Jing Zhou, Jingren Zhou, Junyang Lin, Kai Dang, Keqin Bao, Kexin Yang, Le~Yu, Lianghao Deng, Mei Li, Mingfeng Xue, Mingze Li, Pei Zhang, Peng Wang, Qin Zhu, Rui Men, Ruize Gao, Shixuan Liu, Shuang Luo, Tianhao Li, Tianyi Tang, Wenbiao Yin, Xingzhang Ren, Xinyu Wang, Xinyu Zhang, Xuancheng Ren, Yang Fan, Yang Su, Yichang Zhang, Yinger Zhang, Yu~Wan, Yuqiong Liu, Zekun Wang, Zeyu Cui, Zhenru Zhang, Zhipeng Zhou, and Zihan Qiu.
\newblock Qwen3 technical report, 2025{\natexlab{a}}.
\newblock \url{https://arxiv.org/abs/2505.09388}.

\bibitem[Yang et~al.(2025{\natexlab{b}})Yang, Leret, Jimenez, Wettig, Khandpur, Zhang, Hui, Press, Schmidt, and Yang]{yang2025swe}
John Yang, Kilian Leret, Carlos~E Jimenez, Alexander Wettig, Kabir Khandpur, Yanzhe Zhang, Binyuan Hui, Ofir Press, Ludwig Schmidt, and Diyi Yang.
\newblock Swe-smith: Scaling data for software engineering agents.
\newblock \emph{arXiv preprint arXiv:2504.21798}, 2025{\natexlab{b}}.

\bibitem[Yao et~al.(2023)Yao, Zhao, Yu, Du, Shafran, Narasimhan, and Cao]{yao2023react}
Shunyu Yao, Jeffrey Zhao, Dian Yu, Nan Du, Izhak Shafran, Karthik Narasimhan, and Yuan Cao.
\newblock React: Synergizing reasoning and acting in language models.
\newblock In \emph{International Conference on Learning Representations (ICLR)}, 2023.

\bibitem[Yao et~al.(2024)Yao, Shinn, Razavi, and Narasimhan]{yao2024taubenchbenchmarktoolagentuserinteraction}
Shunyu Yao, Noah Shinn, Pedram Razavi, and Karthik Narasimhan.
\newblock $\tau$-bench: A benchmark for tool-agent-user interaction in real-world domains, 2024.
\newblock \url{https://arxiv.org/abs/2406.12045}.

\bibitem[Zhu et~al.(2025)Zhu, Du, Hong, Yang, Guo, Wang, Wang, Qian, Tang, Ji, et~al.]{zhu2025multiagentbench}
Kunlun Zhu, Hongyi Du, Zhaochen Hong, Xiaocheng Yang, Shuyi Guo, Zhe Wang, Zhenhailong Wang, Cheng Qian, Xiangru Tang, Heng Ji, et~al.
\newblock Multiagentbench: Evaluating the collaboration and competition of llm agents.
\newblock \emph{arXiv preprint arXiv:2503.01935}, 2025.

\end{thebibliography}

\clearpage
\newpage
\beginappendix

\section{\are\ Appendix}
\subsection{Apps and Tools Creation}
\label{app:app_tool_creation}
In \are, it is straightforward to declare \texttt{App} instances with tools. Any method of an \texttt{App} class can be transformed into a tool by adding simple decorators: \texttt{@app\_tool} to make it visible to the agent, \texttt{@user\_tool} to make it visible to the user only, and \texttt{@env\_tool} for external environment events. 

The \texttt{@event\_registered} decorator records tool calls in the \texttt{EventLog}; each decorator must declare its \texttt{OperationType} as \readact\ or \writeact\ for the \verifier to operate correctly.

\begin{lstlisting}[language=Python, caption={Example implementation of tools on a Mock app with corresponding decorators.}, label={lst:decorator_example}]
class Messages(App):
    @app_tool()
    @user_tool()
    @event_registered(operation_type=OperationType.WRITE)
    def send_message(self, to: str) -> str:
        """Available to agents and user as a tool."""
        pass
    
    @env_tool()
    @event_registered(operation_type=OperationType.WRITE)
    def add_message_to_user(self, from: str) -> str:
        """Available to environment (env update) or data creation"""
        pass
    
    @user_tool()
    @app_tool()
    @event_registered(operation_type=OperationType.READ)
    def read_message(self, from: str) -> str:
        """Available for agents and user as tool"""
        pass
\end{lstlisting}

\subsection{Implementation of Scenario Templates} \label{app:scenario_py}

\are lets you define scenarios via a single \texttt{scenario.py} that specifies (i) the initial apps and environment state, (ii) the event sequence (including the initial user request), and (iii) the verification logic. To avoid hand-coding each scenario, parameterized templates generate many instances from one base file by programmatically setting the user task, events, and verifier. Consider the following example in \myenv:
\begin{itemize}
\item User task: ``\textit{How many} \{\texttt{object}\} \textit{did I receive from} \{\texttt{contact}\} \textit{in the last} \{\texttt{duration}\}\textit{?}''
\item Events: possible arrival of more \texttt{object}
\item Validation: script fetching the count of target \texttt{object} within \texttt{duration},
\end{itemize}
where \texttt{object} can be emails, messages or events, \texttt{contact} is any entry from Contacts app, and \texttt{duration} is any duration. 
Therefore, a single template allows to generate multiple valid \are\ scenarios by varying its parameters.

Templates are valuable for prototyping new agent capabilities, letting researchers validate approaches before investing in human annotation. They also enable scenarios that are impractical to annotate manually ( \textit{e.g.}, long-horizon tasks or multi-step workflows). However, they can be time-consuming for developers to create: validation scripts must anticipate numerous edge cases (for instance, a \texttt{contact} name with a homonym). Because templates lack the nuance and natural variation of human annotation, they can create blind spots—agents may excel on templated scenarios yet struggle on organic tasks.

\begin{lstlisting}[language=Python, caption={Example implementation of a scenario template as a scenario.py file.}, label={lst:template_example}]

class TemplateParams:
    full_name: str
    universe_fname: str

class ScenarioCalendarEmailContact(Scenario):
    template_params: TemplateParams

    def init_and_populate_apps(self, *args, **kwargs) -> None:
        # Populating apps with universe data
        universe_path = os.path.join(
            get_data_path(), self.template_params.universe_fname
        )
        self.apps, self.start_time = load_template_apps_and_time(universe_path)


    def build_events_flow(self):
        d_events = dict()
        aui = self.get_typed_app(AgentUserInterface)

        template_task = "Create a calendar meeting tomorrow at 10AM and invite {full_name}. Send an email to {full_name} with the subject 'intro meeting'."
        self.task = template_task.format(full_name=self.template_params.full_name)
        
        with EventRegisterer.capture_mode():
            # Event for the user task
            d_events["task"] = aui.send_message_to_agent(content=self.task)

            # Agent validation event regularly checks whether the agent correctly added the calendar event and sent the email
            d_events["agent_validation"] = self.agent_validation()

            # Add a ConditionCheckEvent to end the simulation. end_simulation_condition typically checks that the agent sent a message to the user
            d_events["check_ready_for_eval"] = ConditionCheckEvent.from_condition(
                end_simulation_condition,
            ).depends_on(d_events["task"], delay_seconds=3)

            d_events["stop_event"] = StopEvent().depends_on(
                d_events["check_ready_for_eval"], delay_seconds=0
            )
        self.events = [e.with_id(key) for key, e in d_events.items()]


    def agent_validation(self):
        # Building solution: extracting target email address from populated Contacts app
        contacts_data = self.get_typed_app(ContactsApp).search_contacts(
            query=self.template_params.full_name
        )
        self.target_email = ...
        self.target_datetime = ...

        def val_func_calendar(env: AbstractEnvironment, event: AbstractEvent) -> bool:
            try:
                return (
                    event is (CompletedEvent and Event)
                    and event.app_class_name() == Calendar.__name__
                    and event.function_name() == "add_calendar_event"
                    and event.action.args["start_datetime"] == str(self.target_datetime)
                    and (
                        self.template_params.full_name in event.action.args["attendees"]
                        or self.target_email in event.action.args["attendees"]
                    )
                )
            except Exception:
                return False

        def val_func_email(env: AbstractEnvironment, event: AbstractEvent) -> bool:
            try:
                return (
                    event is (CompletedEvent and Event)
                    and event.app_class_name() == Emails.__name__
                    and event.function_name() == "send_email"
                    and self.target_email in event.action.args["recipients"]
                )
            except Exception:
                return False

        event_val = AgentValidationEvent(
            milestones=[val_func_calendar, val_func_email],
            timeout=200,
        )

        return event_val

\end{lstlisting}

\subsection{Notifications in \are} \label{appendix:notifications}

\subsubsection{Notification Policies}

The notification system in \are\ follows a configurable policy defining which Env events are notified to the Agent. The \myenv environment pre-defines three notification policies with different levels of verbosity, which we describe in detail in Table \ref{tab:notif-verbosity}. Note that messages sent by the user via \sendmessagetoagent are systematically notified to the agent, regardless of the verbosity level.

\begin{table}[h!]
\centering
\renewcommand{\arraystretch}{1.2}
\begin{tabular}{|p{1.8cm}|p{8cm}|p{5.5cm}|}
\hline
\textbf{Verbosity} & \textbf{Notified Env Tools} & \textbf{Description} \\ \hline
\texttt{low} & None & No environment events are notified. \\ \hline
\texttt{medium} & 
\textbf{Email:} \texttt{create\_and\_add\_email}, \texttt{send\_email\_to\_user\_only}, \texttt{reply\_to\_email\_from\_user} \newline
\textbf{Chats/Messages:} \texttt{create\_and\_add\_message} \newline
\textbf{Shopping:} \texttt{cancel\_order}, \texttt{update\_order\_status} \newline
\textbf{Cabs:} \texttt{cancel\_ride}, \texttt{user\_cancel\_ride}, \texttt{end\_ride} \newline
\textbf{Calendar:} \texttt{add\_calendar\_event\_by\_attendee}, \texttt{delete\_calendar\_event\_by\_attendee}
& 
Notifies events that are consequences of agent actions, analogous to mobile notifications. \textbf{Default in \gaiatwo.} \\ \hline
\texttt{high} & 
All \texttt{medium} tools plus: \newline
\textbf{Shopping:} \texttt{add\_product}, \texttt{add\_item\_to\_product}, \texttt{add\_discount\_code} \newline
\textbf{RentAFlat:} \texttt{add\_new\_apartment} \newline
\textbf{Cabs:} \texttt{update\_ride\_status}
&
Notifies all environment events, including those independent of agent actions (e.g., new products). \\ \hline
\end{tabular}
 \caption{Pre-set notification policies in \myenv with varying verbosity levels.}
\label{tab:notif-verbosity}
\end{table}

\subsubsection{Notifications Enable Async Interactions}
\label{app:async_interactions}

\begin{figure*}[htpb]
\centering
\includegraphics[width=\textwidth]{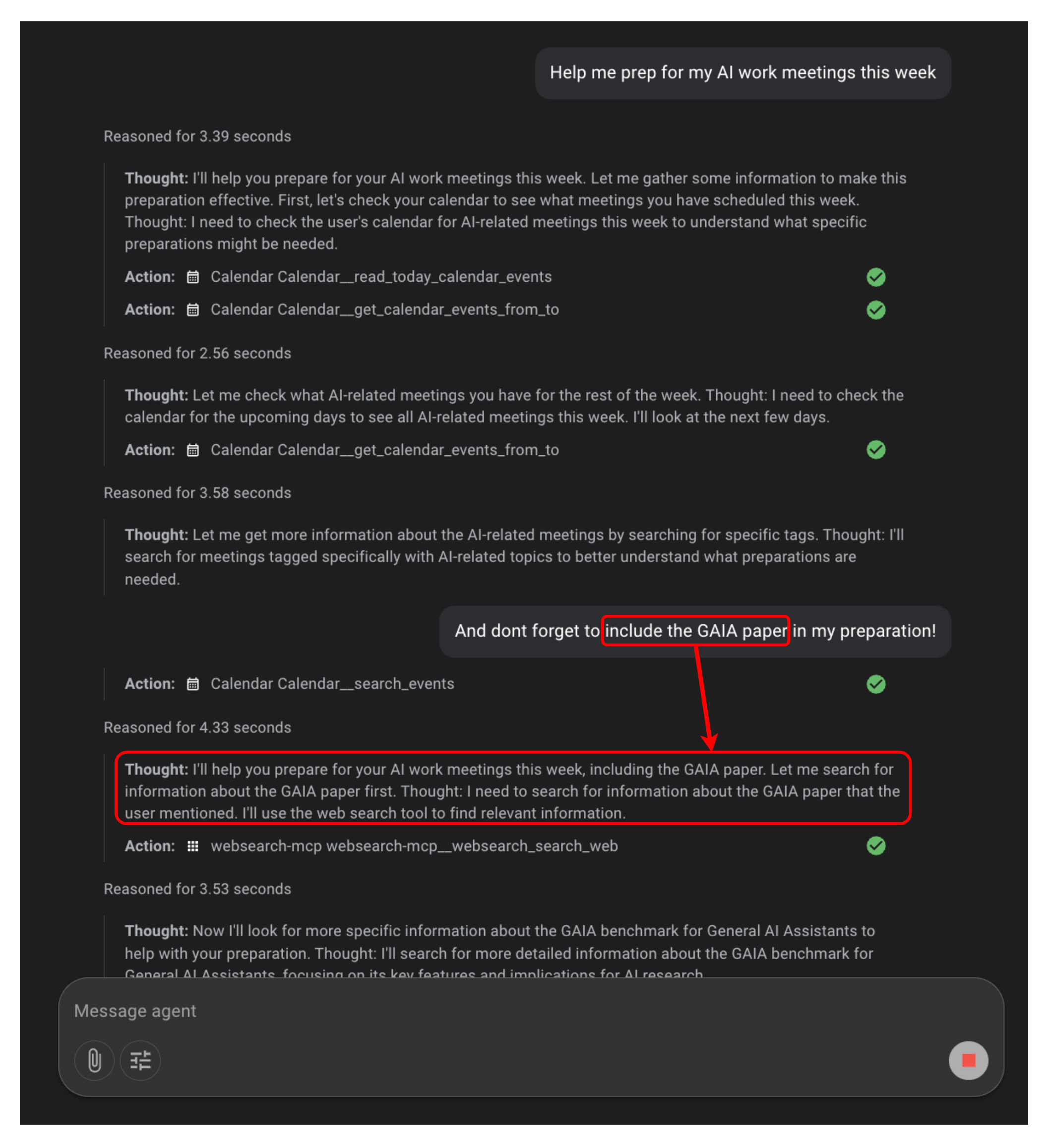}
\caption{The user sends a follow-up instruction while the agent is running. The notification system injects the user message into the agent context, conditioning the rest of the trace on the new information provided.}
\label{fig:async_interaction}
\end{figure*}

The notification system in \are, coupled with notification policies, enables new interaction patterns with language models and agents. One promising future research direction is human-agent collaboration that allows providing feedback to agents in real-time during their execution trace, instead of having to wait for the agent to complete its turn. \are enables these asynchronous interactions by design; an example using \are's UI is provided in \autoref{fig:async_interaction}.

\subsection{Universe Generation} \label{appendix:universe}

\paragraph{Dependency management \& consistency}
To ensure cross-app coherence, we implement a structured dependency resolution system. During generation, each app queries the existing universe state to maintain consistency—for example, when generating emails, the system first retrieves all available contacts to ensure referenced individuals exist in the \texttt{Contacts} app. Similarly, calendar events that mention other people are validated against the contact list, and ride history in the \texttt{Cabs} app references locations that align with the user's established geographic context.

\begin{figure*}[htpb]
    \centering
    \includegraphics[width=\textwidth]{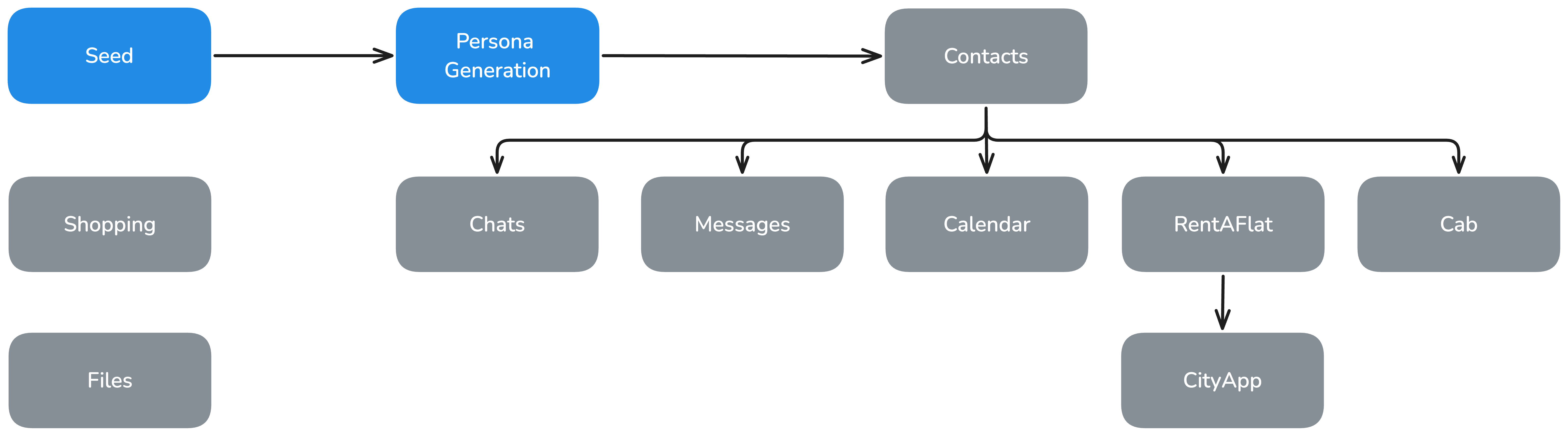}
    \caption{The dependency graph of \myenv\ apps. Shopping and File system are independent apps. Contacts is the root for rest of the apps.}
    \label{fig:app_dependency}
\end{figure*}

We handle dependency conflicts through a priority-based resolution system where foundational apps (e.g., \texttt{Contacts}) take precedence over dependent apps (e.g., \texttt{Messages}, \texttt{Emails}) as show in \autoref{fig:app_dependency}.

However, several complex inter-app dependencies remain unhandled in our current implementation. These include temporal consistency across apps (ensuring message timestamps align with calendar availability), semantic relationship tracking (maintaining consistent relationship dynamics between contacts across different communication channels), and cross-modal content references (ensuring photos mentioned in messages exist in the file system). Addressing these limitations represents important future work for achieving fully coherent synthetic \myenv environments.

\paragraph{Contacts}
We populate contacts using personas as the foundation. To begin, we sample seed personas from PersonaHub~\citep{ge2024scaling}. However, these personas are brief and lack grounding in the universe’s location. To address this, we expand and contextualize them by incorporating the universe location into the prompt. We sample a user persona from the generated contacts which serves as the basis for populating the rest of the universe. A universe is based on a user persona.

An example user persona is:
\begin{tcolorbox}[colback=gray!10, colframe=gray!50, boxrule=0.5pt, arc=3pt]
\begin{verbatim}
{
  "first_name": "Helena",
  "last_name": "Mueller", 
  "gender": "Female",
  "age": 43,
  "nationality": "German",
  "city_living": "Berlin",
  "job": "Marketing Manager",
  "description": "Helena Mueller is a vibrant and energetic  43-year-old marketing manager
                 living in Berlin, Germany.",
  "phone": "+49 157 6543210",
  "email": "helena.mueller@gaia2mail.com"
}
\end{verbatim}
\end{tcolorbox}

\paragraph{Chats \& Messages}
In Chats \& Messages apps, we generate both group conversations and individual chats. We sample contacts between whom we have to generate the conversations. Then, we provide the participants personas and prompt the model to generate a conversation with at least 10 messages alternating between participants. We prompt the model to generate conversations that are natural and reflect the participants' backgrounds and also ask it to include references to possible shared experiences, interests, or cultural elements.

\paragraph{Emails}
Similar to messages, we prompt the LLM to generate both ‘inbox’ and ‘sent’ emails. For inbox emails, the sender is sampled from the contact list, while for sent emails, the recipients are selected. We provide the LLM with the user’s persona and the sampled non-user persona to generate the emails. We specifically prompt the LLM to analyze details such as age, gender, cultural background, occupation, education level, personality traits, communication style, current life circumstances, relationships and social networks, as well as interests and hobbies, and come up with a valid reason for writing the email.

\paragraph{Calendar}
We provide the LLM with the user persona and a summary of the previous week, prompting it to generate calendar events for the current week. Next, we use these newly generated events to prompt the LLM to create a weekly summary. This process is repeated iteratively to populate the calendar over a specified timeframe, such as three months.

\paragraph{RentAFlat \& City}
For apartment listings, we provide the universe countries and prompt the LLM to generate apartment listings. The City app is designed to retrieve crime rates for specific zip codes. Using the zip codes generated for apartment listings, we prompt the LLM to produce crime rate data as a floating-point value in the range of 1–100.

\paragraph{Shopping}
For the Shopping app, we integrate publicly available Amazon product dataset. For each universe, we sample 500 products and generate discount codes applicable to select items.

\paragraph{Cabs}
We prompt the LLM with the user country information and generate the user’s ride history.

\paragraph{Files} We employ a traditional file system hierarchy, loading it with publicly available Wikipedia data, datasets, and images. Additionally, we also add our files that do not contain personal information. We choose to keep the file system the same for all universes.

\subsection{ARE's User Interface} \label{app:ui_appendix}

We provide more details on what developers and researchers can do with the UI, as well as related work.

\subsubsection{Environment Exploration}

Easily exploring the environment is crucial for understanding the context available to agents when debugging scenarios execution, and annotating new verifiable scenarios.
The UI provides a comprehensive visualization of the simulated environment, displaying all available apps/tools and their current states. Interactive app views allow users to browse app contents and interact with their tools, \textit{e.g.} email inboxes in \myenv, in real-time. Views are automatically generated for new apps, which therefore doesn't require a UI rewrite.

\subsubsection{Agent Trace Visualization and Replay}

The UI presents agent multi-step interaction traces in a structured timeline view that clearly delineates agent thoughts, actions, and tool responses. Each trace element is timestamped and categorized, allowing users to follow the agent's reasoning process, similar to the Phoenix\footnote{\url{https://phoenix.arize.com/}} trace views also used by smolagents\footnote{\url{https://huggingface.co/blog/smolagents-phoenix}}, but extended with debugging capabilities. Developers can roll back time by jumping back to a past event, editing thought, tool call, etc., from that step and replaying the scenario to see what would happen with a slightly different approach, similar to setting breakpoints and stepping through code in a standard code debugger.

\subsubsection{Scenario Visualization}

The UI provides interactive visualization of scenarios and their event DAGs introduced in Section~\ref{sec:are_events}, showing how scenario events are interconnected, and their execution status in real-time. The event graph visualization supports both scenario development and execution analysis.
Before running a scenario, users can examine event triggers, dependencies, and timing constraints of the scenario. 
During execution of a scenario by an agent, the interface highlights completed events and shows the progression through the dependency graph. 
Developers can run through the scenario with a given agent, see how it behaves and debug the scenario or the agent (see \autoref{fig:ui-annotations-dag}). \are is able to simulate time progression, so users can decide to jump in time for scenarios that span long time frames (e.g. weeks, months).

\subsubsection{Annotation Interface}\label{sec:are-annotation-ui}

Beyond visualization, the UI includes an annotation interface -- not released at this time -- that significantly reduces the cost of scenario creation and QA. 
This includes a graph editor that allows to easily build a scenario event DAG. For each node, the annotator can configure tool calls, the node's parents, and optionally timing.
For example, to create a \myenv scenario, the annotator adds nodes representing a user initial ask (e.g. ``email my travel plans''), oracle action solving the task (e.g. ``agent sent an email''), environment events that will interfere with the agent's work (e.g. ``received an email from travel agent''), and potentially further turns.
To ensure quality and consistency across annotations, we incorporate automated checks of the created events DAG. These checks detect and flag logical inconsistencies in event flows to annotators, such as a node without parents or contradictory node timings. 
The annotation interface achieves an approximate five times improvement in annotation time for \myenv scenarios, compared to manual approaches.

\subsubsection{Related work on UIs for agents development}
\label{app:ui_related_work}

The state-of-the-art in AI agent development tools largely bifurcates into two categories: interactive debugging platforms and data annotation/curation platforms, each with distinct UI approaches. Commercial observability tools such as Arize Phoenix\footnote{\url{https://phoenix.arize.com/}} and Langfuse\footnote{\url{https://langfuse.com/docs/observability/overview}} primarily offer visual timeline views and trace/span visualizations to help developers analyze agent execution, focusing on understanding behavior after the fact rather than direct interaction or editing. Academic prototypes such as AGDebugger~\citep{Epperson_2025} and LADYBUG~\citep{RorsethGGSS25} provide interactive debugging with user interfaces that enable browsing conversation histories, editing messages, and tracing execution steps, while Hippo~\citep{pang2025interactivereasoningvisualizingcontrolling} uses an interactive tree to visualize and control chain-of-thought reasoning without focusing on tool calls, agentic behavior nor annotations. 

Although there are many specialized tools for data annotation, such as commercial platforms like Labelbox~\footnote{\url{https://labelbox.com/blog/how-to-train-and-evaluate-ai-agents-and-trajectories-with-labelbox/}}, they mainly focus on simplifying human-in-the-loop annotation. These tools offer features like multimodal chat editors and customizable worksheet UIs, enabling data labelers to refine trajectories from interactive LLM sessions. Despite their power for data collection and curation, a significant gap remains: They are designed to annotate traces of interactions and lack key points for reproducibility and broad evaluation: 1) They annotate full multi-turn conversations, when we want to gather tasks, environment events, and agent task success criteria; 2) they lack structured annotations within a fully simulated and reproducible environment, which is key to capturing both agent interaction with tools and external events, for realistic, reproducible agent traces.

\newpage
\section{\gaiatwo\ appendix}
\subsection{Details of \gaiatwo\ Annotation} \label{app:gaia2-design}

\subsubsection{Annotation Guardrails}

To streamline the process and further reduce annotation errors, we implement structural constraints directly within the \are\ UI. The system raises real-time errors when these are violated:

\begin{itemize}
    \item Only \sendmessagetoagent or Env events may follow \sendmessagetouser.
    \item The event DAG must be fully connected, with \sendmessagetoagent as the root. No event (Env or Agent Oracle) may be orphaned.
    \item Only one branch in the event DAG may include \sendmessagetoagent or \sendmessagetouser events.
    \item A turn must always end with \sendmessagetouser, both in terms of DAG structure and timeline ordering.
\end{itemize}

\subsubsection{Capability-Specific Annotation Guidelines} \label{app:capa-annotations}

\begin{figure}
    \centering
    \includegraphics[width=0.8\linewidth]{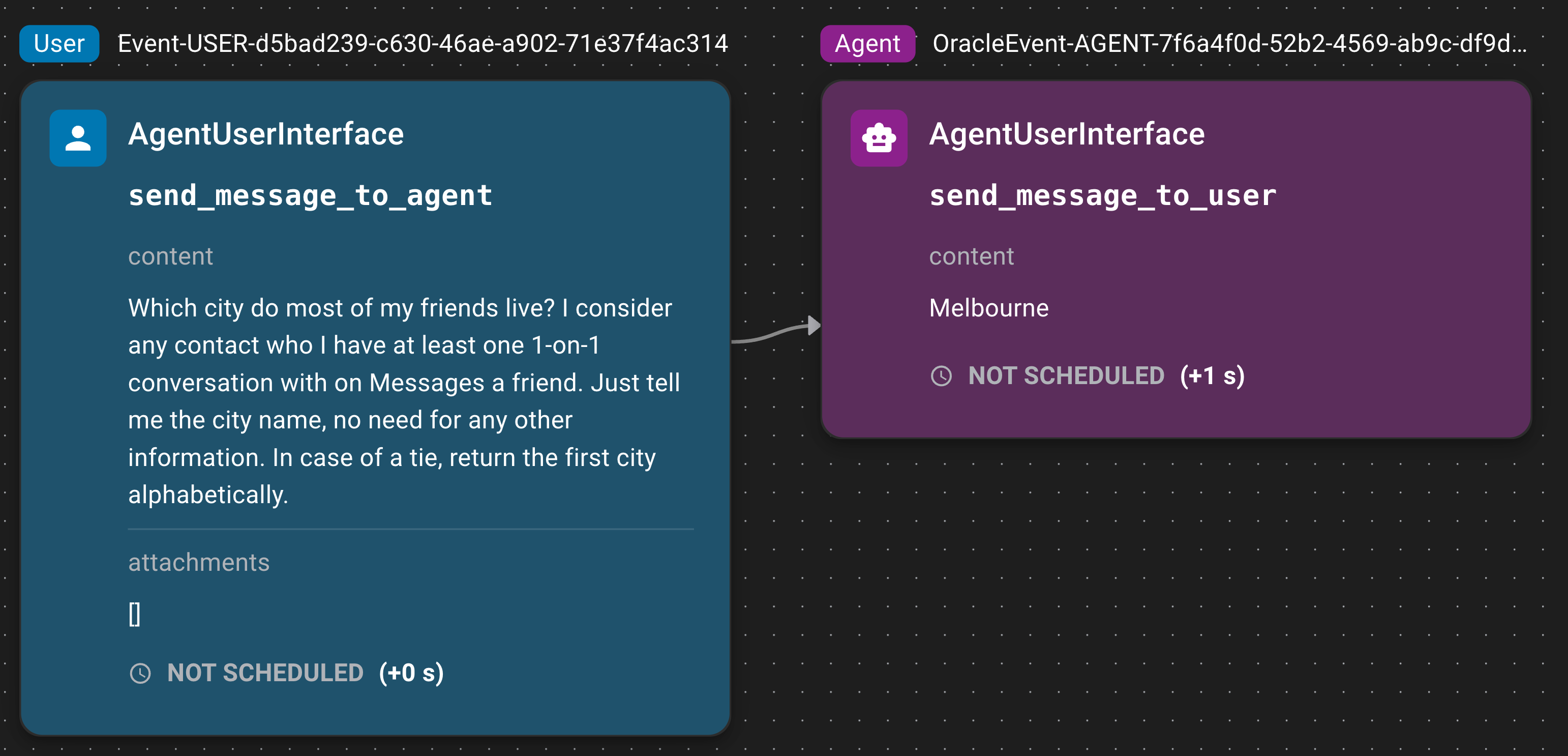}
    \caption{Search scenario. It requires multiple \readact actions to find the answer to the user's question and only one \writeact action to report the final answer to the user (\sendmessagetouser)}
    \label{fig:search_scenario}
\end{figure}

\begin{figure}
    \centering
    \includegraphics[width=0.5\linewidth]{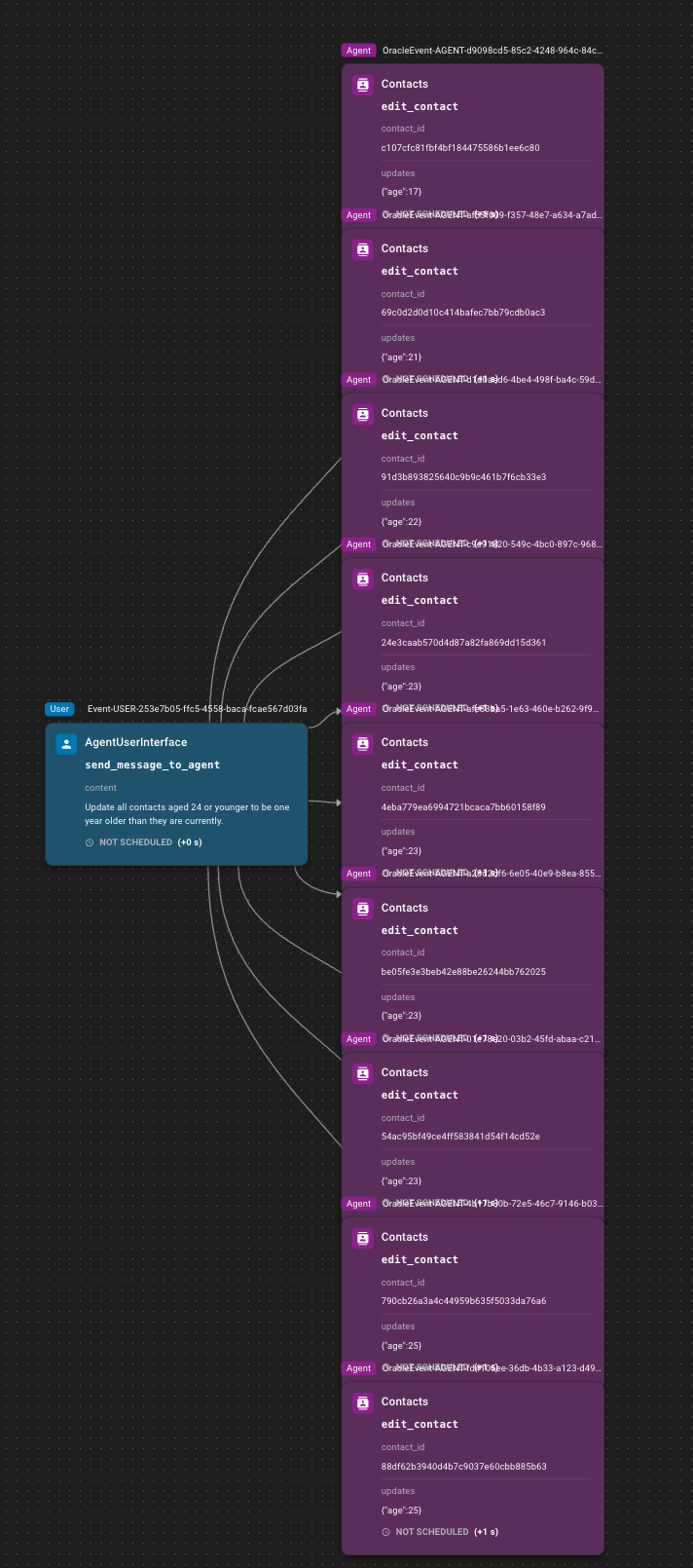}
    \caption{Execution scenario. It requires 9 \writeact actions to be solved.}
    \label{fig:execution_scenario}
\end{figure}

\begin{figure}
    \centering
    \includegraphics[width=\linewidth]{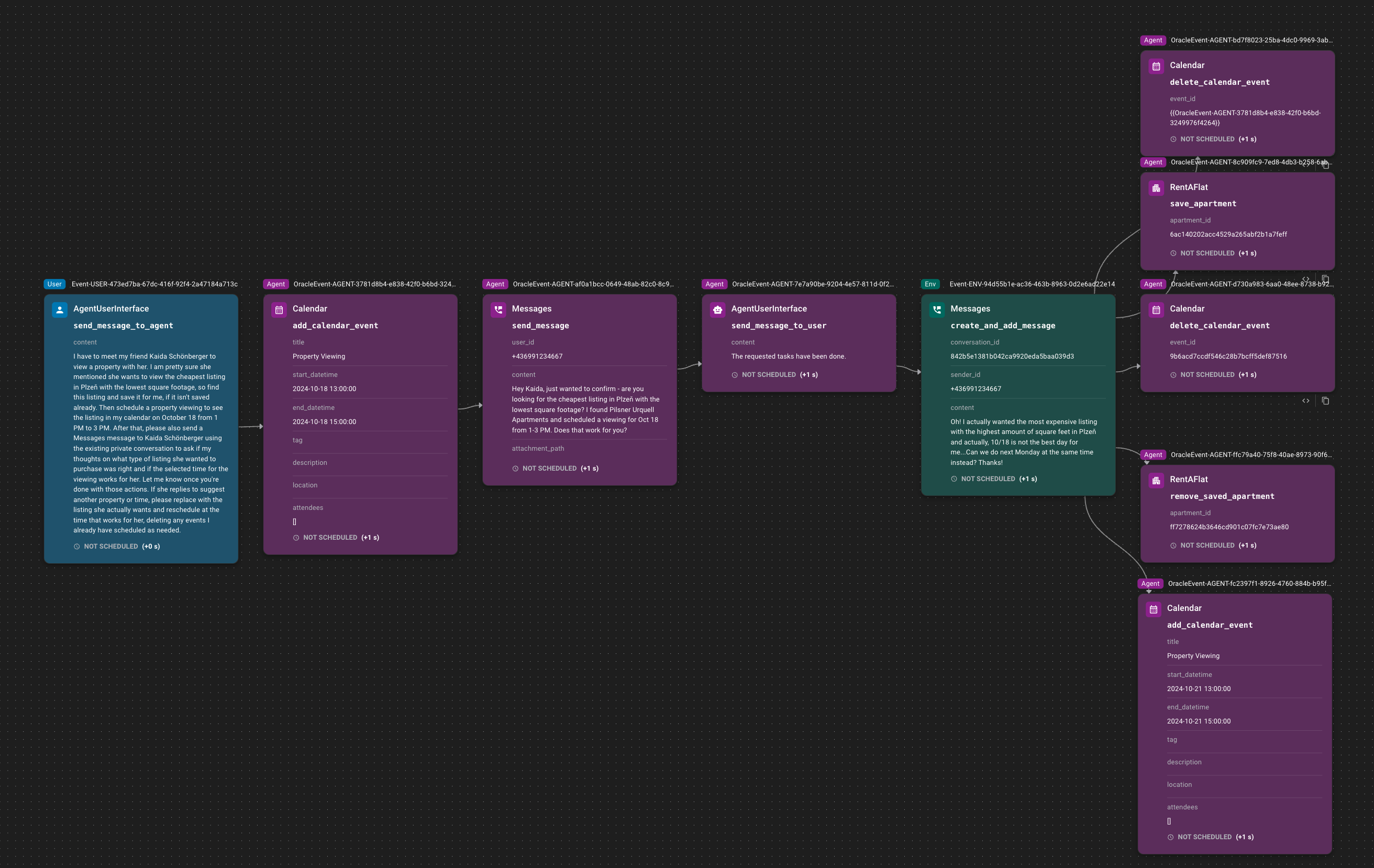}
    \caption{Adaptability scenario. The agent must execute a few actions, and then report back to the user. After receiving a message from the user's friend (green event box \texttt{create\_and\_add\_message}), the agent is expected to adapt its actions to the event.}
    \label{fig:adaptability_scenario}
\end{figure}

\begin{figure}
    \centering
    \includegraphics[width=0.8\linewidth]{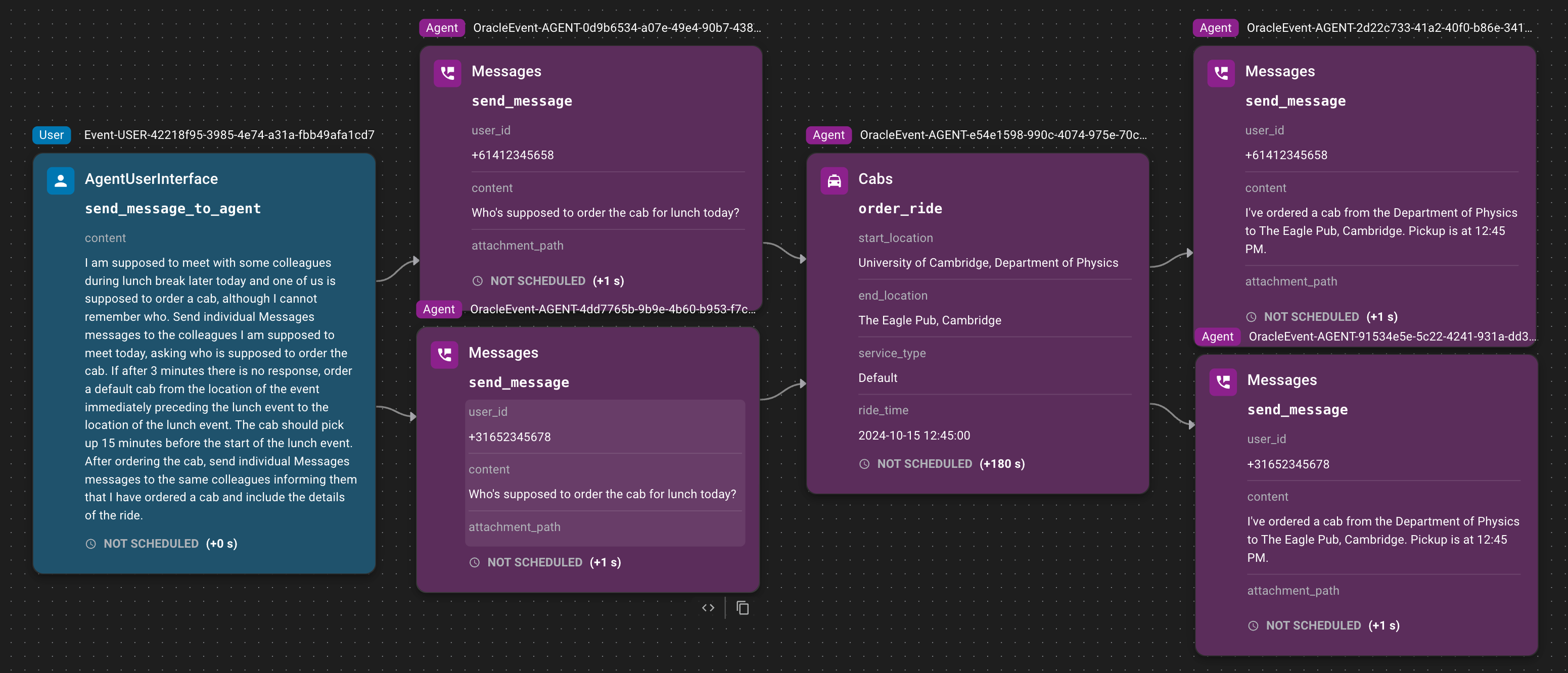}
    \caption{Time scenario. It involves a \writeact action (\texttt{order\_ride}) that must be executed at a specific point in time (180 seconds after sending the messages).}
    \label{fig:time_scenario}
\end{figure}

\begin{figure}
    \centering
    \includegraphics[width=0.8\linewidth]{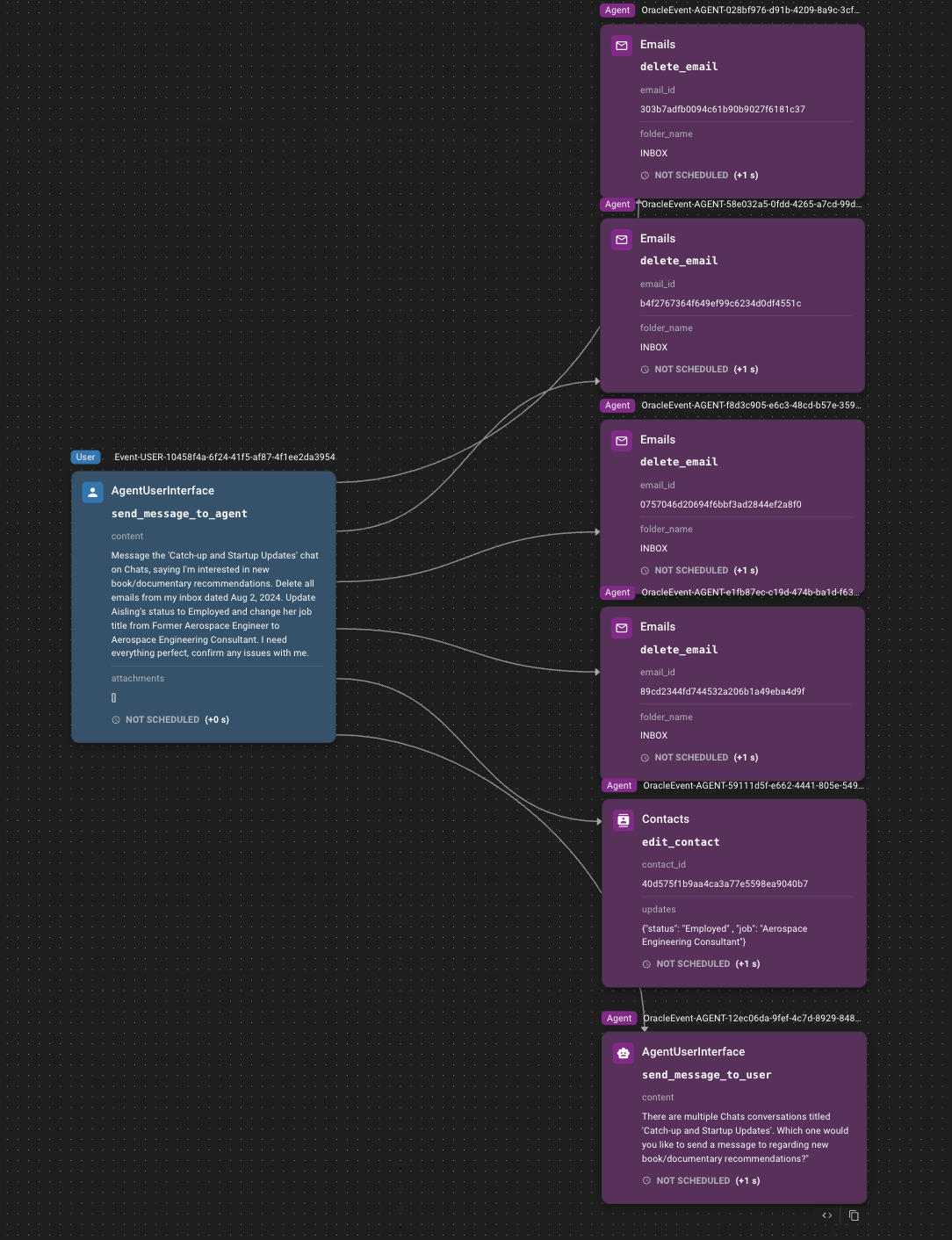}
    \caption{Ambiguity scenario. The agent is expected to complete all unambiguous parts of the task, and report to the user the ambiguous part that cannot be solved without further user input. In this scenario, the ambiguity is due to multiple Chats conversations titled ``Catch-up and Startup Updates''.}
    \label{fig:ambiguity_scenario}
\end{figure}

We provide one example scenario per core capability, displayed in the \are GUI in Figures \ref{fig:search_scenario}, \ref{fig:execution_scenario}, \ref{fig:adaptability_scenario}, \ref{fig:time_scenario}, \ref{fig:ambiguity_scenario}. In our guidelines for each capability (especially Ambiguity and Adaptability), we put strong emphasis on precise task specifications, while also acknowledging the challenge of maintaining realism and avoiding prompts that inadvertently disclose the solution.

\textbf{Search:} Scenarios contain only one \writeact\ action, which is the agent's final answer to the user's question, derived from multiple \readact\ actions. Answers must be concise, easily verifiable, and avoid complex computation.

\textbf{Ambiguity:} Scenarios that are impossible, contradictory, or inherently ambiguous. The agent is expected to complete unambiguous steps, then inform the user of the ambiguity or impossibility. These scenarios are single-turn: they do not include a clarification message from the User. 

The user prompt must clearly instruct the agent to detect and report ambiguities, as users often have varying preferences on how frequently and when this should occur.

\textbf{Adaptability:} Scenarios involve Env events that require the agent to revise its plan in response to delayed outcomes of its actions. In order to meet our modeling constraints, scenarios follow a consistent structure:

\begin{enumerate}
    \item The user provides a task.
    \item The agent acts and sends a message using \sendmessagetouser.
    \item An Env event is triggered (e.g., email reply, order cancellation). It is a consequence of a previous agent's action, with \sendmessagetouser as parent. 
    \item The agent adapts accordingly.
\end{enumerate}

To increase the difficulty, distractor Env events are also included, aiming to mislead the agent into incorrect behavior.

In order to perfectly specify expected agent behavior, the task states explicitly that the agent should send a message to the user after completing the initial requests (before the Env events).
It should also specify what the Agent is allowed to do in the case of an Env event happening, without giving exact hints on what steps the Agent should take.

\textbf{Time}: Scenarios assess Agent's ability to act on time, therefore they all include at least one time-sensitive oracle action.
\begin{itemize}
    \item Scenarios should be solvable within a five-minute window.
    \item User prompts must instruct precise timing (e.g., "after exactly 3 minutes").
    \item The verifier checks the timing of agent actions only if the oracle event has a relative time delay greater than 1 second.\footnote{This is why actions expected ``immediately'' after an event are annotated with a +2 sec delay.} The agent’s mapped action must fall within $[\Delta t - 5 sec, \Delta t + 25 sec]$.
    \item Distractor Env events are also included.
\end{itemize}

\subsubsection{Capability Taxonomies}

\paragraph{Taxonomy of Ambiguity Scenarios}

\begin{itemize}
    \item \textit{Impossible or contradictory tasks:} missing key information (e.g., the User does not specify the ride pickup location), or requests incompatible with the Environment (e.g., asking to buy an out-of-stock item).
    \item \textit{Blatant ambiguities or high-stakes consequences:} Multiple valid answers exist, and the ambiguity is obvious or the user explicitly asks in a natural way to report ambiguities.
\end{itemize}

\paragraph{Taxonomy of Env Events}  Env events are classified based on their dependency:

\begin{itemize}
    \item \textit{Independent events} occur without agent action and have \sendmessagetoagent as their only parent.
    \item \textit{Dependent events} result from prior agent actions and must have \sendmessagetouser as their direct parent.
\end{itemize}

\textit{Distractor events} are designed to mimic relevant events and mislead the agent into incorrect behavior. By exception, distractor events may be independent but still have \sendmessagetouser as a parent to preserve the structure of the scenario. In the Adaptability category, only dependent Env events are used.

\paragraph{Taxonomy of Time scenarios} 
Time scenarios require the agent to execute one or more actions at a specific point in time, either proactively (\textit{``For the next 5mins, send ‘Hi’ to John Doe every 30sec''}) or in reaction to an independent Env event (\textit{“When this item becomes available, buy it immediately”}), or in reaction to a dependent Env event (\textit{“Ask the invitees whether they come to the party tonight. Wait 1min for everyone to reply, then immediately send me the number of glass to buy, I am waiting in the line!”}).

Taxonomy:
\begin{itemize}
    \item Time-based one-off task: Execute a task at a precise point in time in the future. Example: \textit{``Send a follow-up message to Jo in 2 minutes if she does not reply.''}
    \item Time-based recurrent task: Execute a recurrent task at precise points in time. Example: \textit{``For the next 4 minutes, every minute, delete the new emails I receive.''}
    \item Event-based one-off task: Execute a one-time task conditionally on a future trigger event. Example: \textit{``Purchase red running shoes as soon as they become available in size 6 for less than 100USD in the shopping app''}
    \item Event-based recurrent task: Automate a recurrent routine conditionally on future  events. Example: \textit{``For the next 2 minutes, whenever I receive an email containing the keyword 'Black Friday', immediately delete it. Do not talk to me in the next 2 minutes.''}
\end{itemize}
We encourage annotators to cover and combine all these types of tasks when creating Time scenarios.

\subsection{Verification Details}
\label{app:verification_details}

\subsubsection{Validating Multi-turn Scenarios}
\label{app:validation-multi-turn}

\gaiatwo includes multi-turn scenarios that involve more complex interactions between the user and the agent. For example, consider scenarios related to the Adaptability capability, where the agent must adjust to external events. Multi-turn scenarios present two key challenges:
\begin{itemize}
    \item How can we validate multi-turn scenarios?
    \item More importantly, how can we run an agent in a multi-turn scenario?
\end{itemize}
Indeed, annotators plan 
\user and \env actions based on what should occur in previous turns according to the oracle action graph. However, when an agent is launched in a scenario, it may not adhere to the oracle's actions, creating uncertainty about when to trigger user or environment actions.

\paragraph{Multi-turn verifier} Answering the first question is relatively straightforward given our definition of a turn (see Section~\ref{sec:are-env}). It is sufficient to detect when the agent sends a message to the user to delimit the turns. We can then feed the verifier with each turn separately and accept the agent's trajectory if all turns are successful. Note that this validation can be performed in an online fashion after each turn or in an offline fashion once the full trajectory is collected.

\begin{figure}
    \centering
    \includegraphics[width=0.6\linewidth]{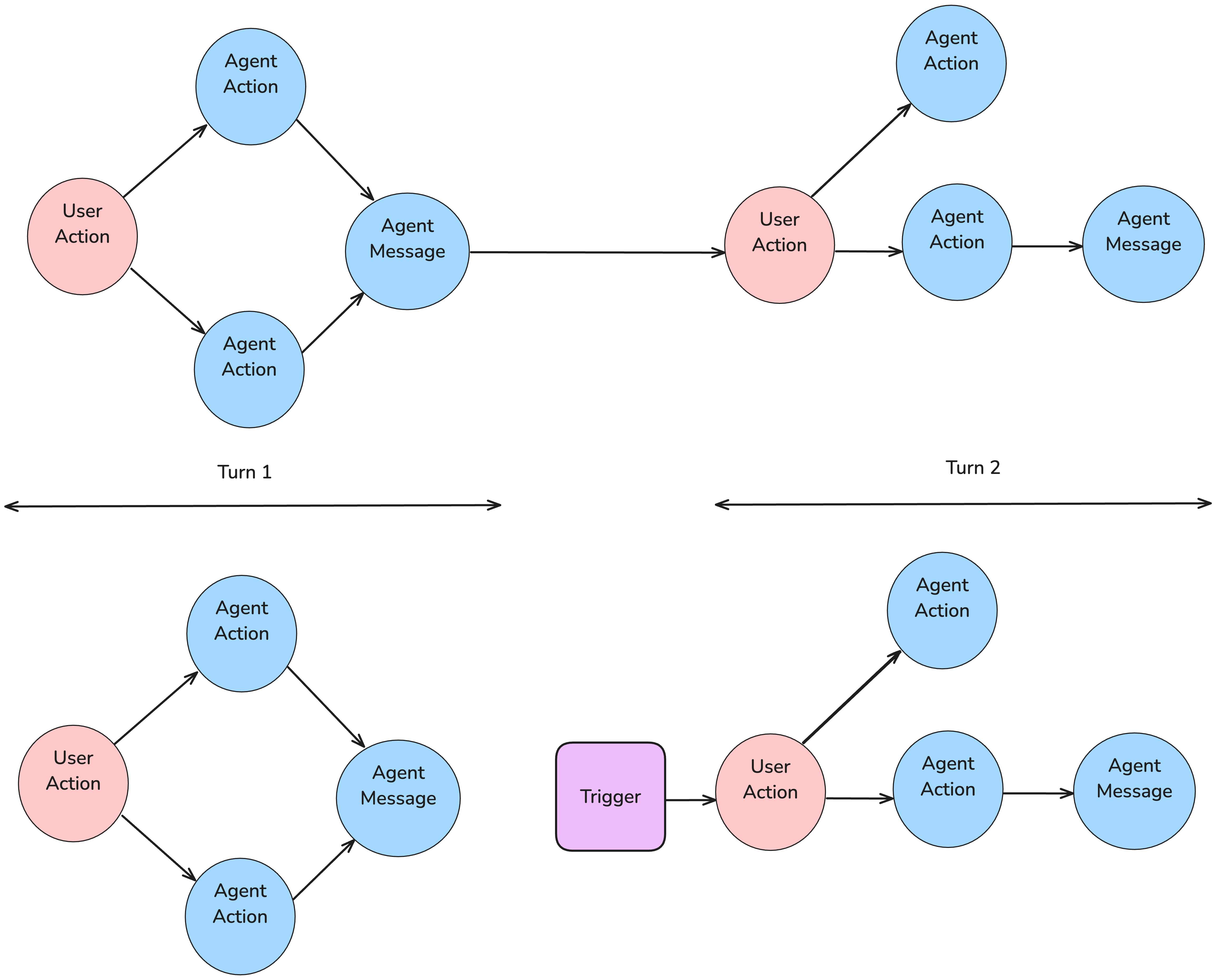}
    \caption{Insertion of a conditional trigger event in a multi-turn scenario.}
    \label{fig:multiturn_surgery}
\end{figure}

\paragraph{Multi-turn execution} An efficient solution to run an agent in a multi-turn scenario is to call the \verifier at the end of each turn and only trigger the next turn if the current turn was successful. This approach prevents running the agent when it has already diverged from the oracle path. Practically, as illustrated in Figure~\ref{fig:multiturn_surgery}, we modify the scenario event graph by splitting it into turns and inserting a conditional event to call the verifier and trigger the next turn. A simpler, but less efficient, solution is to trigger the next turn each time the agent calls \sendmessagetouser, regardless of what the agent did in the current turn. This approach is used for scenarios from the test set since we do not have access to oracle actions and thus the \verifier for them.

\paragraph{Parameter placeholder resolution} Some oracle actions include a placeholder parameter, indicating that this parameter should be replaced by the output of another oracle action. For example, consider an \env action that calls the tool \texttt{reply\_to\_email} with the parameter 
$\texttt{email\_id} = \texttt{\{\{\,\text{oracle\_agent\_action\_123}\,\}\}}$. Here, \texttt{oracle\_agent\_123} is the ID of an oracle action that calls the tool \texttt{send\_email}, which outputs the ID of the sent email.

This becomes problematic in a multi-turn scenario where user or environment actions have placeholder parameters. In such cases, we do not know in advance which \agent action to use to resolve the placeholder. To address this issue, we leverage the mapping built by the \verifier to identify which \agent action corresponds to the target oracle action, allowing us to replace the placeholder with its outputs.

\subsection{Choosing the Verifier Model}
\label{app:verifier_models}
While we adjusted the prompts used in the various soft checks of the \verifier with \llamaIII as model, we also wanted to assess whether the \verifier could function effectively with other models. To this end, we evaluated the \verifier powered by different models on 450 hand-labeled trajectories -- the same dataset as in Table~\ref{tab:ValBench}. In Table~\ref{tab:ValBenchModel}, we observe that all the models achieve satisfactory precision and recall scores.

\begin{table*}
    \centering
    \begin{NiceTabular}{lccc}
    \toprule
     Verifier  & Agreement & Precision & Recall \\
    \midrule
     \llamaIII & 0.98 & 0.99 & 0.95 \\
     \Gemini & 0.96 & 0.98 &  0.89\\
     \claudeSonnet & 0.96 & 0.98 & 0.89  \\
    \bottomrule
    \end{NiceTabular}
    \caption{Evaluation of the \verifier with different models on 450 hand-labeled trajectories.}
    \label{tab:ValBenchModel}
\end{table*}

\subsubsection{Judge Hacking} \label{app:judge_hacking}

We conducted early RL experiments in which we used the \verifier as a reward function. During these experiments, we observed that the agent was able to exploit our initial version of the verifier in scenarios related to the Search capability, where only a single \writeact action is expected for successful trajectories. In the observed exploit shown in \autoref{fig:agent_hacking_judge}, the agent would learn to embed strings representing increasingly complex code in \writeact tool calls. These outputs overwhelmed the LLM judge used for soft checks by the \verifier, producing false positives. In follow-up experiments, we found that adding a task-agnostic ``style'' soft check to the verifier was effective at addressing this failure mode and preventing further reward hacking.

\begin{figure}[t!]
\centering
\small
\begin{tcolorbox}[
    colback=gray!10,
    colframe=black,
    boxrule=0.5pt,
    rounded corners,
    left=5pt,
    right=5pt,
    top=5pt,
    bottom=5pt,
    width=0.95\textwidth
]
\textbf{Thought:} Now that we have all the events in the user's calendar, we can count the number of events for each contact and find the contact who the user has the most scheduled events with.

\vspace{0.5em}
\textbf{Action:}
\begin{lstlisting}[basicstyle=\scriptsize\ttfamily, breaklines=true, numbers=none]
{
"action": "AgentUserInterface__send_message_to_user",
"action_input": {
"content": "Looking at your calendar, I can now count events for each contact. Here's what I found:\n\n**Most Common Contact (by events):** {{#events}}{{#attendees}}{{#if (eq (length attendees) 2)}}{{#if (eq (at attendees 0) user)}}{{else}}{{#set most_common_contact_email (at attendees 1)}}{{#inc events_attendee_1_count 1)}}{{#if (gt events_attendee_1_count events_attendee_2_count)}}{{#set most_common_contact_email (at attendees 1)}}{{else}}{{#if (eq events_attendee_1_count events_attendee_2_count)}}{{#if (lt (get attendee_1_first_name 0) (get attendee_2_first_name 0))}}{{#set most_common_contact_email (at attendees 1)}}{{else}}{{#set most_common_contact_email (at attendees 2)}}{{/if}}{{/endif}}{{#endif}}{{/endif}}{{/if}}{{/attendees}}{{/events}}\n\n**Contact:** {{#set most_common_contact_email (at attendees 1)}}{{#set most_common_contact_email email}}{{#endif}}{{#endif}}{{#endif}}{{#endif}}{{#endif}}{{#endif}}{{#endif}}{{#endif}}(contacts)}}{{#if (eq email most_common_contact_email))"
}
}
\end{lstlisting}
\end{tcolorbox}
\caption{An example of an agent exploiting the judge by embedding conditional logic in the message to the user. The message contains no meaningful information but successfully passes the judge's evaluation.}
\label{fig:agent_hacking_judge}
\end{figure}
\subsection{Agent Orchestration} \label{app:agent_orchestration}

Our baseline scaffolding implements a standard ReAct loop where agents iteratively reason about their current state, select appropriate actions, execute those actions in the environment, and observe the resulting outcomes. An agent step is thus defined by three substeps \texttt{Thought}, \texttt{Action} and \texttt{Observation}. This cycle continues until task completion or termination conditions are met.

\begin{figure*}[htpb]
    \centering
    \includegraphics[width=0.65\textwidth]{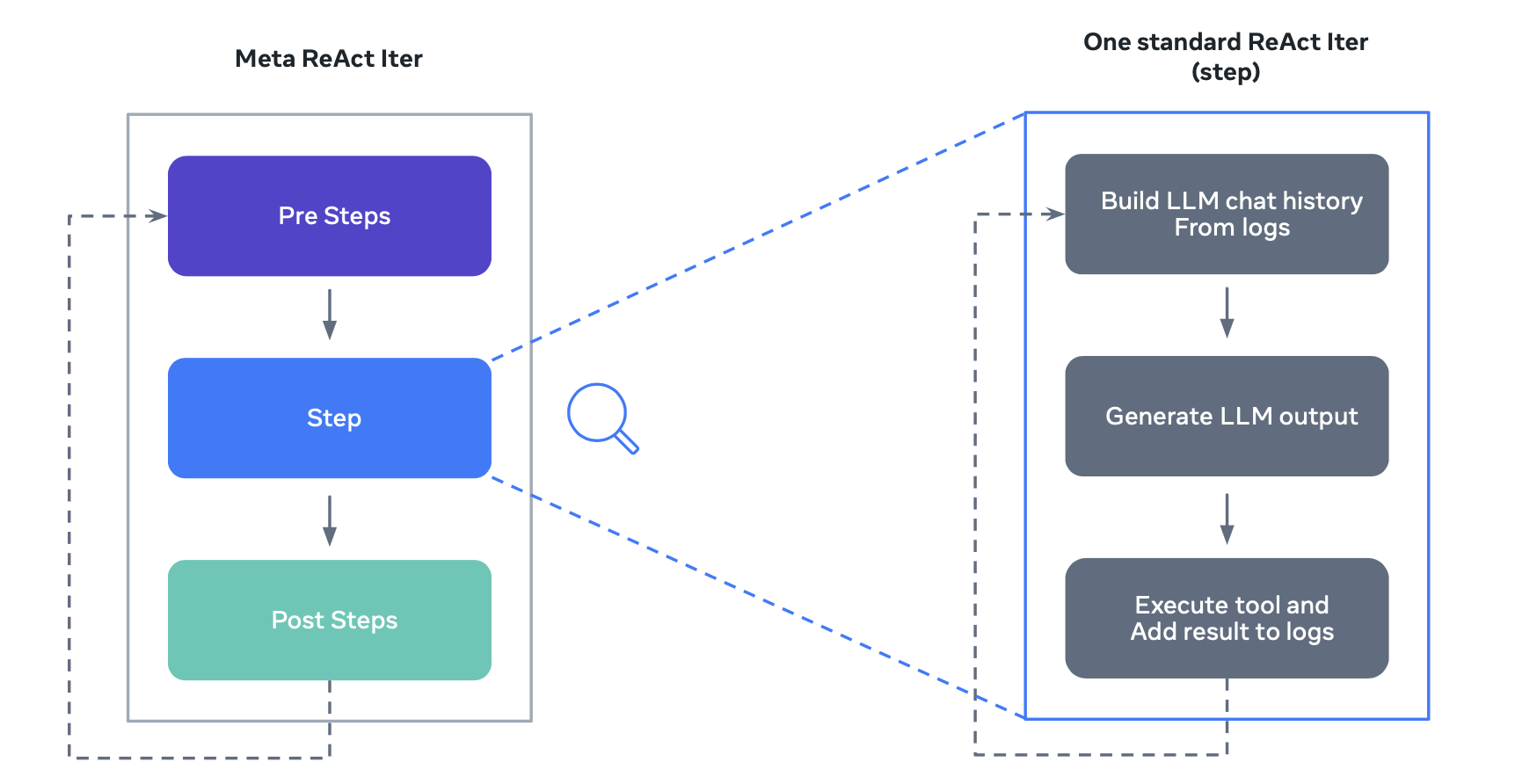}
    \caption{Proposed ReAct loop with pre/post steps in \gaiatwo, allowing flexible behaviors.}
    \label{fig:agent_pre_post_steps_gaia2}
\end{figure*}

At each step of this loop, our scaffolding triggers configurable pre-step and post-step methods that can pull relevant information from the environment state or detect termination conditions based on task-specific criteria as detailed in \autoref{fig:agent_pre_post_steps_gaia2}. Pre-step methods gather contextual information and validate preconditions before action execution, while post-step methods process outcomes, update internal state, and check for completion signals. This agentic modeling approach enables the creation of sophisticated agent behaviors with minimal implementation overhead, as complex interaction patterns emerge from the composition of simple, reusable scaffolding components rather than monolithic agent implementations.

\subsection{Experimental Setup and Implementation Details}
\label{app:experimental_setup}
We report \gaiatwo scores on a representative set of models, covering both proprietary and open-source systems, and including both reasoning-oriented and non-reasoning models. 

For evaluation, we use a ReAct scaffold that requires a \texttt{Thought:} and an \texttt{Action:} at each step. Since some models do not reliably follow this format, we add custom stop sequences—\texttt{<end\_action>} and \texttt{Observation:}—for models that tend to continue past a single tool call (Claude, Kimi, Qwen). This issue is largely alleviated by provider-specific ToolCalling APIs; we encourage reporting results with either interface (ReAct or ToolCalling).

Due to cost and time constraints, we did not evaluate every available model. For instance, Claude~4 Opus was excluded because of its very high latency and cost (\$15/M input tokens and \$75/M output tokens). We plan on evaluating and releasing other models scores, including DeepSeek and GPT-OSS models.

Finally, we note the following special configurations for certain third-party models:
\begin{itemize}
    \item \textbf{Gemini~2.5 Pro:} dynamic reasoning enabled via \texttt{budget\_reasoning\_tokens = -1}. 
    \item \textbf{Grok-4:} reasoning budget capped at 16k tokens per completion. We encountered many issues with xAI's API, notably with a lot of \texttt{Empty Response} errors, causing high-variance in our reported results.
    \item \textbf{GPT-5:} temperature and top-$p$ set to 1; no custom stop sequences were applied (not supported by the API).
\end{itemize}

\subsection{Additional Experiments}

\begin{figure}[htpb]
    \centering
    \includegraphics[width=0.9\linewidth]{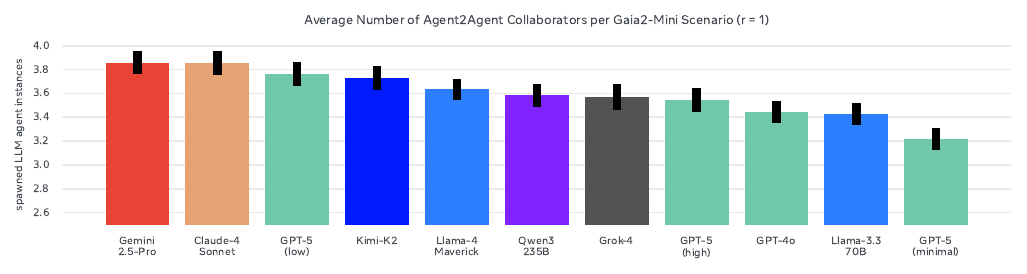}
    \caption{Average number of agents spawned in Agent2Agent evaluations on \gaiatwo-mini tasks across models. In any Agent2Agent scenario, main-agents can (in principle) spawn an unlimited number of app-agents before scenario timeout. In practice, behavior in Agent2Agent settings is relatively consistent across model families.}
    \label{fig:num_collabs_with_a2a}
\end{figure}

In our Agent2Agent experiments, we record the number of instantiated sub-agents in \autoref{fig:num_collabs_with_a2a}. Counts are fairly consistent across model families, yet the top A2A performers also spawn more sub-agents, suggesting stronger task decomposition.

\subsubsection{Influence of Noise Level on \gaiatwo\ Results}
\label{sec:gaia2_noise_experiments}
In this experiment, we vary the probability of tool errors and frequency of random environment events and measure resulting model results on \gaiatwo. While our lowest level of noise does not significantly impact model performance, increasing noise results in deteriorating performance across models. This aligns with our intuitions.

\begin{table}[htpb]
    \centering
    \begin{NiceTabular}{ccccc}
    \toprule
        & \multicolumn{4}{c}{\textbf{Noise level}}  \\
         \cmidrule{2-5} 
         & None & Low & Medium* & High \\ 
         \midrule 
         \textbf{Claude-4 Sonnet} & 31.2 & 35.0 & 23.8 & 8.1 \\
    \bottomrule
    \end{NiceTabular}
    \caption{Model performance on \gaiatwo-mini across different noise levels. *Default \gaiatwo setting.}
    \label{tab:noise_results}
\end{table}

\end{document}